\documentclass[10pt,twocolumn,letterpaper]{article}

\usepackage{iccv}
\usepackage{times}
\usepackage{epsfig}
\usepackage{graphicx}
\usepackage{amsmath}
\usepackage{amssymb}
\usepackage{cite}
\usepackage{amsmath,amssymb,amsfonts}
\usepackage{textcomp}
\usepackage{subfigure}
\usepackage[linesnumbered,boxed,ruled,commentsnumbered]{algorithm2e}
\usepackage{algpseudocode}  
\usepackage{amsmath}  
\UseRawInputEncoding
\newtheorem{theorem}{Theorem}
\newtheorem{proposition}{Proposition}
\newtheorem{lemma}{Lemma}
\newtheorem{remark}{Remark}
\newtheorem{problem}{Problem}
\newtheorem{proof}{Proof}
\usepackage[pagebackref=true,breaklinks=true,letterpaper=true,colorlinks,bookmarks=false]{hyperref}

\iccvfinalcopy 

\def\iccvPaperID{****} 

\ificcvfinal\fi

\begin{document}

\title{IRON: Invariant-based Highly Robust Point Cloud Registration}

\author{Lei Sun\\
School of Mechanical and Power Engineering\\
East China University of Science and Technology\\
{\tt\small leisunjames@126.com}
}

\maketitle
\ificcvfinal\thispagestyle{empty}\fi

\begin{abstract}
   In this paper, we present IRON (Invariant-based global Robust estimation and OptimizatioN), a non-minimal and highly robust solution for point cloud registration with a great number of outliers among the correspondences. To realize this, we decouple the registration problem into the estimation of scale, rotation and translation, respectively. Our first contribution is to propose RANSIC (RANdom Samples with Invariant Compatibility), which employs the invariant compatibility to seek inliers from random samples and robustly estimates the scale between two sets of point clouds in the meantime. Once the scale is estimated, our second contribution is to relax the non-convex global registration problem into a convex Semi-Definite Program (SDP) in a certifiable way using Sum-of-Squares (SOS) Relaxation and show that the relaxation is tight. For robust estimation, we further propose RT-GNC (Rough Trimming and Graduated Non-Convexity), a global outlier rejection heuristic having better robustness and time-efficiency than traditional GNC, as our third contribution. With these contributions, we can render our registration algorithm, IRON. Through experiments over real datasets, we show that IRON is efficient, highly accurate and robust against as many as 99\% outliers whether the scale is known or unknown, outperforming the existing state-of-the-art algorithms.
\end{abstract}

\section{Introduction}

Point cloud registration is a cornerstone in computer vision and robotics and has been widely applied in scene reconstruction~\cite{blais1995registering,henry2012rgb,choi2015robust}, object localization~\cite{drost2010model,papazov2012rigid,guo20143d}, medical imaging~\cite{audette2000algorithmic}, Simultaneous Localization and Mapping (SLAM)~\cite{zhang2014loam}, etc. It aims to estimate the transformation (scale, rotation and translation) between two point clouds.

Iterative Closest Point (ICP)~\cite{besl1992method} is a common local-minimum registration approach but is too dependent on the initial guess of the transformation, thus easy to fail without a good initialization. Therefore, an initialization-free way to address the registration problem\raisebox{0.3mm}{---}to build correspondences between point clouds from 3D keypoints~\cite{tam2012registration}, has grown increasingly popular. Many correspondence-based methods~\cite{horn1987closed,arun1987least,zhou2016fast,briales2017convex,bustos2017guaranteed,yang2019polynomial,yang2020teaser} have been proposed to solve the registration problem in the presence of Gaussian noise.

However, there exists a difficulty for all these correspondence-based methods. In real-world scenes, the input correspondences are often corrupted by outliers since the keypoint detecting and matching process~\cite{rusu2008aligning,tombari2013performance} may render mismatched or spurious results. This situation makes the robust registration methods of paramount significance.

Though RANSAC~\cite{fischler1981random} is a common robust heuristic~\cite{andrew2001multiple}, it has two critical disadvantages: (i) its runtime increases exponentially with the outlier ratio, and (ii) it is fragile when encountering extreme outlier ratios (e.g. 90\%+). Thus, an outlier removal method GORE \cite{bustos2017guaranteed} as well as non-minimal solutions FGR \cite{zhou2016fast} and TEASER/TEASER++ \cite{yang2019polynomial,yang2020teaser} have been proposed to deal with over 90\% outliers when the scale is known. But when the scale varies (e.g. cross-source point clouds), none of them could tolerate more than 90\% outliers, which necessitates a general solver capable of handling not only known-scale scenes but unknown-scale ones.

\begin{figure}[t]
\centering
\subfigure{
\begin{minipage}[t]{0.49\linewidth}
\centering
\includegraphics[width=1\linewidth]{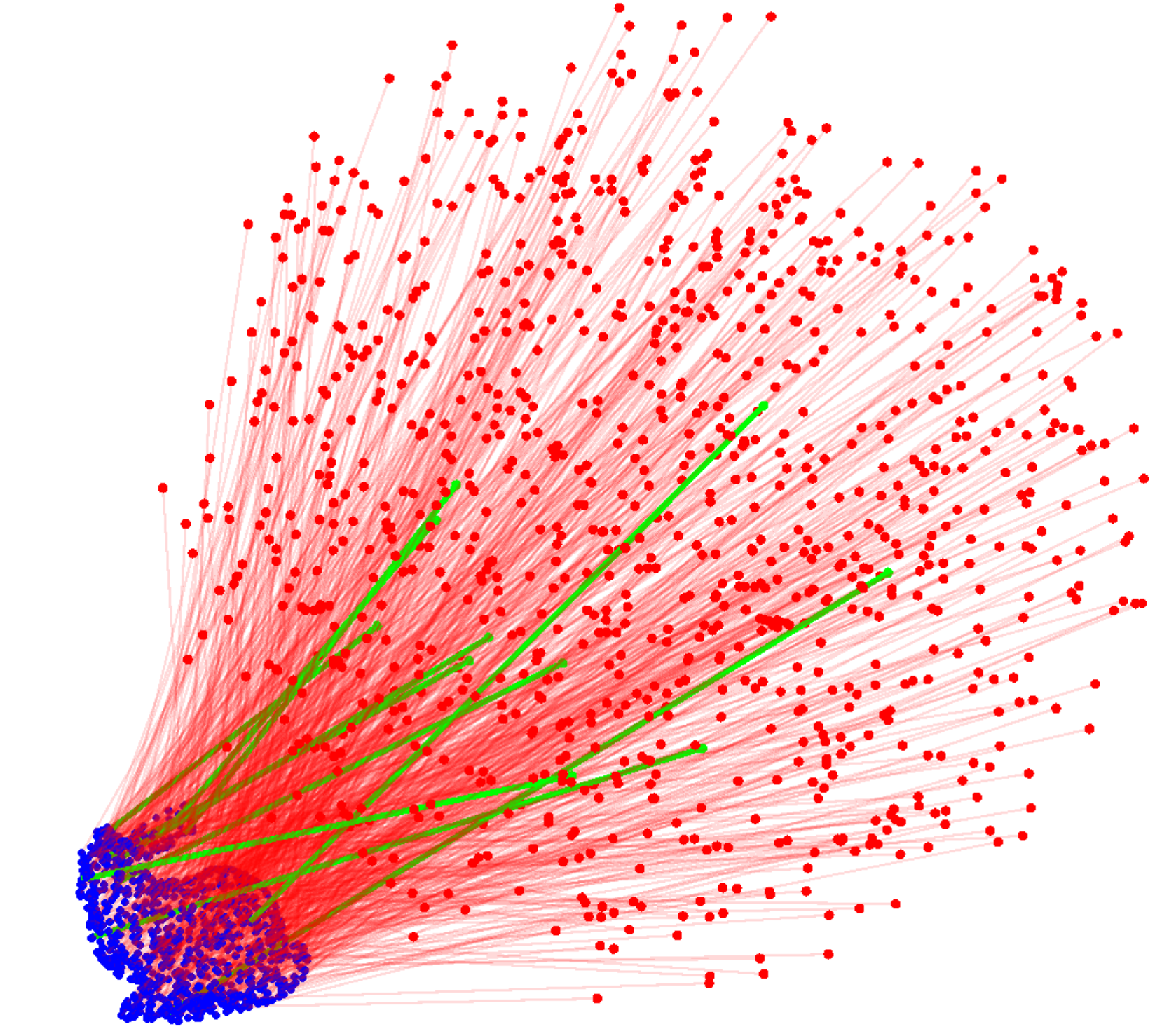}
\end{minipage}
}%
\subfigure{
\begin{minipage}[t]{0.49\linewidth}
\centering
\includegraphics[width=1\linewidth]{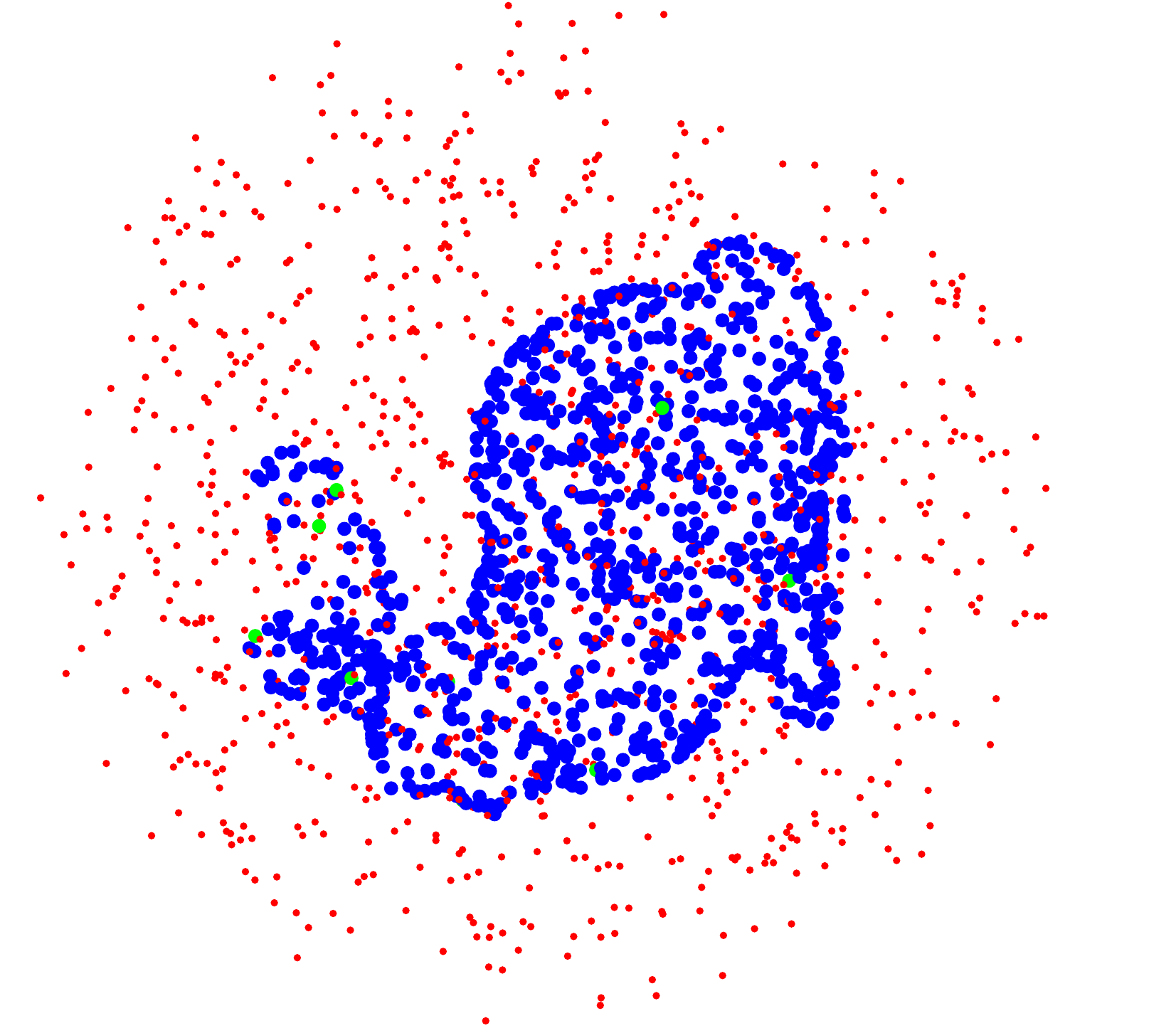}
\end{minipage}
}%

\centering
\caption{Demonstration of an unknown-scale noisy point cloud registration problem. \textbf{Left:} The transformed 'bunny' point cloud ($N=1000$) \cite{curless1996volumetric} is corrupted by 99\% outliers (red lines) and only 10 inlier correspondences (green lines) remain. \textbf{Right:} Our solver IRON can still estimate the transformation ($\mathit{s},\boldsymbol{R},\boldsymbol{t}$) precisely.}
\label{BG}
\end{figure}

In this paper, we design a non-minimal correspondence-based approach, named IRON (Invariant-based global Robust estimation and OptimizatioN), to robustly estimate the scale, rotation and translation in cascade.

\textbf{Highlights.} First, we invent RANSIC (RANdom Samples with Invariant Compatibility), a novel scale estimation and inlier searching method based on invariant compatibility that can tolerate very extreme outliers (e.g. 99\%).

Second, we employ the Sum-of-Squares (SOS) Relaxation of Lasserre's Hierarchy~\cite{lasserre2001global} to the known-scale robust global registration problem, leading to a certifiably globally optimal solver with strong optimality guarantees.

Third, we propose RT-GNC (Rough Trimming and Graduated Non-Convexity) to significantly accelerate the convergence and promote robustness simultaneously, as a highly efficient and robust heuristic for the registration problem.

Finally, we test IRON over real datasets and show that: (i) IRON is precise and robust even when 990 of 1000 correspondences are outliers in both known-scale and unknown-scale problems, outshining other methods, and (ii) IRON can be effectively applied in object localization problems.

\section{Related Work}

In this section, we provide reviews on: (i) some typical correspondence-based registration solvers by dividing them into two categories: the naive (outlier-free) ones and robust ones, and (ii) popular robust heuristics for outlier rejection.

\textbf{Naive Registration Solvers.} Naive registration problems can be solved using closed-form solutions \cite{horn1987closed,arun1987least} by separating the estimation of rotation, scale and translation. Besides, Branch-and-Bound (BnB) has been applied to solve the registration problem	\cite{olsson2008branch,parra2014fast} in a globally optimal way. And more recently, the Lagrangian Dual Relaxation is developed to produce a globally optimal generalized registration solver~\cite{briales2017convex} with good optimality guarantees.

\textbf{Robust Registration Solvers.} The minimal closed-form solution \cite{horn1987closed} can be combined with RANSAC to deal with outliers. In terms of non-minimal robust solvers, FGR~\cite{zhou2016fast} is the first method using Graduated Non-Convexity (GNC), which solves the transformation matrix with a Gauss-Newton method. Though not globally optimal, it is efficient for known-scale registration. GORE~\cite{bustos2017guaranteed} is a preprocessing outlier removal approach that guarantees that the correspondences removed must not be inliers. Currently, TEASER/TEASER++~\cite{yang2019polynomial,yang2020teaser} are globally optimal solvers that can dispose of 99\% outliers for known-scale registration. Another recent solution~\cite{li2021point} utilizes RANSAC for scale and translation estimation. Unfortunately, all these solvers are either incapable of estimating the scale or unable to tolerate over 90\% outliers for unknown-scale registration.

\textbf{Robust Heuristics.} RANSAC is a hypothesize-and-test heuristic, which usually requires minimal solvers (or very fast non-minimal solvers) to obtain the best consensus set. Its limitations lie in exponentially increasing runtime and relatively low robustness. When the solver is non-minimal, GNC~\cite{yang2020graduated} is a general-purpose heuristic operated by alternating optimization, which can typically handle 70-80\% outliers in many robotic or computer vision problems.

\section{Preliminaries}

\subsection{Problem Formulation}

With two corresponding 3D point sets $\{\mathbf{P}_i\}_{i=1}^N$ and $\{\mathbf{Q}_i\}_{i=1}^N$ in the correspondence-based point cloud registration problem, we can have the following relation:
\begin{equation}\label{Definition}
\mathit{s}\boldsymbol{R}\mathbf{P}_i+\boldsymbol{t}+\boldsymbol{\varepsilon}_i=\mathbf{Q}_i,
\end{equation}
where $\mathit{s}>0$ is the scale, $\boldsymbol{R}\in SO(3)$\footnote[1]{The explicit expression of $SO(3)$ can be found in \cite{tron2015inclusion,briales2017convex}.} is the rotation matrix, $\boldsymbol{t}\in\mathbb{R}^3$ is the translation vector, tuple $(\mathbf{P}_i, \mathbf{Q}_i)$ denotes the {$i_{th}$} putative point-to-point correspondence, and $\boldsymbol{\varepsilon}_i\in\mathbb{R}^3$ represents the measurement of Gaussian noise with isotropic covariance $\sigma^2\mathbf{I}_3$ and 0-mean w.r.t. the {$i_{th}$} correspondence. We then define a registration problem below.

\begin{problem}[Point Cloud Registration Problem]\label{Prob-1}
Solving the point cloud registration problem with noise is equivalent to the following global minimization problem:
\begin{equation}\label{Optimization}
\underset{\mathit{s}>0,\boldsymbol{R}\in\mathbf{SO}3, \boldsymbol{t}\in\mathbb{R}^3}{\min} \sum_{i=1}^N||\mathit{s}\boldsymbol{R}\mathbf{P}_i+\boldsymbol{t}-\mathbf{Q}_i||^2,
\end{equation}
where function \eqref{Optimization} is a Sum-of-Squares (SOS) polynomial.
\end{problem}

\subsection{SOS Relaxation of Lasserre's Hierarchy}

Lasserre~\cite{lasserre2001global} derived a sequence of SOS Relaxations of Lasserre's Hierarchy for the global optimization of SOS polynomials. Based on Putinar's Positivstellensatz~\cite{putinar1993positive} and Nie's finite convergence theory~\cite{nie2014optimality}, the SOS Relaxation with equality constraints can be summarized as follows.

\begin{theorem}[SOS Relaxation with Equality Constraints]\label{SOS-R}
Assume that we have a constrained optimization problem over the variables $\boldsymbol{x}\in\mathbb{R}^n$ such that:
\begin{equation}
\begin{gathered}
f^{\star}:=\underset{\boldsymbol{x}\in\mathcal{K}}{\min}{f}(\boldsymbol{x}), \\
\mathcal{K}:=\left\{{h}_i(\boldsymbol{x})=0, i=1,2,\dots,m\right\},
\end{gathered}
\end{equation}
where the objective ${f}(\boldsymbol{x})\in\mathbb{R}[\boldsymbol{x}]$ \footnote[2]{Here $\mathbb{R}[\boldsymbol{x}]$ denotes the ring of real-valued polynomials w.r.t. $\boldsymbol{x}$.} is a SOS polynomial and $\mathcal{K}$ is the feasible set in which ${h}_i(\boldsymbol{x})\in\mathbb{R}[\boldsymbol{x}]$.
If the $2\theta_{th}$ \textit{truncated ideal} of $\boldsymbol{h}:=(h_1,h_2,\dots,h_m)$ that is given by
\begin{equation}
\langle \boldsymbol{h}\rangle_{2 \theta}:=\left\{\sum_{i=1}^{m} {\phi}_{i} {h}_{i} | \begin{array}{c}{\phi}_{i} \in \mathbb{R}[\boldsymbol{x}], \operatorname{degree}\left({\phi}_{i} h_{i}\right) \leq 2 \theta,
\end{array}\right\}
\end{equation}
is Archimedean (e.g. $X-\sum_{j=1}^{n}x_j^2\in \langle \boldsymbol{h}\rangle_{2\theta}$ for some $\theta\in\mathbb{N}$ and $X>0$), then we can derive a sequence of SOS Relaxations of relaxation order $\theta$ based on Lasserre's Hierarchy:
\begin{equation}
\begin{gathered}
f^{*}_{\theta}:=\max \tau, \\
s.t.\,\, {f}(\boldsymbol{x})-\tau = {\phi}_0(\boldsymbol{x})+g(\boldsymbol{x}) ,
\end{gathered}
\end{equation}
where ${\phi}_0(\boldsymbol{x})$ is a SOS polynomial of degree at most $2\theta$ and $g(\boldsymbol{x}) \in \langle \boldsymbol{h} \rangle_{2\theta}$. And this hierarchy has finite convergence generically, meaning that $f^{*}_{\theta}=f^{\star}$ with some $\theta$.
\end{theorem}

\subsection{Our Method}

Since simultaneously estimating the scale, rotation and translation in the presence of both noise and outliers is difficult and impractical, our strategy is to decouple the registration problem into three subproblems given in the following sections: (i) estimating the scale and finding a portion of inliers from random samples using the compatibility of invariants (RANSIC), (ii) certifiably globally solving the optimal rotation with SOS Relaxation as well as recovering the translation, and (iii) combining the naive solver with a new robust heuristic to globally reject outliers (RT-GNC).

\section{Scale Estimation and Inlier Searching}
\label{Scale Estimation}

This section provides a novel approach, RANSIC, to estimate the scale and search for inliers at the same time. Different from most previous methods (e.g.~\cite{yang2019polynomial,yang2020teaser,bustos2017guaranteed}), we aim to seek the inliers, though not all, rather than removing outliers to diminish the correspondence set. TEASER and ROBIN~\cite{shi2020robin} make use of pairwise invariants to roughly reject outliers and build the undirected graph; in this work, we explore the use of invariants in finding inliers. Our method is built upon the geometric constraints of an inlier set consisting of 3 point correspondences, partially inspired by Horn's closed-form solution~\cite{horn1987closed} and ROBIN.

\subsection{First Compatibility: Scale and Translation}

Following Problem \ref{Prob-1}, assume that we have 3 random non-colinear correspondences {$(\mathbf{P}_i, \mathbf{Q}_i), i=1,2,3$}, here defined as a \textit{3-point set}. The centroids w.r.t. $\{\mathbf{P}_i\}_{i=1}^3$ and $\{\mathbf{Q}_i\}_{i=1}^3$ can be computed, respectively, as
\begin{equation}\label{RANSIC-1}
\mathbf{\bar{P}}=\frac{1}{3} \sum^{3}_{i=1}\mathbf{P}_i, \,\,\,\mathbf{\bar{Q}}=\frac{1}{3}\sum^{3}_{i=1}\mathbf{Q}_i.
\end{equation}
Hence, each point's corresponding vector directed from the centroid to its own position in 3D space can be written as
\begin{equation}\label{RANSIC-2}
\mathbf{\widetilde{P}}_i=\mathbf{P}_i-\mathbf{\bar{P}}, \,\,\,\mathbf{\widetilde{Q}}_i=\mathbf{Q}_i-\mathbf{\bar{Q}}.
\end{equation}
Here, we introduce a translation-free technique (first appeared in~\cite{arun1987least}) based on the centroids \eqref{RANSIC-1} such that
\begin{equation}\label{RANSIC-3}
\boldsymbol{t}=\mathbf{\bar{Q}}-\boldsymbol{R}\mathbf{\bar{P}},
\end{equation}
where since noise $\boldsymbol{\varepsilon}_i$ is isotropic, we can omit the measurement of noise in this equation.
\begin{proposition}[Scale-based Invariant] \label{Prop1} We define a set of functions $\{{s}_i(\mathbf{P}_i, \mathbf{Q}_i)\}^{3}_{i=1}$ satisfying that
\begin{equation}\label{RANSIC-4}
{s}_i(\mathbf{P}_i, \mathbf{Q}_i)=\frac{||\mathbf{\widetilde{Q}}_i||}{||\mathbf{\widetilde{P}}_i||}=\mathit{s}+\epsilon_i,
\end{equation}
where each ${s}_i(\mathbf{P}_i, \mathbf{Q}_i)$ is invariant to $\boldsymbol{R}$ and $\boldsymbol{t}$ and mathematically represents the scale computed from the $i_{th}$ correspondence, and $\epsilon_i$ is a scalar denoting the difference between the computed scale $\mathit{s}_i$ and the ground-truth scale $\mathit{s}$.
\end{proposition}

Proposition \ref{Prop1} indicates that by building a 3-point set, we can have 3 functions invariant to the rotation and translation, in other words, solely variant to the scale. Intuitively, each function denotes the scale obtained with its respective correspondence under the perturbance of noise. The next step is to set constraints to these functions based on noise.

According to Problem \eqref{Prob-1} , it is apparent to obtain
\begin{equation}\label{RANSIC-4+5}
\mathit{s}\boldsymbol{R}\mathbf{\widetilde{P}}_i+\boldsymbol{\varepsilon}_i=\mathbf{\widetilde{Q}}_i, \Rightarrow \, ||\mathit{s}\boldsymbol{R}\mathbf{\widetilde{P}}_i||=||\mathbf{\widetilde{Q}}_i-\boldsymbol{\varepsilon}_i||.
\end{equation}
Using the triangle inequality, we can derive that
\begin{equation}\label{RANSIC-6}
\begin{gathered}
||\mathbf{\widetilde{Q}}_i||-||\boldsymbol{\varepsilon}_i|| \leq ||\mathit{s}\boldsymbol{R}\mathbf{\widetilde{P}}_i||= \mathit{s}||\mathbf{\widetilde{P}}_i|| \leq ||\mathbf{\widetilde{Q}}_i||+||\boldsymbol{\varepsilon}_i||,
\end{gathered}
\end{equation}
and if we let $\alpha$ be the bound of the norm of noise $\boldsymbol{\varepsilon}_i$ $(||\boldsymbol{\varepsilon}_i|| \leq\alpha)$, then we can have
\begin{equation}\label{RANSIC-7}
\begin{gathered}
\frac{||\mathbf{\widetilde{Q}}_i||-\boldsymbol{\varepsilon}_i}{||\mathbf{\widetilde{P}}_i||} \leq \mathit{s} \leq \frac{||\mathbf{\widetilde{Q}}_i||+\boldsymbol{\varepsilon}_i}{||\mathbf{\widetilde{P}}_i||},
\end{gathered}
\end{equation}
which leads to the following compatibility condition.
\begin{proposition}[Scale-based Compatibility] \label{Prop2} In the presence of noise, if the 3-point set only consists of inliers, the invariant functions $\{{s}_i(\mathbf{P}_i, \mathbf{Q}_i)\}_{i=1}^3$ should satisfy the compatibility condition such that
\begin{equation}\label{RANSIC-8}
||{s}_i(\mathbf{P}_i, \mathbf{Q}_i)-{s}_j(\mathbf{P}_j, \mathbf{Q}_j)||\leq \alpha(\frac{1}{||\mathbf{\widetilde{P}}_i||}+\frac{1}{||\mathbf{\widetilde{P}}_j||}),
\end{equation}
where $(i,j)\in\{(1,2),(1,3),(2,3)\}$.
\end{proposition}

Proposition \ref{Prop2} states that as long as the 3 correspondences in the 3-point set are all inliers, they must satisfy the inequalities \eqref{RANSIC-8} regarding the scale, which makes up the first part of our first compatibility condition. The second part is a translation-based condition. But before that, we need to roughly compute the scale and rotation from the 3-point set first. For the scale, we use the least-squares solution in~\cite{yang2019polynomial}:
\begin{equation}\label{RANSIC-9}
\mathit{\hat{s}}=\frac{1}{\sum_{i=1}^3 \upsilon_i}\cdot \sum_{i=1}^3 \upsilon_i\boldsymbol{s}_i(\mathbf{P}_i, \mathbf{Q}_i),
\end{equation}
where $\upsilon_i=\frac{||\mathbf{\widetilde{P}}_i||^2}{\alpha^2}$.
For the rotation, we adopt the triad-based method in \cite{horn1987closed}, which merely requires some dot and cross products and is rather efficient to implement. Based on a 3-point triad, the rotation\footnote[3]{The explicit computation can be found in the supplementary material.} can be linearly obtained as
\begin{equation}\label{RANSIC-10}
\boldsymbol{\widetilde{R}}=\left[\begin{array}{c}
\mathbf{n}_1\, \,\mathbf{n}_2\, \,\mathbf{n}_3
\end{array}\right]\left[\begin{array}{c}
\mathbf{m}_1 \,\,\mathbf{m}_2 \,\,\mathbf{m}_3
\end{array}\right]^{\top},
\end{equation}
where unit vectors $\mathbf{m}_i, \mathbf{n}_i\in\mathbb{R}^3$, and $\{\mathbf{m}_i\}_{i=1}^3$ and $\{\mathbf{n}_i\}_{i=1}^3$ denote the coordinate systems before and after the transformation, respectively.

\begin{proposition}[Translation-based Invariant] \label{Prop3} We then define another set of functions $\{\boldsymbol{t}_i(\mathbf{P}_i, \mathbf{Q}_i)\}^{3}_{i=1}$ satisfying that
\begin{equation}\label{RANSIC-14}
\boldsymbol{t}_i(\mathbf{P}_i, \mathbf{Q}_i)=\mathbf{Q}_i-\mathit{\hat{s}}\boldsymbol{\widetilde{R}}\mathbf{P}_i=\boldsymbol{t}+\boldsymbol{\varepsilon}_i,
\end{equation}
where each $\boldsymbol{t}_i(\mathbf{P}_i, \mathbf{Q}_i)$ is invariant to $\mathit{s}$ and $\boldsymbol{R}$, and it represents the translation computed from the $i_{th}$ correspondence perturbed by noise.
\end{proposition}

Now we can impose another compatibility condition for the difference between any two $\boldsymbol{t}_i(\mathbf{P}_i, \mathbf{Q}_i)$.
\begin{proposition}[Translation-based Compatibility] \label{Prop4} In the presence of noise, if the 3-point set only consists of inliers, functions $\{\boldsymbol{t}_i(\mathbf{P}_i, \mathbf{Q}_i)\}^{3}_{i=1}$ should satisfy that
\begin{equation}\label{RANSIC-15}
||\boldsymbol{t}_i(\mathbf{P}_i, \mathbf{Q}_i)-\boldsymbol{t}_j(\mathbf{P}_j, \mathbf{Q}_j)||\leq 2\beta,
\end{equation}
where $(i,j)\in\{(1,2),(1,3),(2,3)\}$ and $||\boldsymbol{\varepsilon}_i||\leq\beta$.
\end{proposition}

Proposition \ref{Prop2} and \ref{Prop4} jointly constitute the first compatibility condition (scale and translation compatibility) so that we can eliminate a great deal of outliers from random 3-point set samples and enable the potential inliers that pass the first condition to be further tested by the second compatibility.

\subsection{Second Compatibility: Completeness}

Assume that we already have $M$ 3-point sets that pass the first compatibility condition: {$\mathcal{M}=\{(\mathbf{P}_i, \mathbf{Q}_i)_k, i=1,2,3\}_{k=1}^M$}, called the \textit{potential 3-point sets}. We can now define a new set of rotation-based invariant functions.
\begin{proposition}[Rotation-based Invariant] \label{Prop5} We define a set of functions $\{\boldsymbol{r}_k(\mathbf{P}_i, \mathbf{Q}_i)\}_{k=1}^M$ satisfying that
\begin{equation}\label{RANSIC-16}
\boldsymbol{r}_k(\mathbf{P}_i, \mathbf{Q}_i)=\boldsymbol{\widetilde{R}}_k=\boldsymbol{{R}}\cdot \mathbf{Exp}(\boldsymbol{\eta}_k)
\end{equation}
where, intuitively, $\boldsymbol{r}_k(\mathbf{P}_i, \mathbf{Q}_i)$ is the rotation $\boldsymbol{\widetilde{R}}_k$ computed from the $k_{th}$ set and invariant to $\mathit{s}$, $\boldsymbol{R}$ and $\boldsymbol{t}$, and $\mathbf{Exp}$ is the exponential map \cite{barfoot2017state} used to describe the noise perturbance.
\end{proposition}

Then, we can build another invariant compatibility condition for every two potential 3-point sets.
\begin{proposition}[Complete Compatibility] \label{Prop6} For any two potential 3-point sets {$\{(\mathbf{P}_i, \mathbf{Q}_i)_k, i=1,2,3\}_{k=1}^2\subset\mathcal{M}$}, in the presence of noise, we have the following conditions: \\ (i) functions $\{\boldsymbol{r}_k(\mathbf{P}_i, \mathbf{Q}_i)\}^{2}_{k=1}$ should satisfy that
\begin{equation}\label{RANSIC-15}
trace(\boldsymbol{r}_1(\mathbf{P}_i, \mathbf{Q}_i)^{\top}\boldsymbol{r}_2(\mathbf{P}_j, \mathbf{Q}_j))\leq \gamma,
\end{equation}
where $\gamma$ here denotes the bound of noise;\\
(ii) concatenating the two 3-point sets as a 6-point set: {$\{(\mathbf{P}_i, \mathbf{Q}_i), l=1,2,\dots,6\}$}, we can also have 
\begin{align} ||{s}_a(\mathbf{P}_a, \mathbf{Q}_a)-{s}_b(\mathbf{P}_b, \mathbf{Q}_b)||&\leq \alpha(\frac{1}{||\mathbf{\widetilde{P}}_a||}+\frac{1}{||\mathbf{\widetilde{P}}_b||}), \\ ||\boldsymbol{t}_a(\mathbf{P}_a, \mathbf{Q}_a)-\boldsymbol{t}_b(\mathbf{P}_b, \mathbf{Q}_b)||&\leq 2\beta, 
\end{align}
where subscript $a,b\in\{1,2,\dots,6\}\,\,(a\neq b)$ and there are $\left(\begin{array}{c}
6 \\ 2 \end{array}\right)\cdot 2=30$ inequalities in total for condition (ii).
\end{proposition}

Proposition \ref{Prop6} (its proof can be seen in the supplementary material) constitutes the second compatibility condition (complete compatibility), as our ultimate condition to extract qualified inliers from the potential 3-point sets.

\subsection{RANSIC}
\textbf{Algorithm.} Our method RANSIC can be implemented in the following steps: (i) to obtain a random 3-point set from the correspondences, (ii) to check the first compatibility condition (Proposition \ref{Prop2} and \ref{Prop4}) and preserve all potential 3-point sets, and (iii) to check the second compatibility condition (Proposition \ref{Prop6}) between any two inlier 3-point sets to find the ultimate inliers. These steps are specified by the pseudocode in Algorithm \ref{Algo1-RANSIC}, where function \textit{getUniqueSet} gets the serial numbers of all the correspondences in the inlier 3-point set $\mathcal{N}^{In}$ and removes the repeated ones.

\textbf{Novelty.} Different from RANSAC, RANSIC can eliminate a large number of outliers at the beginning using the scale-based condition \ref{Prop2} so that only a small part of the samples are used to compute the rotation and translation, where the computation process can be rapidly achieved with simple dot and cross products and matrix transformations. Besides, the goal of RANSIC is merely to find a part of inliers (usually fewer than 10), so building the consensus set by calculating the residual errors w.r.t all correspondences is not required, making RANSIC much more efficient than RANSAC (Section \ref{Subsection-subproblems}). Furthermore, RANSIC is a preprocessing algorithm for our heuristic RT-GNC since the inliers it sought can make the RT-GNC more robust (Section \ref{Heuristic}).

\textbf{Advantage.} Unlike other scale estimators (e.g. \cite{yang2020teaser,li2021point}), RANSIC's runtime is completely independent of the correspondence number and only relies on the outlier ratio, hence applicable for large problems. And when the outlier ratio is no more than 98\%, RANSIC has outstanding running speed.

\begin{algorithm}
\caption{RANSIC}
\label{Algo1-RANSIC}
\SetKwInOut{Input}{\textbf{Input}}
\SetKwInOut{Output}{\textbf{Output}}
\Input{correspondences $\mathcal{N}=\{(\mathbf{P}_i,\mathbf{Q}_i)\}_{i=1}^N$; noise thresholds $\alpha, \beta, \gamma$; minimum number of the inlier 3-point sets $\mathit{X}$\;}
\Output{estimated scale $\mathit{\hat{s}}$; inlier set $\mathcal{N}^{In}$\;}
\BlankLine
Initiate $\mathcal{N'}=\emptyset$ \;
\While {true}{
{Randomly select $\mathcal{S}=\{(\mathbf{P}_j,\mathbf{Q}_j)\}_{j=1}^3\subset\mathcal{N}$}\;
\If {$\mathcal{S}$ can pass the first compatibility \ref{Prop2}\&\ref{Prop4}}{
Search for all the 3-point sets $\mathcal{S}_i\in\mathcal{N'}$ ($i=1,2,\dots,x$) that are in the second compatibility \ref{Prop6} with $\mathcal{S}$\;
\If {$x\geq \mathit{X}$}{
$\mathcal{N}^{In}=\{\mathcal{S}\}\cup\{\mathcal{S}_i\}_{i=1}^{x}$ \;
\textbf{break}}
$\mathcal{N'}=\mathcal{N'}\cup \{\mathcal{S}$\} \;
}
}
$\mathcal{N}^{In}=\textit{getUniqueSet}(\mathcal{N}^{In})$\;
Compute the scale $\mathit{\hat{s}}$ from inliers in $\mathcal{N}^{In}$ using \eqref{RANSIC-9}\;
\Return scale $\mathit{\hat{s}}$ and inlier set $\mathcal{N}^{In}$\;
\end{algorithm}

\section{Globally Optimal Point Cloud Registration}

In this section, we develop a certifiably globally optimal solver for the known-scale robust point cloud registration.
\begin{theorem}[Translation-free Robust Registration]\label{Th-1}
Problem \ref{Prob-1} with outliers is equivalent to the following translation-free weighted global optimization problem:
\begin{equation}\label{Go-1}
\underset{\boldsymbol{R}\in{SO}3}{\min} \sum_{i=1}^N \omega_i\left(\mathit{\hat{s}}^{2}\mathbf{\widetilde{P}}_i^{\top}\mathbf{\widetilde{P}}_i+\mathbf{\widetilde{Q}}_i^{\top}\mathbf{\widetilde{Q}}_i-2\mathit{\hat{s}}\mathbf{\widetilde{Q}}_i^{\top}\boldsymbol{R}\mathbf{\widetilde{P}}_i\right),
\end{equation}
where $\omega_i$ ($i=1,2,\dots,N$) denotes the weight w.r.t the $i_{th}$ correspondence, $\mathit{\hat{s}}$ is the scale estimated by RANSIC, and
\begin{equation}\label{Go-2}
\mathbf{\widetilde{P}}_i=\mathbf{P}_i-\mathbf{\bar{P}}_i,\,\,
\mathbf{\widetilde{Q}}_i=\mathbf{Q}_i-\mathbf{\bar{Q}}_i,
\end{equation}
where $\mathbf{\bar{P}}$ and $\mathbf{\bar{Q}}$ are derived from the translation-free technique in \eqref{RANSIC-3} such that
\begin{equation}\label{Go-3}
\mathbf{\bar{P}}=\frac{\sum_{i=1}^N(\omega_i \mathbf{P}_i)}{\sum_{i=1}^N\omega_i},
\,\,\mathbf{\bar{Q}}=\frac{\sum_{i=1}^N(\omega_i \mathbf{Q}_i)}{\sum_{i=1}^N\omega_i}.
\end{equation}
Once the optimal rotation $\boldsymbol{\hat{R}}$ is solved by minimizing \eqref{Go-1}, the optimal translation can be recovered by
\begin{equation}\label{t-free}
\boldsymbol{\hat{t}}=\mathbf{\bar{Q}}-\boldsymbol{\hat{R}}\mathbf{\bar{P}}.
\end{equation}
\end{theorem}

The proof can be seen in the supplementary material. Since the first two terms in \eqref{Go-1} are constant, we can further simplify the minimization problem by removing the constants and parametrize $\boldsymbol{\hat{R}}$ with unit quaternions, leading to a Quadratically Constrained Quadratic Program (QCQP).

\begin{proposition}[QCQP for Robust Registration]\label{Prop7} The global minimization problem \eqref{Go-1} can be equivalently reformulated as the following quaternion-based QCQP:
\begin{equation}\label{Go-4}
\begin{gathered}
f^{\star}:=\underset{\boldsymbol{q}\in\mathbb{R}^4}{\min}{f}(\boldsymbol{q})=\underset{\boldsymbol{q}\in\mathbb{R}^4}{\min} \sum_{i=1}^N \boldsymbol{q}^{\top}\mathbf{C}\boldsymbol{q}, \\
s.t.\,\, {h}(\boldsymbol{q})=1-\boldsymbol{q}^{\top}\boldsymbol{q}=0,
\end{gathered}
\end{equation}
where ${f}(\boldsymbol{q})$ is a quadratic polynomial, ${h}(\boldsymbol{q})$ is a quadratic constraint, $\mathbf{C}=-\mathit{\hat{s}}\sum_{i=1}^N \omega_i \mathbf{\Pi}_1(\mathbf{\widetilde{P}}_i^*)^{\top}\mathbf{\Pi}_2(\mathbf{\widetilde{Q}}_i^*)$ 
and $\mathbf{\widetilde{P}}_i^*=\left[\begin{array}{c} \mathbf{\widetilde{P}}_i \\ 0 \end{array}\right]$ and $\mathbf{\widetilde{Q}}_i^*=\left[\begin{array}{c} \mathbf{\widetilde{Q}}_i \\ 0 \end{array}\right]$. (The expression of $\mathbf{\Pi}_1$ and $\mathbf{\Pi}_2$ and the proof can be found in the supplementary material.)
\end{proposition}

Note that the objective \eqref{Go-1} is a SOS polynomial, as in~\cite{yang2020perfect}, we can apply the SOS Relaxation of Lasserre's Hierarchy \cite{lasserre2001global} to reduce \eqref{Go-4} to a moment SDP problem.

\begin{proposition}[SOS Relaxations for \eqref{Go-1}]\label{SOS-RP}
After applying Lasserre's Hierarchy to the global minimization \eqref{Go-1} and based on QCQP \eqref{Go-4}, it can be equivalently formulated as
\begin{equation}\label{Go-5}
\begin{gathered}
{f}^{*}_{\theta}:=\max \tau,\\
s.t.\,\, {f}(\boldsymbol{q})-\tau=\phi_0(\boldsymbol{q})+{\zeta}\cdot{h}(\boldsymbol{q}),
\end{gathered}
\end{equation}
where $\theta=1$, $\tau$ is the global lower bound of ${f}(\boldsymbol{q})$, $\phi_0(\boldsymbol{q})$ is a SOS polynomial (degree $\leq 2$), and ${\zeta}$ is a constant multiplier.
\end{proposition}

The SOS program \eqref{Go-5} can be readily modeled as a convex moment SDP problem~\cite{lasserre2001global} in a Generalized Moment Problem (GMP) solver {gloptipoly3}~\cite{henrion2009gloptipoly} and subsequently solved by a standard SDP solver (e.g. SeDuMi~\cite{sturm1999using}, Mosek~\cite{mosek2015mosek}). In Section \ref{Experiments}, we will show that the SOS Relaxation applied here has very promising global optimality. 

For faster implementation, the QCQP \eqref{Go-4} can be also solved by finding the eigenvector w.r.t. the smallest eigenvalue of the symmetric matrix $\mathbf{C}$ (IRON* in Section~\ref{Experiments}).

\section{Better Heuristic: Robust Outlier Rejection}\label{Heuristic}

This section proposes a new robust heuristic RT-GNC for the registration problem, which outperforms the traditional heuristic GNC~\cite{yang2020graduated} in two ways: (i) the robustness against outliers is better, and (ii) the iteration number needed for convergence is greatly reduced, thus faster for practical use.

\subsection{GNC with the Leclerc Cost Function}

We first introduce a GNC process using the Leclerc cost function \cite{leclerc1989constructing,black1996unification} as the robust function ${\rho}(\,\,)$, called GNC-LC, which underlies our heuristic RT-GNC.
\begin{lemma}[GNC-LC]\label{Lem1}
The robust registration problem \eqref{Go-4} can be globally solved by the following iterative optimization problem with outlier process:
\begin{equation}\label{GNC1}
\underset{\omega_i^{t}\in[0,1]}{\min} \sum_{i=1}^N \left[\omega_i^{t} r_i^{t}+\boldsymbol{\Psi}(\omega_i^{t},\mu^{t},\bar{r})\right]^{t},
\end{equation}
where $t$ indicates that this process is in the $t_{th}$ iteration, $\omega_i^{t}$ is the weight and $r_i^{t}={res}_i(\boldsymbol{\hat{q}}^{t})$ is the residual error w.r.t the $i_{th}$ correspondence, and function $\boldsymbol{\Psi}(\omega_i^{t},\mu^{t},\bar{r})$ is the outlier process satisfying that
\begin{equation}\label{GNC2}
\boldsymbol{\Psi}(\omega_i^{t},\mu^{t},\bar{r})={\mu^t}^2\bar{r}^2\left(\omega_i^t\ln(\omega_i^t)-\omega_i^t+1\right),
\end{equation}
where $\mu^{t}$ is the controlling parameter that is updated in each GNC iteration (usually by $\mu^{t}=\mu^{t-1}/1.05$ in the registration problem) and $\bar{r}$ is the threshold of all residual errors only dependent on noise, and the Leclerc robust function ${\rho}(r_i,\mu,\bar{r})$ can be expressed as 
\begin{equation}\label{GNC3}
{\rho}(r_i^{t},\mu^{t},\bar{r})=1-e^{-\frac{{r_i^t}^2}{{\mu^{t}}^2 \bar{r}^2}}.
\end{equation}
\end{lemma}

\textbf{Why the Leclerc function?} Our heuristic is built upon the GNC process and we empirically find that the Leclerc ${\rho}$-function \eqref{GNC3} is the one that fits our heuristic best.

\textbf{Improving Robustness and Efficiency.} Though the proposed GNC-LC can reject 90\% outliers within 50 iterations, already superior to GNC-TLS/-GM (85\%) \cite{yang2020graduated}, we can further improve it by introducing a trimming technique.

\subsection{RT-GNC}

ADAPT~\cite{tzoumas2019outlier} is an efficient approach for robust estimation, which employs minimally trimmed squares to reject outliers in an adaptive way. Now we combine the trimming technique with GNC-LC, aiming to not only boost the robustness but accelerate the convergence of the heuristic.
\begin{proposition}[Rough Trimming and GNC]\label{Prop8}
The Black-Rangarajan-Duality-and-GNC problem \eqref{GNC1} can be improved by a minimally trimming strategy as
\begin{equation}\label{GNC4}
\underset{i\in\mathcal{I}^t,\omega_i^t\in[0,1]}{\min} \sum_{i=1}^{|\mathcal{I}^t|} \left[\omega_i^t r_i^t+\boldsymbol{\Psi}(\omega_i^t,\mu^t,\bar{r})\right]^t,
\end{equation}
where set $\mathcal{I}^t$ is a subset of $\mathcal{I}^{t-1}$ satisfying that
\begin{equation}\label{GNC5}
\begin{gathered}
\mathcal{I}^t=\arg\underset{\mathcal{I}^t\subset\mathcal{I}^{t-1}}{\max}\,\, |\mathcal{I}^t|,\\
s.t. \sum_{j\in\mathcal{I}^t} {r_j^t}^2\leq\xi_{t},
\end{gathered}
\end{equation}
where $\xi_{t}$, reduced in each iteration, is a positive bound used to differentiate outliers and inliers, and in the first iteration $\mathcal{I}^1$ includes all the correspondences ($\mathcal{I}^1=\{1,2,\dots,N\}$).
\end{proposition}

For more convenient use of Proposition \ref{Prop8}, we provide a remark on its practical implementation procedures.
\begin{remark}[How to implement RT-GNC?]\label{Examp1}
Proposition \ref{Prop8} can be equivalently solved by the following alternating optimization process: in the $t_{th}$ iteration,  (i) we make an estimate by optimizing \eqref{Go-4} with weights $\boldsymbol{\omega}^t=[\omega_1,\omega_2,\dots,\omega_{N^t}]$ ($\omega_i$ are all 1 in the first iteration) to obtain the optimal solutions $\boldsymbol{\hat{q}}^t$ and ${\boldsymbol{\hat{t}}}^t$ and compute all the residual errors $r_i^t$, (ii) we update the weights in closed-form:
\begin{equation}\label{GNC6}
\omega_{i}^{t+1}=\left\{\begin{array}{ll}
0 & \text { if } {r^t_{i}}^2 > \xi^t \,or\, \omega_i^{t}=0, \\
e^{-\frac{{r^t_{i}}^2}{{\mu^t}^2\bar{r}^2}} & \text { if } {r^t_{i}}^2 \leq \xi^t,
\end{array}\right.
\end{equation}
and (iii) we update the controlling parameter by $\mu^{t+1}=\mu^{t}/\eta$ (default:$\eta=1.05$) and the inlier bound by
\begin{equation}\label{trimming-update}
\xi^{t+1}=\min\left(\xi^{t},\underset{i\in\mathcal{N}}{\max}({r_i^t}^2)\right)\cdot\nu,
\end{equation}
where usually $\nu\in[0.4,0.7]$ and $\xi^{1}=\mathit{\hat{s}}^2\cdot\max( {r^1_{i}})^2$. In this way, we can iteratively repeat this alternating process until the global minimum of objective \eqref{Go-4} converges or $\mu<1$.
\end{remark}

Rough Trimming can make GNC converge within 20 iterations and tolerate 2-3\% outliers more. But it can be even faster and more robust by taking the advantage of RANSIC.

RANSIC (Section \ref{Scale Estimation}) has already found the inliers $\mathcal{N}^{In}$. Now if we deliberately increase the weights of these inliers hundredfold in the first 3 iterations $\omega_i^{(1-3)}$ ($i\in\mathcal{N}^{In}$) while keeping the other weights unchanged $\omega_i^t\in[0,1]$ ($i\notin\mathcal{N}^{In}$), and subsequently update their weights in each iteration by
\begin{equation}\label{GNC7}
\omega_{i}^{t+1}=e^{-\frac{{r^t_{i}}^2}{{\mu^t}^2\bar{r}^2}},\, i\in\mathcal{N}^{In},
\end{equation}
rather than by \eqref{GNC6}, then the resulting heuristic, which is the ultimate version of RT-GNC, will become robust against over 99\% outliers and only require 10-14 iterations before the global minimum (or residual errors) converges (Section \ref{Subsection-subproblems}). Its pseudocode is given in Algorithm \ref{Algo2-RT-GNC}, in which the condition \textit{Converge} means: $|f^t-f^{t-1}|/|f^t|\leq10^{-5}$, where $f^t$ is the global minimum of \eqref{Go-4} in the {$t_{th}$} iteration.

\begin{algorithm}
\caption{RT-GNC}
\label{Algo2-RT-GNC}
\SetKwInOut{Input}{\textbf{Input}}
\SetKwInOut{Output}{\textbf{Output}}
\Input{correspondence set $\mathcal{N}^*=\{1,2,\dots,N\}$; inlier set $\mathcal{N}^{In}$; estimated scale $\mathit{\hat{s}}$; updating parameters $\nu$, $\eta$\;}
\Output{optimal rotation and translation: $\boldsymbol{\hat{R}}$ and $\boldsymbol{\hat{t}}$\;}
\BlankLine
Set big weights $\omega_i^{(1-3)}=big\_num$ (e.g. 200) where $i\in\mathcal{N}^{In}$ for the known inliers\;
Set initial weights $\omega_i^1=1$ where $i\notin\mathcal{N}^{In}$, and set $\mu^1=10$, $\xi^{1}=\mathit{\hat{s}}^2\cdot\max( {r^1_{i}}^2 )$, $max\_it=15$,  $t=1$\;
\While {$t\leq max\_it$}{
Solve minimization \eqref{Go-4} with moment SDP and recover optimal $\boldsymbol{\hat{R}}^t$ and $\boldsymbol{\hat{t}}^t$ using \eqref{t-free} with $\boldsymbol{\hat{q}}^t$\;
{\If {Converge}{\textbf{break}}}
Compute all residual errors $r_i^t$ where $i\in\mathcal{N}^*$ \;
Update weights $\omega_i^{t+1}$: if $i\notin\mathcal{N}^{In}$, then update with \eqref{GNC6}; if $i\in\mathcal{N}^{In}$, then update with \eqref{GNC7}\;
Update parameters: $\mu^{t+1}=\mu^t/\eta$ and \eqref{trimming-update}\;
$t=t+1$\;}
\Return optimal $\boldsymbol{\hat{R}}^{t-1}$ and $\boldsymbol{\hat{t}}^{t-1}$\;
\end{algorithm}

\section{Experiments}\label{Experiments}

We evaluate IRON, as well as its solutions for the subproblems: RANSIC, SOS Relaxation and RT-GNC, against the state-of-the-art solvers in multiple registration experiments, which are all implemented in Matlab on a laptop with a Core i7-7700HQ CPU and 16GB of RAM without using any parallelism programming technique.

\subsection{Testing of Subproblems}\label{Subsection-subproblems}

\textbf{Experimental Setup.} In our experiments, we adopt the Stanford 'bunny' point cloud~\cite{curless1996volumetric}. We first randomly downsample the bunny to 1000 points and resize it to fit in a $[-0.5,0.5]^3$ cube, as our first point set $\{\mathbf{P}_i\}_{i=1}^{1000}$. Then we generate a random transformation: $\mathit{s}, \boldsymbol{R},\boldsymbol{t}$, where $\mathit{s}\in[1,5]$, $\boldsymbol{R}\in SO3$ and $||\boldsymbol{t}||\leq \frac{\sqrt{3}}{2}$, to transform the first point set. We add random Gaussian noise with $\sigma=0.01$ and 0-mean to the transformed point set to have our second point set $\{\mathbf{Q}_i\}_{i=1}^{1000}$. To create outliers in clutter, a portion of the points in $\{\mathbf{Q}_i\}_{i=1}^{1000}$ are replaced by random points inside a sphere of diameter $\mathit{s}\sqrt{3}$ according to the outlier ratio (0\%-99\%), as displayed in Figure~\ref{OB}(a). All the experimental results are obtained over 50 Monte Carlo runs.

\begin{figure}[t]
\centering
\subfigure[Scale Estimation]{
\begin{minipage}[t]{1.00\linewidth}
\centering
\includegraphics[width=0.495\linewidth]{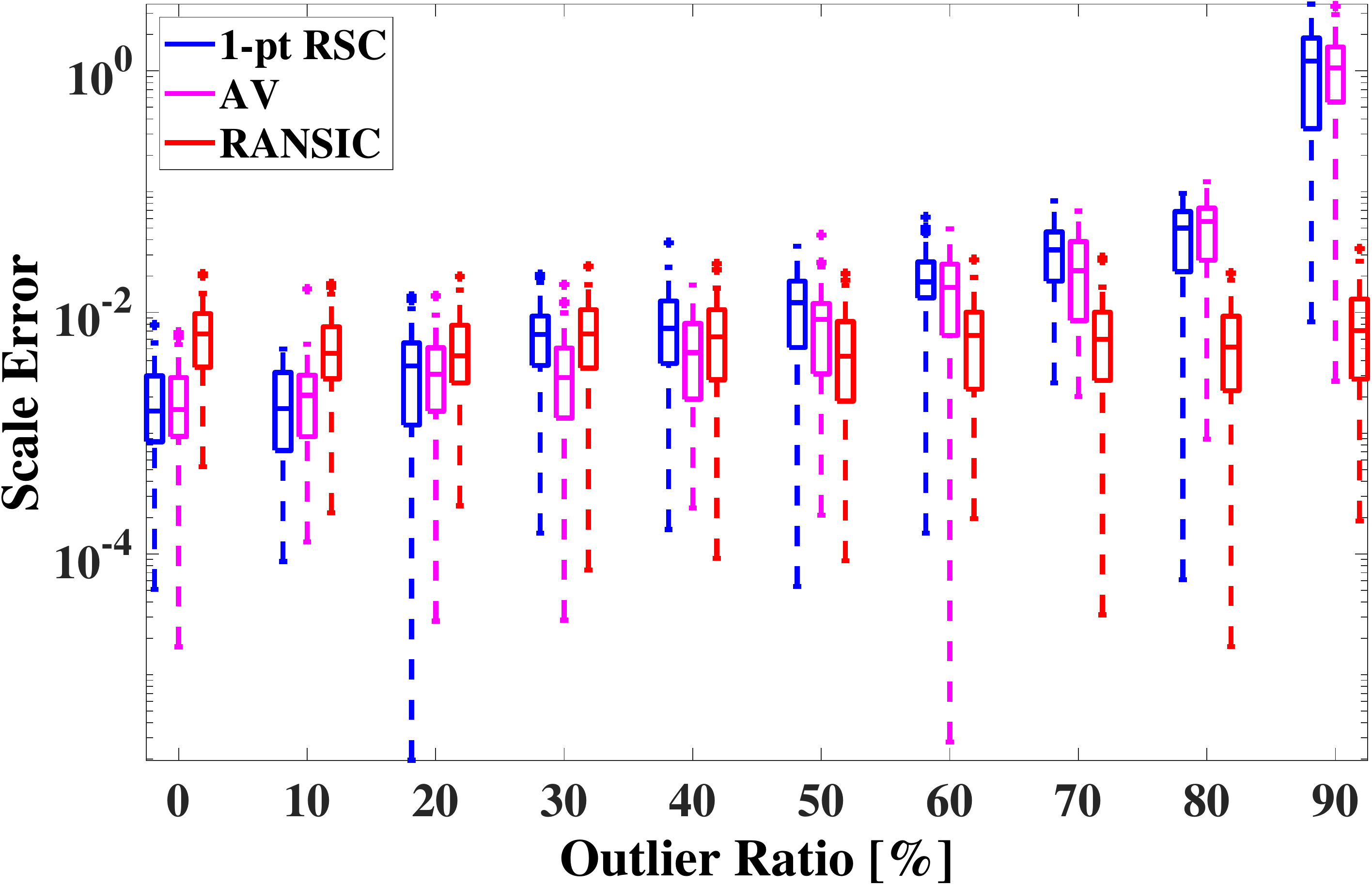}
\includegraphics[width=0.495\linewidth]{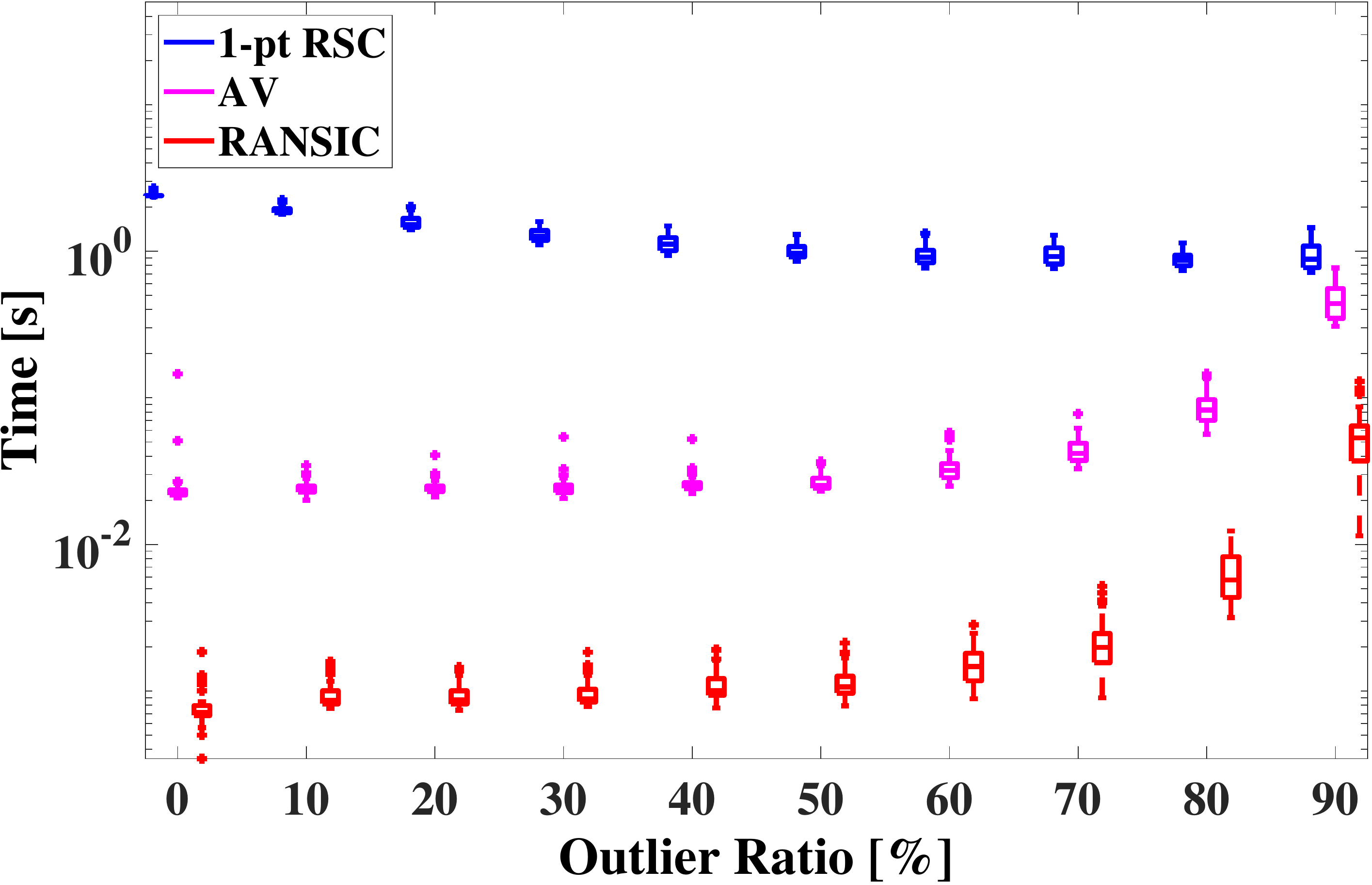}
\end{minipage}
}%
\vspace{-2mm}

\subfigure[Naive Registration]{
\begin{minipage}[t]{1.00\linewidth}
\centering
\includegraphics[width=0.495\linewidth]{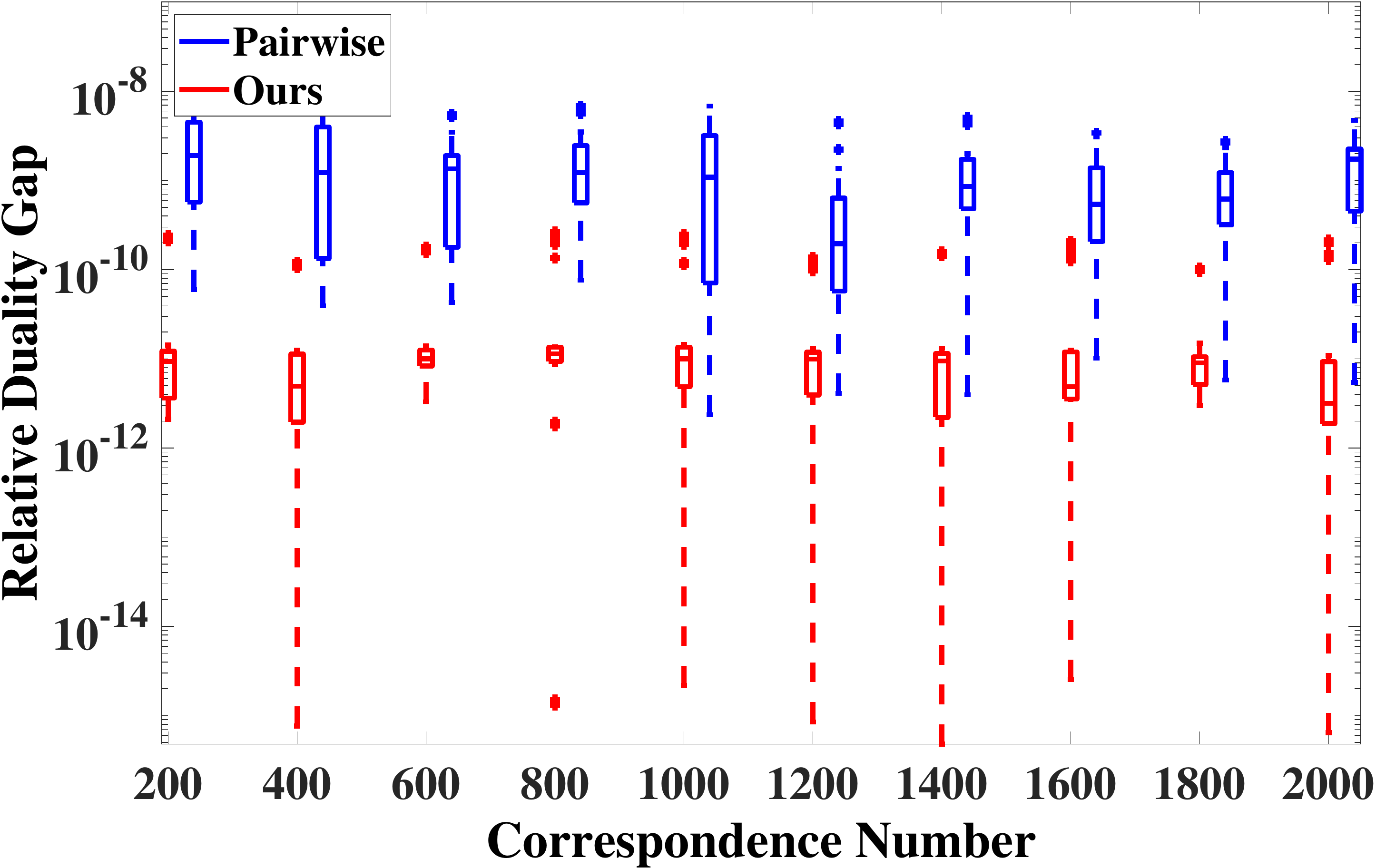}
\includegraphics[width=0.495\linewidth]{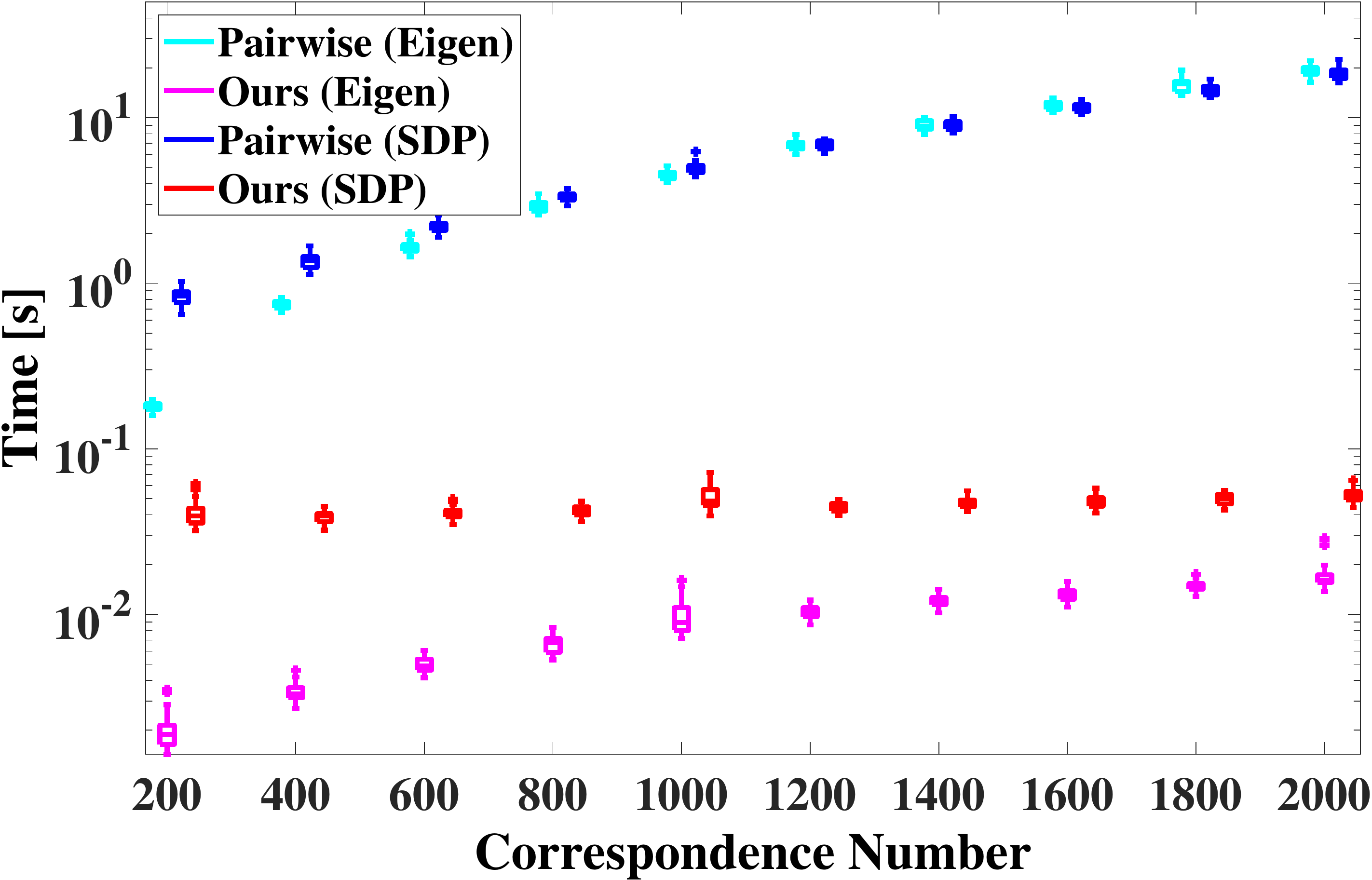}
\end{minipage}
}%
\vspace{-2mm}

\subfigure[Outlier Rejection]{
\begin{minipage}[t]{1.00\linewidth}
\centering
\includegraphics[width=0.495\linewidth]{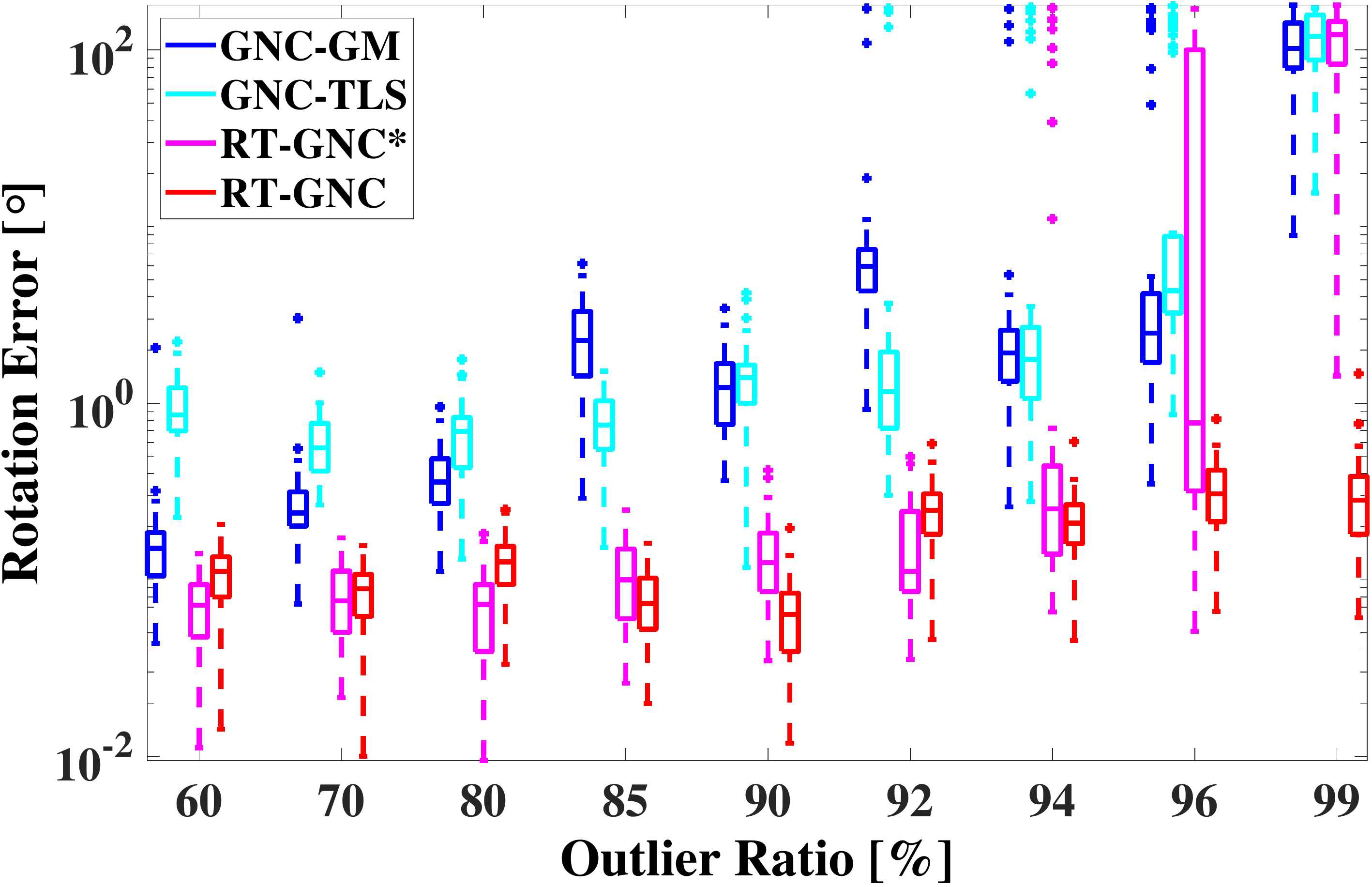}
\includegraphics[width=0.495\linewidth]{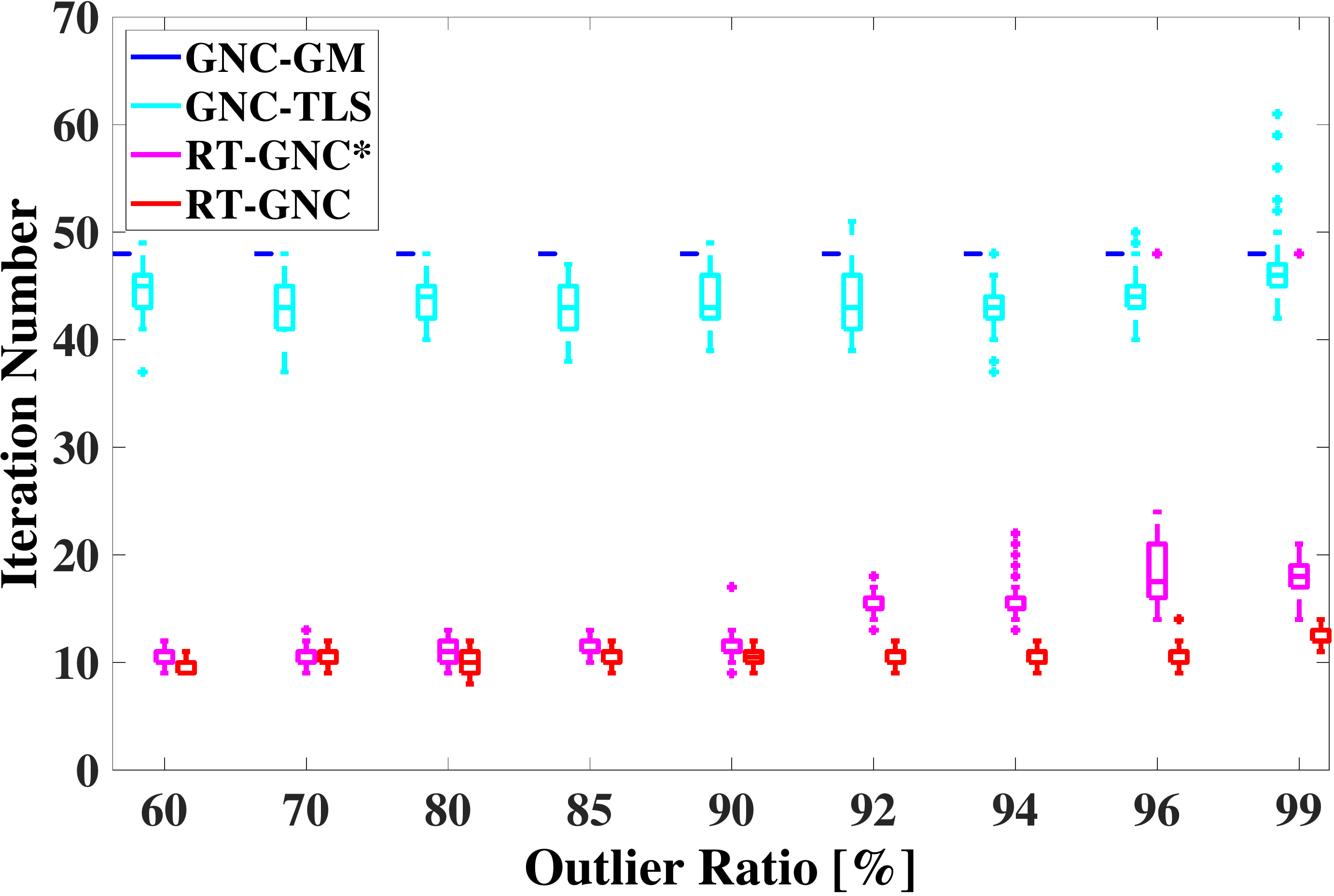}
\end{minipage}
}%
\vspace{-1mm}
\centering
\caption{Testing subproblems in the registration problem. \textbf{(a)} Accuracy and runtime of 1-pt RSC, AV and RANSIC. \textbf{(b)} Tightness and runtime of TEASER++'s pairwise solver and our solver (we adopt two versions for runtime). \textbf{(c)} Accuracy and convergence speed of two traditional GNC heuristics, RT-GNC and RT-GNC*.}
\label{Subproblems}
\end{figure}

\textbf{Scale Estimation.} We first benchmark RANSIC against Adaptive Voting (AV) in TEASER/TEASER++ and one-point RANSAC (1-pt RSC) in \cite{li2021point}. We downsample the correspondences to $N=100$ since AV will run in hours if with 1000 correspondences. In each run, scale $\mathit{s}$ is randomly chosen within $[1,5]$. Figure~\ref{Subproblems}(a) shows the scale errors: $|\mathit{\hat{s}}-\mathit{s}|$ and runtime w.r.t. increasing outlier ratios. We can observe that both AV and 1-pt RSC fail at 90\% outliers while RANSIC is robust against 90\% outliers and faster. In fact, RANSIC is precise even with 99\% outliers (Figure~\ref{OB}).

\textbf{Naive Registration.} We evaluate the global optimality (tightness) of the relaxation, defined as the relative duality gap: $(\hat{f}-f^{*})/\hat{f}$ ($\hat{f}$ is the objective recalculated with the optimal solutions). We compare our solver (using gloptipoly3+SeDuMi) with the pairwise-invariant rotation solver in TEASER++ (using CVX~\cite{grant2014cvx}+SeDuMi), both in their SDP version. Also, we introduce their eigen-based linear versions in time evaluation. As shown in Figure~\ref{Subproblems}(b), we test the tightness and runtime w.r.t. 200 to 2000 correspondences. We can see that our solver is tighter and is almost independent of the correspondence (since its modeling process is fast) while the runtime of TEASER++'s pairwise solver increases greatly with the correspondence number.

\textbf{Outlier Rejection.} We test our heuristic RT-GNC (Algorithm \ref{Algo2-RT-GNC}) and its RANSIC-free version RT-GNC* (Proposition \ref{Prop8}) against traditional GNC solvers, including GNC-TLS and GNC-GM. Figure~\ref{Subproblems}(c) shows the rotation errors and iteration numbers w.r.t. increasing outlier ratios. We find that both RT-GNC and RT-GNC* are over 3 times as fast as and more robust than the two traditional GNC heuristics; particularly, RT-GNC can even tolerate 99\% outliers.

\subsection{Overall Benchmark}

We provide a benchmark over the 'bunny'. More experiments on dataset~\cite{curless1996volumetric} are in the supplementary material.

\begin{figure*}[t]
\centering

\subfigure[Registration with Known Scale ($\mathit{s}=1$)]{
\begin{minipage}[t]{0.33\linewidth}
\centering
\includegraphics[width=0.49\linewidth]{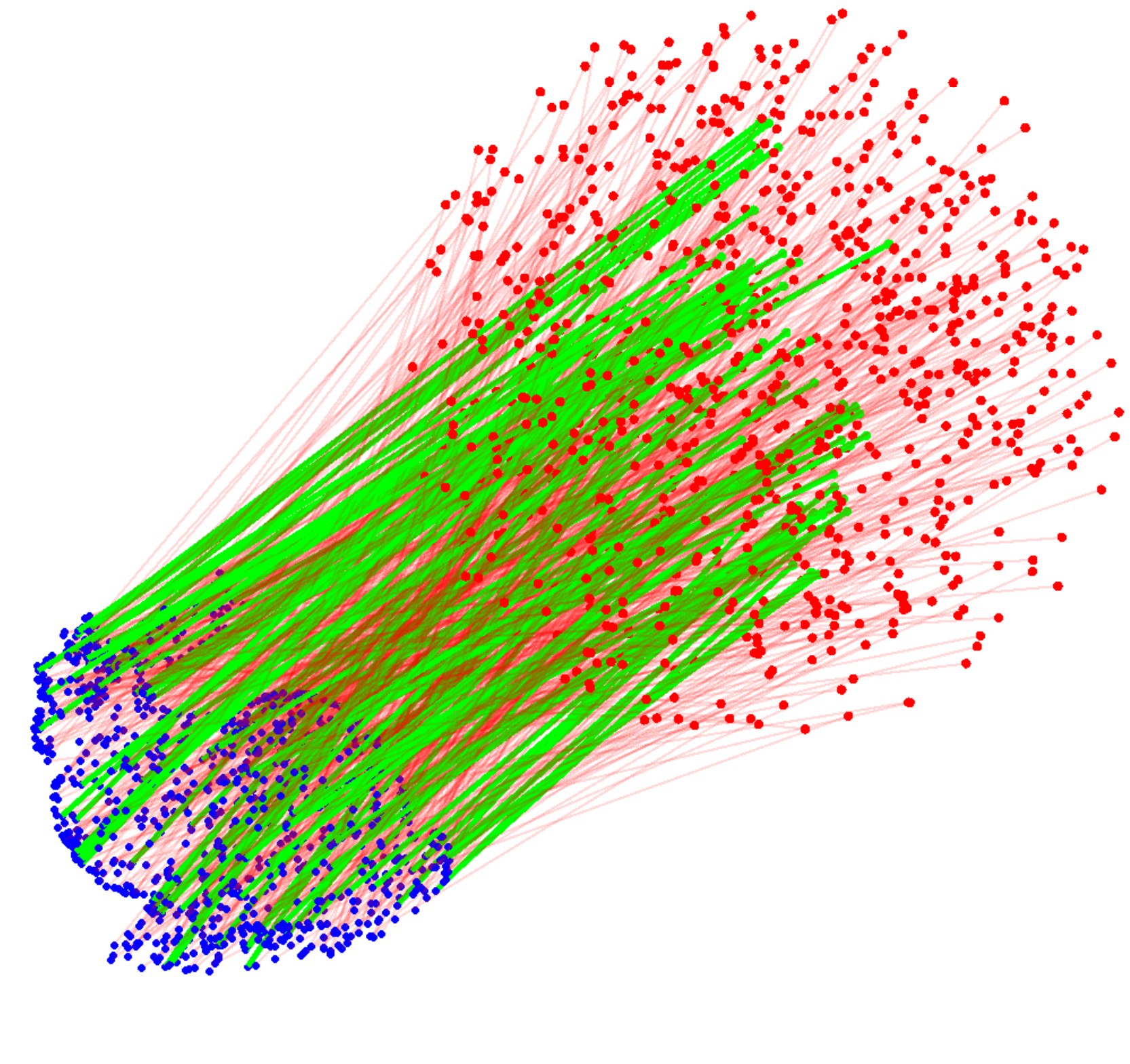}\hspace{-1mm}
\includegraphics[width=0.49\linewidth]{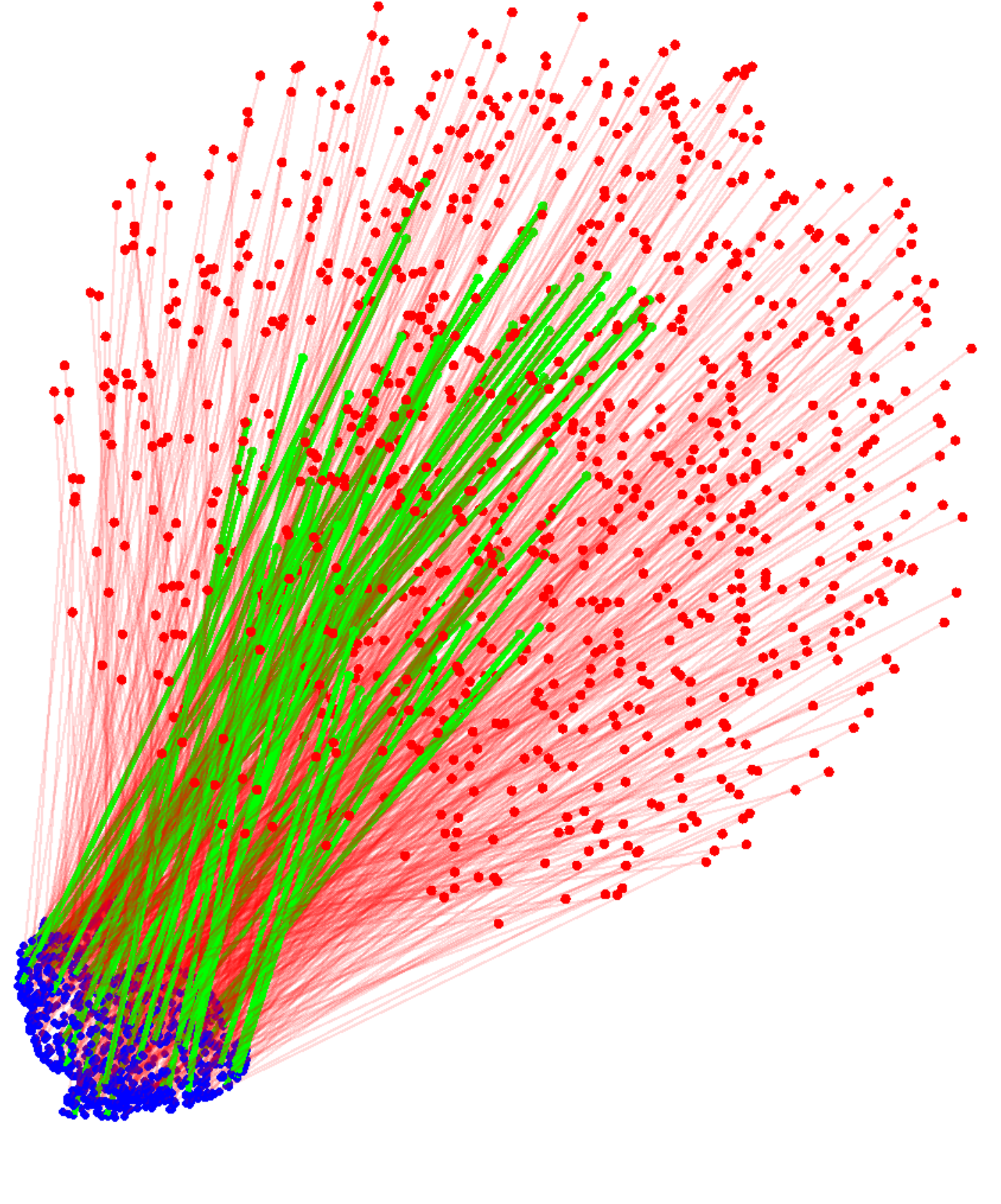}
\includegraphics[width=1\linewidth]{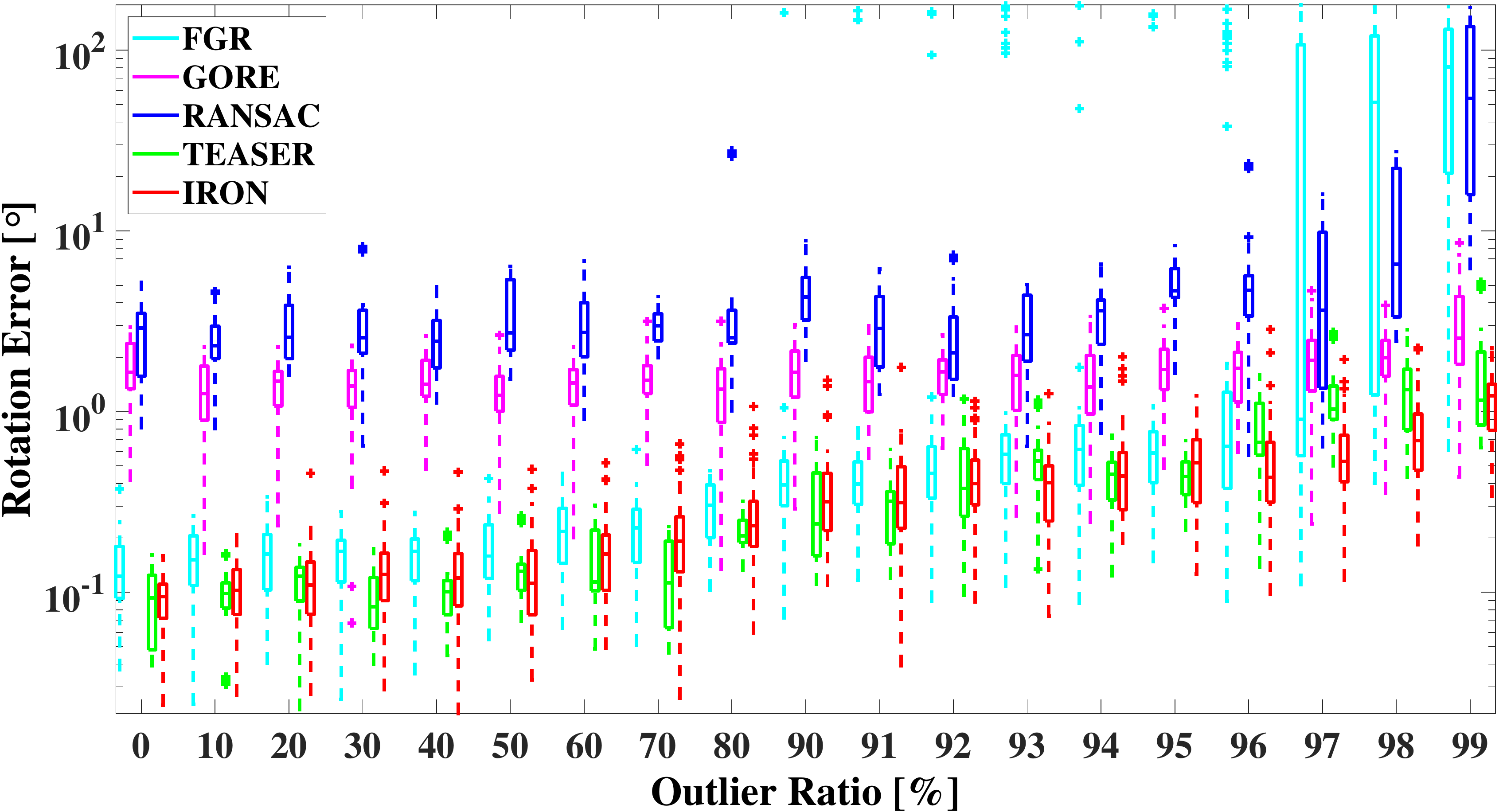}
\includegraphics[width=1\linewidth]{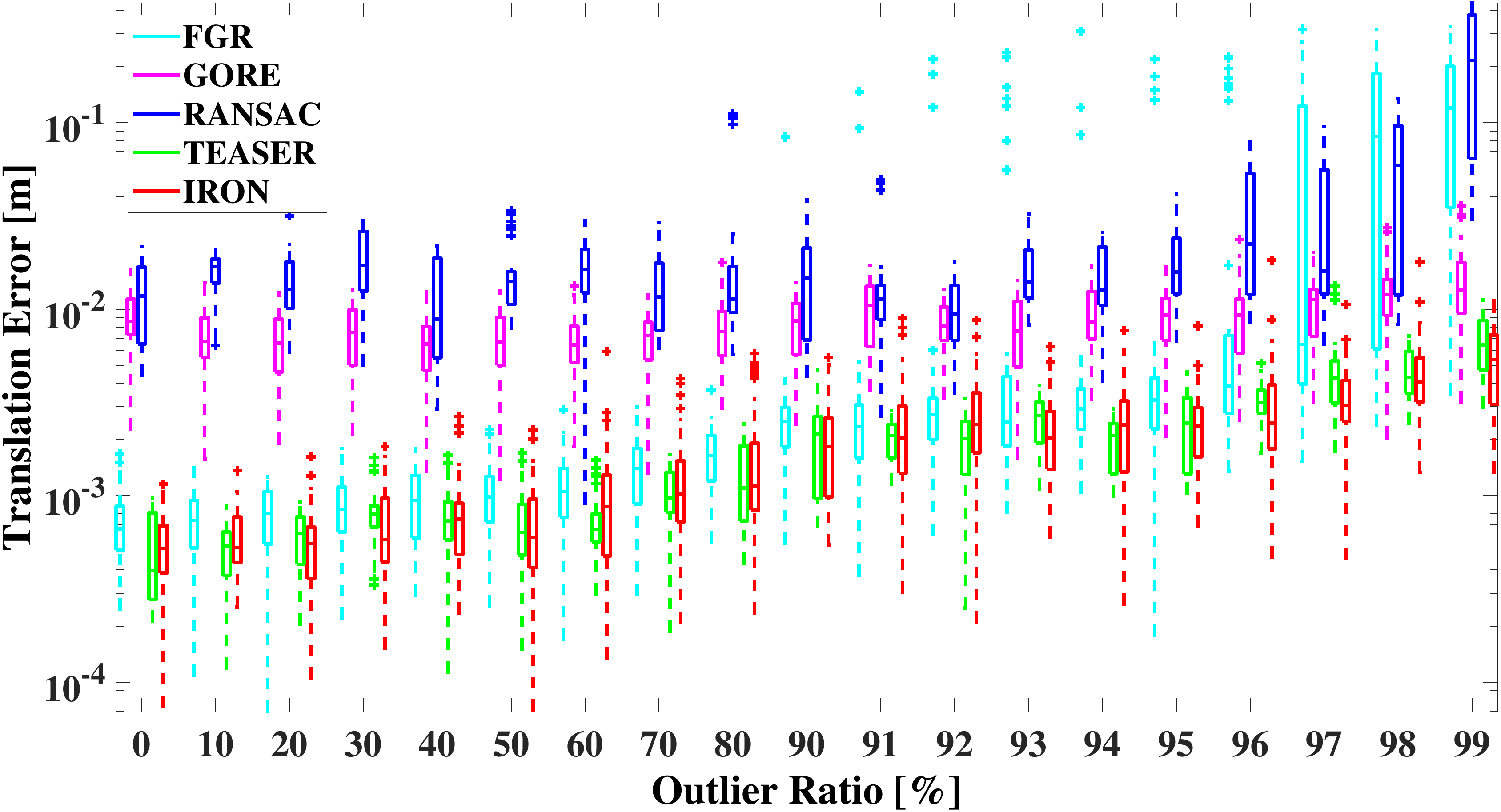}
\includegraphics[width=1\linewidth]{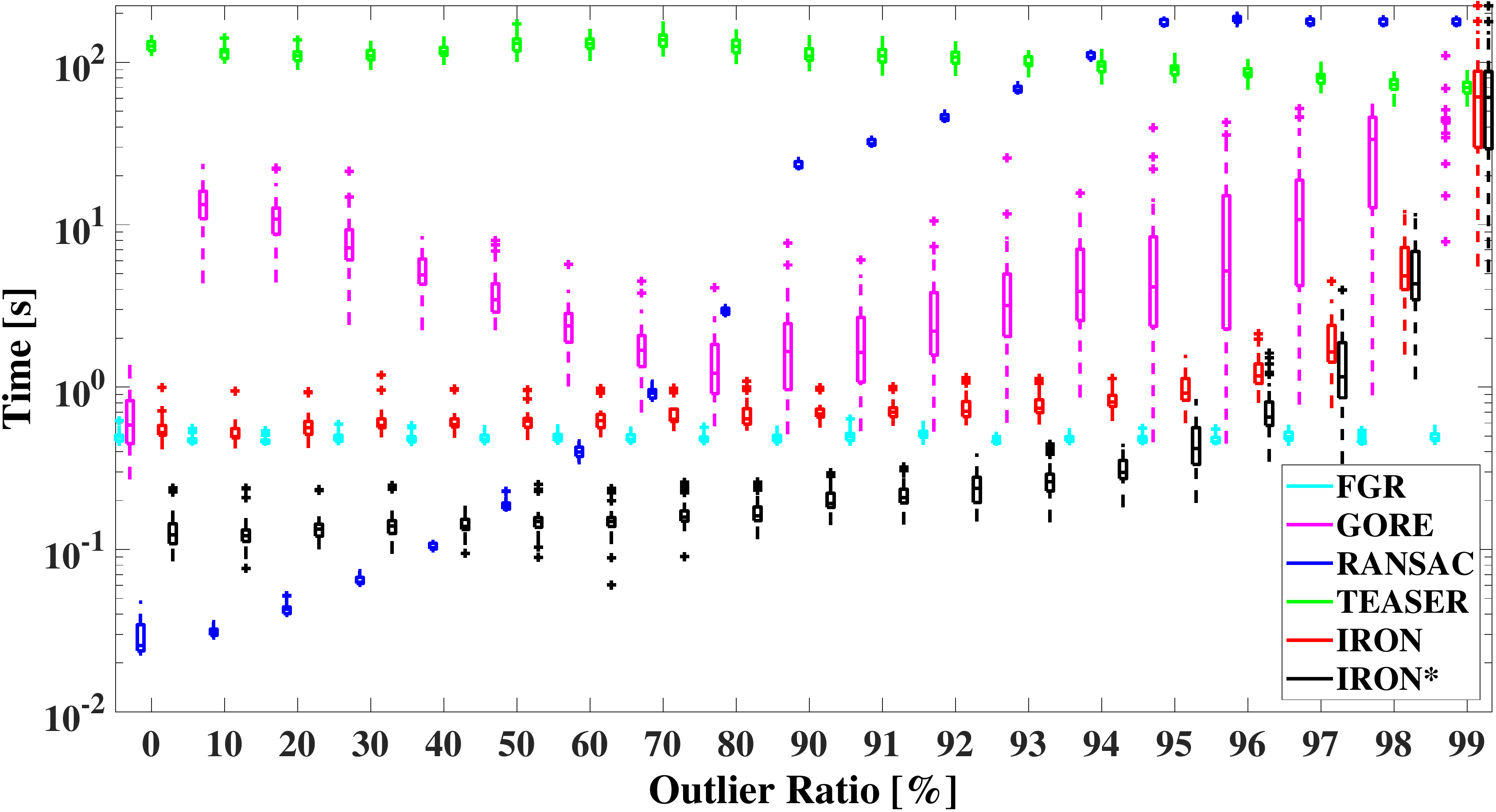}
\end{minipage}
}%
\subfigure[Registration with Unknown Scale]{
\begin{minipage}[t]{0.33\linewidth}
\centering
\includegraphics[width=1\linewidth]{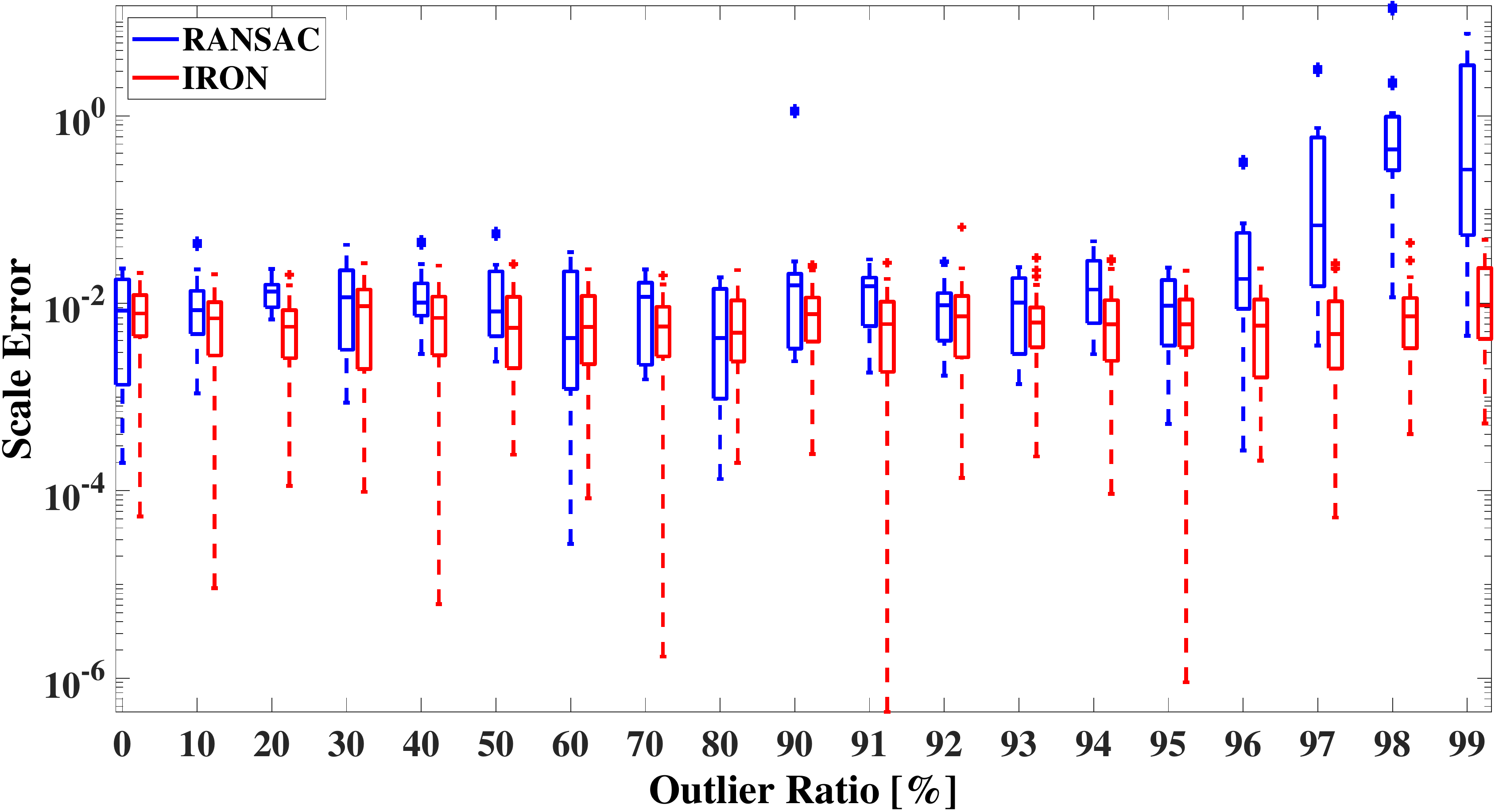}
\includegraphics[width=1\linewidth]{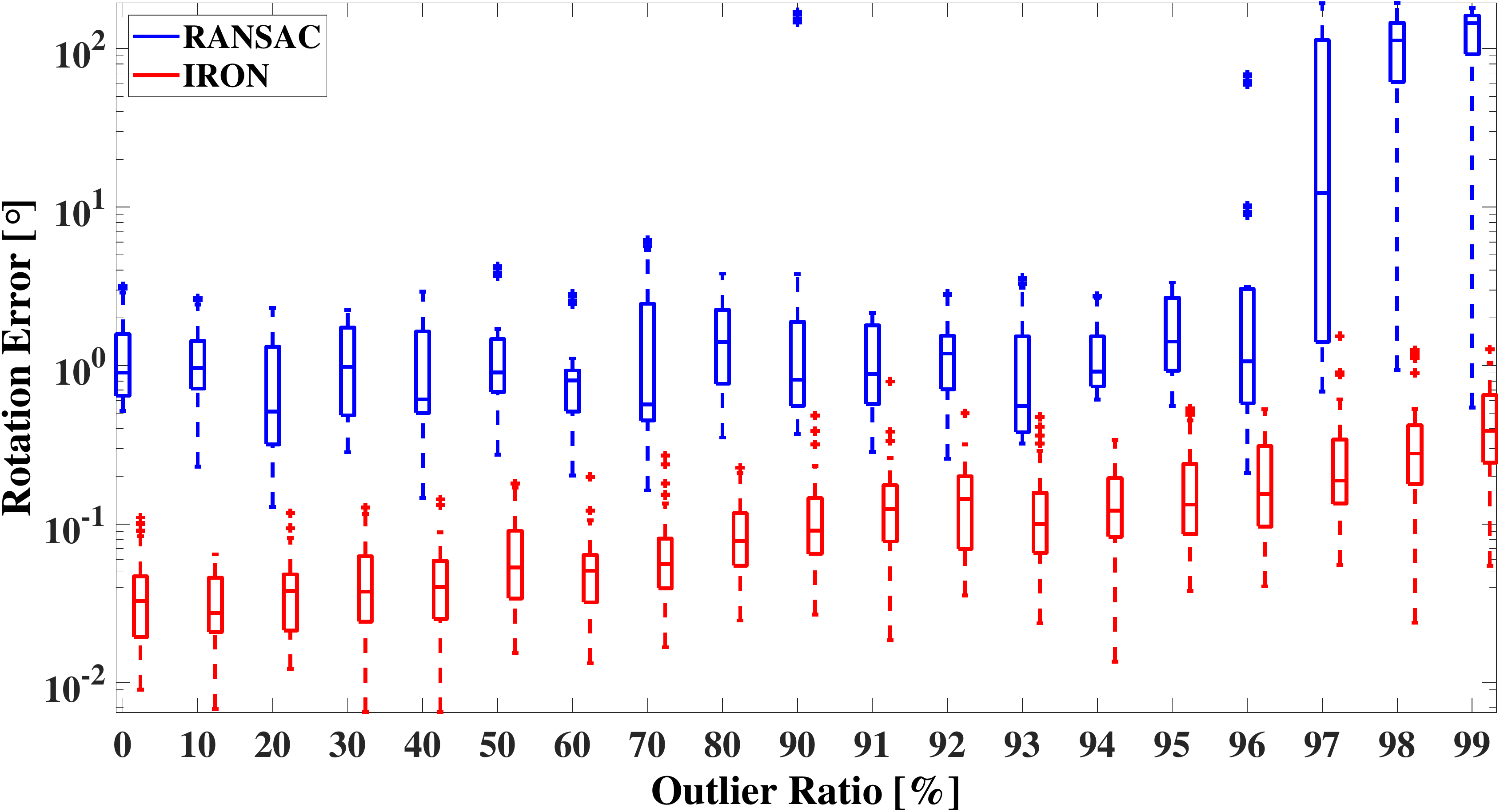}
\includegraphics[width=1\linewidth]{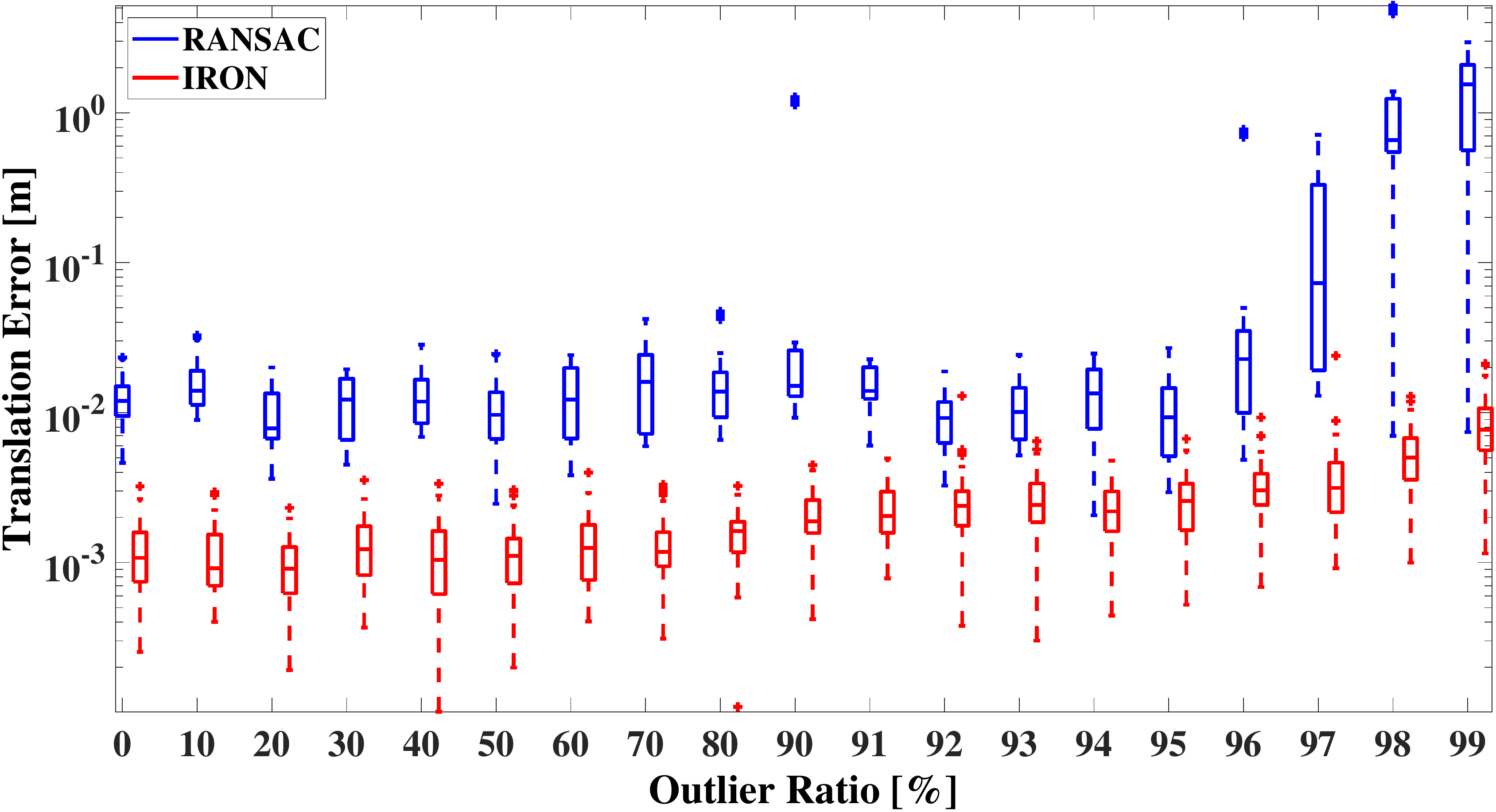}
\includegraphics[width=1\linewidth]{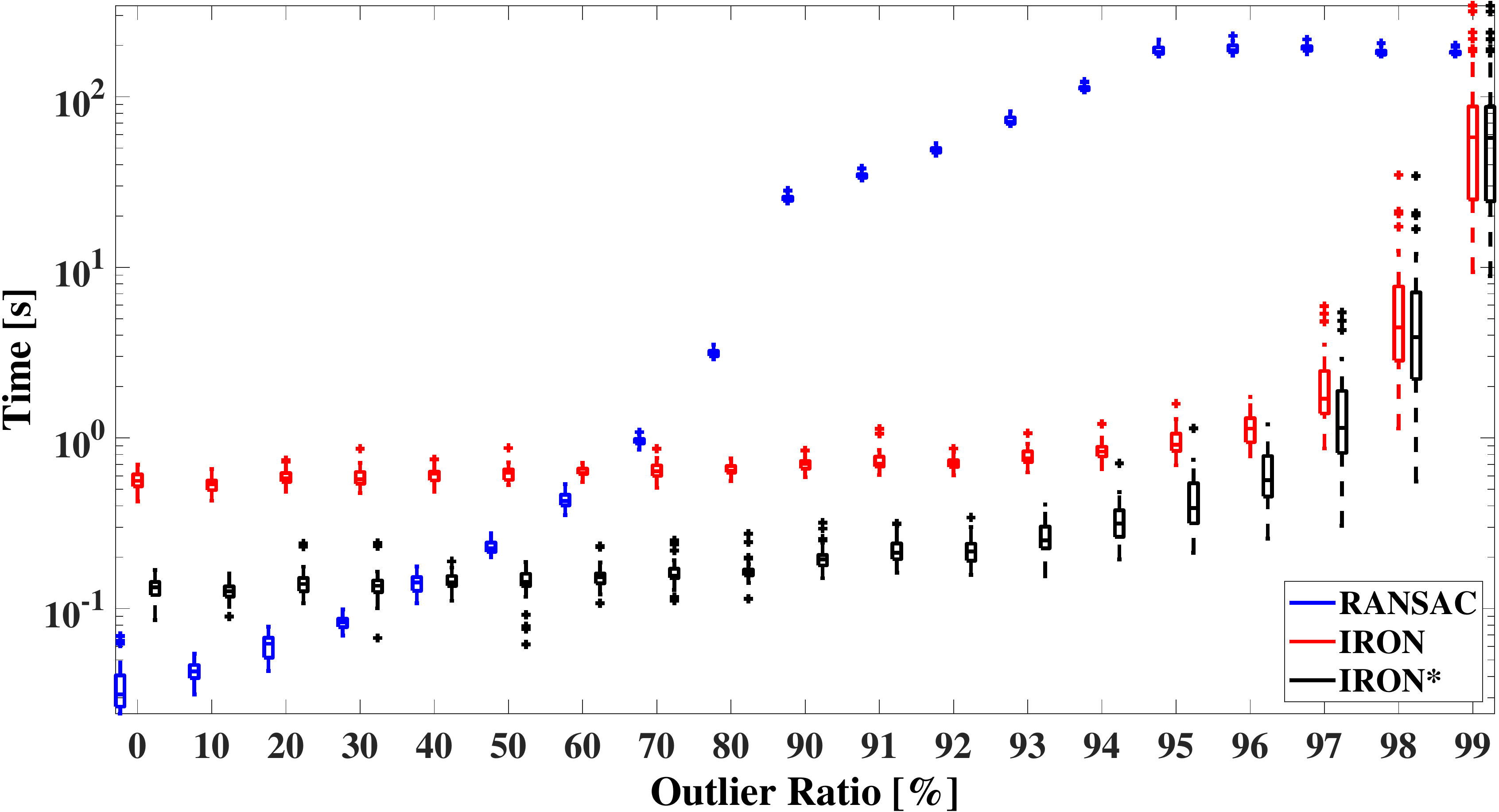}
\end{minipage}
}%
\subfigure{
\begin{minipage}[t]{0.33\linewidth}
\centering
\includegraphics[width=1\linewidth]{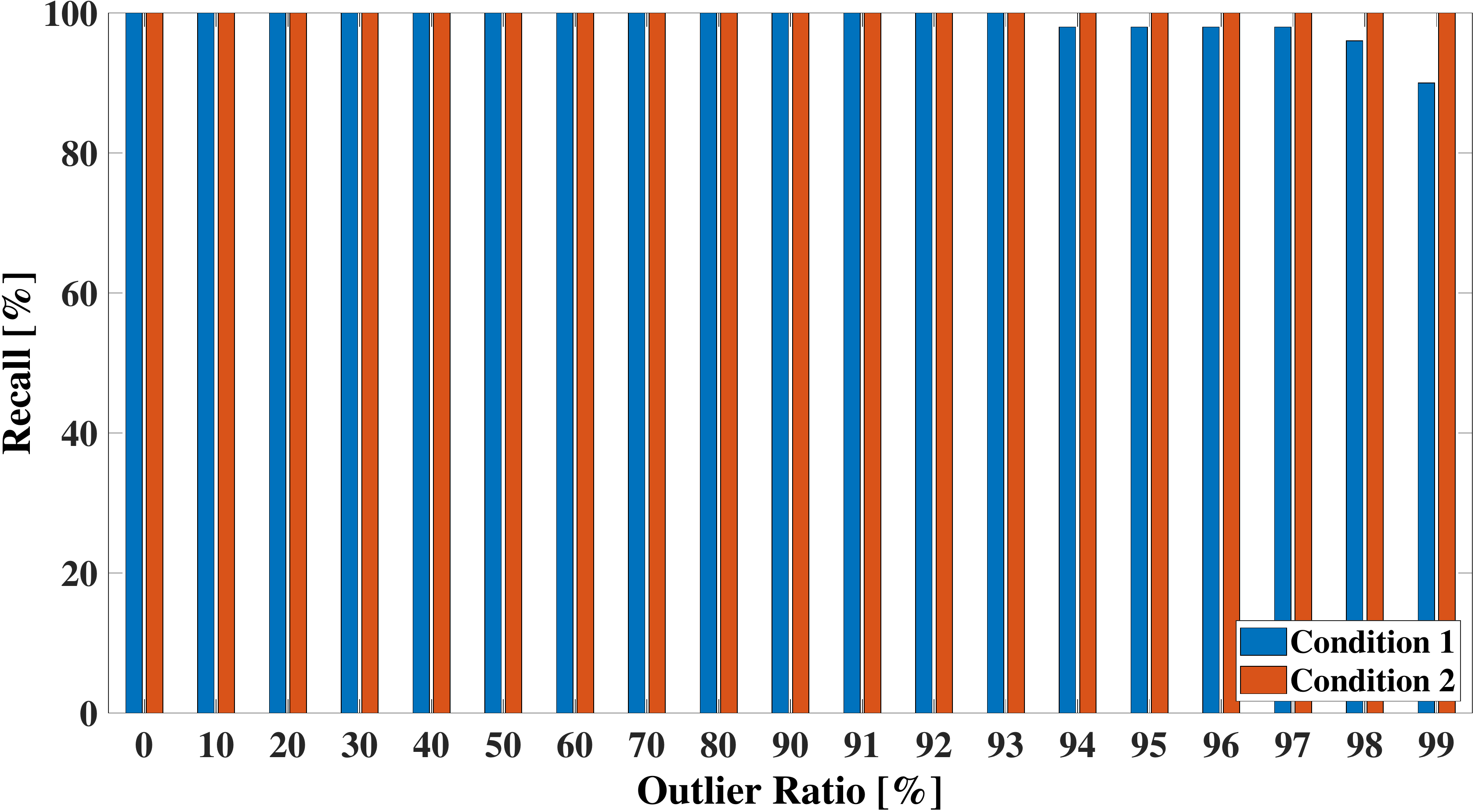}
\footnotesize{(c) Recall Ratio}\vspace{5mm}
\includegraphics[width=1\linewidth]{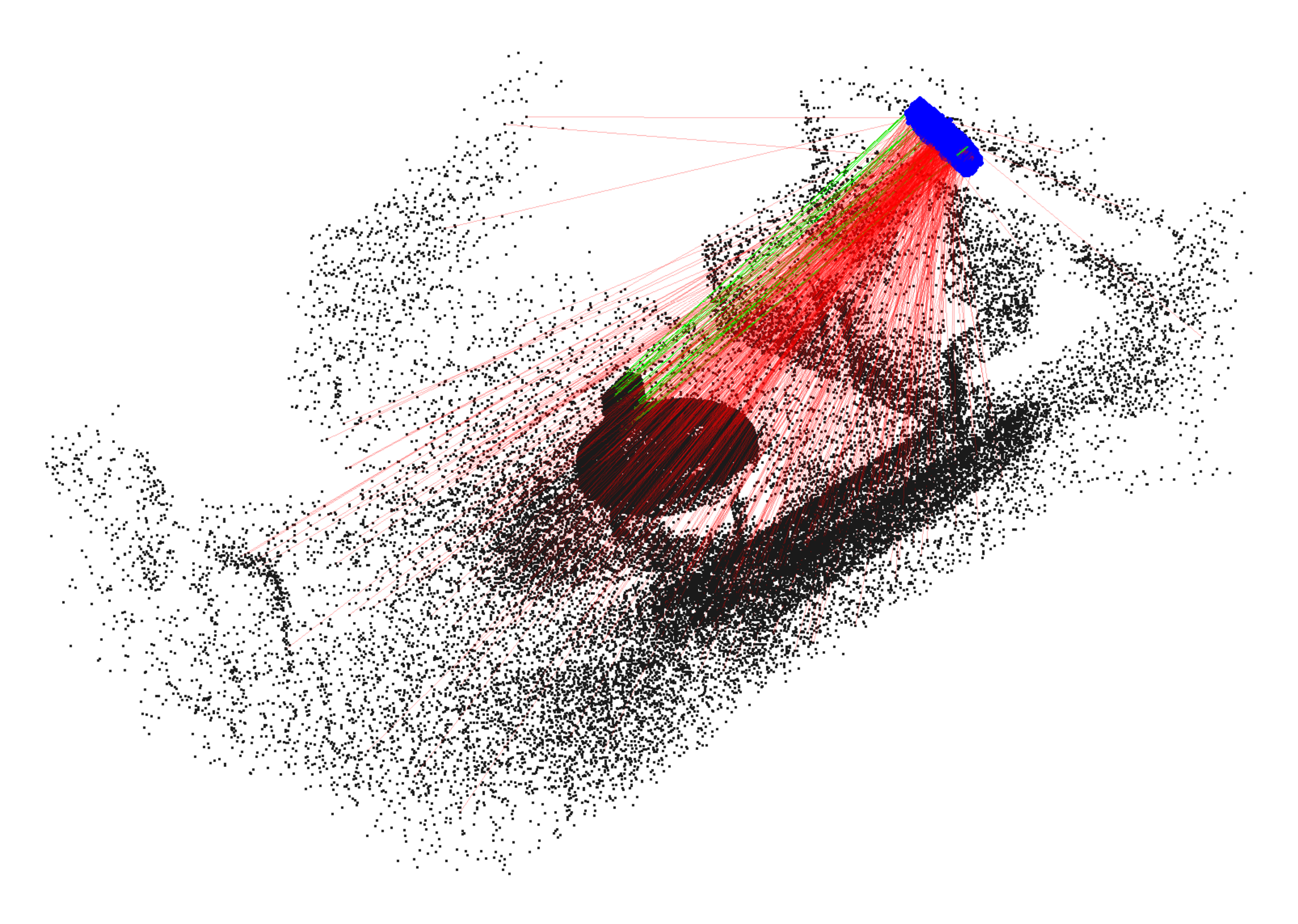}
\vspace{4.0mm}
\includegraphics[width=1\linewidth]{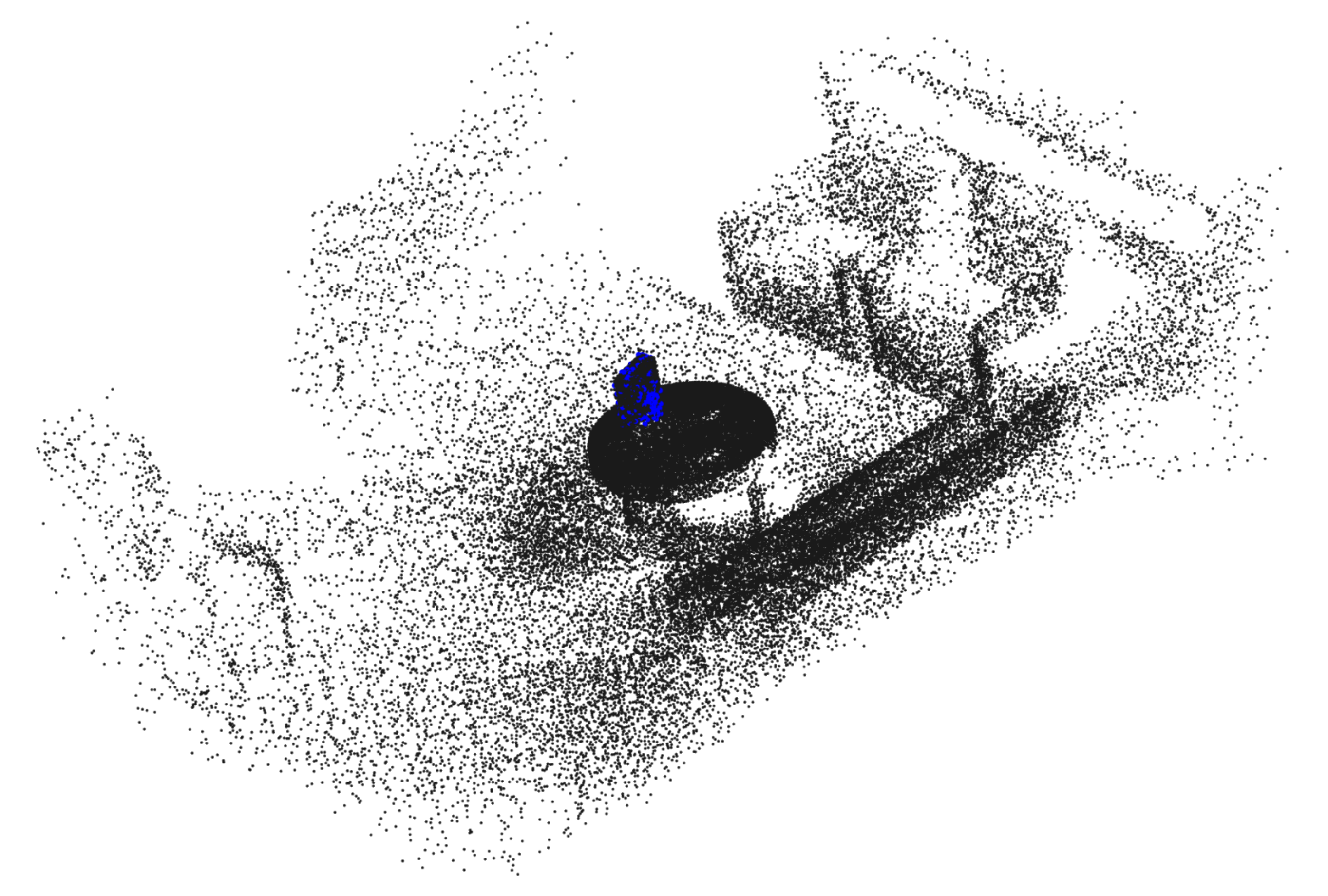}
\footnotesize{(d) Object Localization}
\end{minipage}
}%

\centering
\caption{Benchmark on standard registration problems. \textbf{(a)} Estimation accuracy and runtime of FGR, GORE, RANSAC, TEASER and IRON/IRON* with $\mathit{s}=1$. The top-row images demonstrate a known-scale registration problem with 90\% outliers and an unknown-scale registration problem with 95\% outliers. \textbf{(b)} Estimation accuracy and runtime of RANSAC and IRON/IRON*. \textbf{(c)} IRON's recall ratio of inliers. \textbf{(d)} Object localization with unknown-scale registration over dataset~\cite{lai2011large} (\textit{Scene-09}). In the first image, inliers are shown in green lines and outliers are in red lines. The reprojection result using the transformation estimated by IRON is displayed in the second image. Here we have 605 correspondences with 95.87\% outliers, and the errors of ($\mathit{s}, \boldsymbol{R}, \boldsymbol{t}$) are: ($2.83\times10^{-7}$, ${8.87\times10^{-6}}^{\circ}$, $3.98\times10^{-7} m$).}
\label{OB}
\vspace{-9pt}
\end{figure*}

\textbf{Known-scale registration:} We adopt FGR, GORE, RANSAC, TEASER and IRON. As for RANSAC, we use Horn's minimal (3-point) method \cite{horn1987closed} to solve the rigid transformation. When its iteration number exceeds $4/(1-\textit{outlier ratio})^3$ (where we empirically find RANSAC can converge) or 30000 (where the runtime is about 3 minutes), it will be stopped. In terms of TEASER, we adopt the GNC heuristic version of TEASER++ \cite{yang2020teaser}, but differently: (i) we do not use parallelism programming in order to ensure fairness for all solvers, and (ii) it is implemented in Matlab, not C++, using the maximum clique solver \cite{eppstein2010listing}. IRON is the complete algorithm in this article; IRON* denotes its fast eigen-based version, whose estimation errors are omitted as they are exactly the same as IRON's. From Figure~\ref{OB}(a), we observe that: (i) FGR yields wrong results with over 90\% outliers and RANSAC ($\leq30000$) begins to break at 96\%, and (ii) though GORE, TEASER and IRON are all robust against 99\% outliers, IRON is more precise than GORE and IRON/IRON* are faster than TEASER with 0-98\% outliers.

\textbf{Unknown-scale registration:} We only compare IRON with RANSAC because: (i) FGR and GORE are not designed for scale estimation, and (ii) scale estimator AV in TEASER cannot handle 1000 correspondence within an hour and is also unable to tolerate over 90\% outliers (Figure~\ref{Subproblems}(a)), thus no need for comparison. Figure~\ref{OB}(b) shows that: (i) IRON is accurate and robust even with 99\% outliers while RANSAC ($\leq30000$) fails at 96\%, and (ii) RANSAC is slower than IRON* when the outlier ratio is over 50\%.

\textbf{Recall of Inliers:} We evaluate IRON's recall ratio of inliers in the unknown-scale registration problem under two recall conditions, as shown in Figure~\ref{OB}(c). For condition 1, we compute the ratio of inliers that have weights 1 to the total inlier number. For condition 2, we compute the ratio of outliers that have weights 0 to the total outlier number. We find that IRON can always reject all the outliers and recall 90\% of inliers even when the outlier ratio is as high as 99\%.

\subsection{Application: 3D Object Localization}

We evaluate IRON over the RGB-D scenes datasets~\cite{lai2011large} for 3D object localization. We extract the 3D point cloud of the targeted object (\textit{cereal box} in this case) from the scene according to the labels provided and impose a random rigid transformation on the object. We then adopt the FPFH feature descriptor~\cite{rusu2009fast} to build correspondences between the scene and the transformed object. The correspondences and the estimation result by IRON of an unknown-scale registration problem are demonstrated in Figure~\ref{OB}(d). Experiments over more scenes are shown in the supplementary material.

\section{Conclusion}

We present a highly robust solver for correspondence-based point cloud registration. We propose RANSIC for scale estimation and inlier finding, apply SOS Relaxation for globally optimal robust registration, and design RT-GNC for outlier rejection. The resulting algorithm, IRON, has the most state-of-the-art robustness (robust against 99\% outliers) and very promising time-efficiency for both known-scale and unknown-scale registration problems.

{\small
\bibliographystyle{ieee_fullname}
\bibliography{egbib}

\begin{thebibliography}{10}\itemsep=-1pt

\bibitem{andrew2001multiple}
Alex~M Andrew.
\newblock Multiple view geometry in computer vision.
\newblock {\em Kybernetes}, 2001.

\bibitem{arun1987least}
K~Somani Arun, Thomas~S Huang, and Steven~D Blostein.
\newblock Least-squares fitting of two 3-d point sets.
\newblock {\em IEEE Transactions on pattern analysis and machine intelligence},
  (5):698--700, 1987.

\bibitem{audette2000algorithmic}
Michel~A Audette, Frank~P Ferrie, and Terry~M Peters.
\newblock An algorithmic overview of surface registration techniques for
  medical imaging.
\newblock {\em Medical image analysis}, 4(3):201--217, 2000.

\bibitem{barfoot2017state}
Timothy~D Barfoot.
\newblock {\em State estimation for robotics}.
\newblock Cambridge University Press, 2017.

\bibitem{besl1992method}
PJ Besl and Neil~D McKay.
\newblock A method for registration of 3-d shapes.
\newblock {\em IEEE Transactions on Pattern Analysis and Machine Intelligence},
  14(2):239--256, 1992.

\bibitem{black1996unification}
Michael~J Black and Anand Rangarajan.
\newblock On the unification of line processes, outlier rejection, and robust
  statistics with applications in early vision.
\newblock {\em International journal of computer vision}, 19(1):57--91, 1996.

\bibitem{blais1995registering}
G{\'e}rard Blais and Martin~D. Levine.
\newblock Registering multiview range data to create 3d computer objects.
\newblock {\em IEEE Transactions on Pattern Analysis and Machine Intelligence},
  17(8):820--824, 1995.

\bibitem{breckenridge1999quaternions}
WG Breckenridge.
\newblock Quaternions proposed standard conventions.
\newblock {\em Jet Propulsion Laboratory, Pasadena, CA, Interoffice Memorandum
  IOM}, pages 343--79, 1999.

\bibitem{briales2017convex}
Jesus Briales and Javier Gonzalez-Jimenez.
\newblock Convex global 3d registration with lagrangian duality.
\newblock In {\em Proceedings of the IEEE Conference on Computer Vision and
  Pattern Recognition}, pages 4960--4969, 2017.

\bibitem{bustos2017guaranteed}
Alvaro~Parra Bustos and Tat-Jun Chin.
\newblock Guaranteed outlier removal for point cloud registration with
  correspondences.
\newblock {\em IEEE transactions on pattern analysis and machine intelligence},
  40(12):2868--2882, 2017.

\bibitem{choi2015robust}
Sungjoon Choi, Qian-Yi Zhou, and Vladlen Koltun.
\newblock Robust reconstruction of indoor scenes.
\newblock In {\em Proceedings of the IEEE Conference on Computer Vision and
  Pattern Recognition}, pages 5556--5565, 2015.

\bibitem{curless1996volumetric}
Brian Curless and Marc Levoy.
\newblock A volumetric method for building complex models from range images.
\newblock In {\em Proceedings of the 23rd annual conference on Computer
  graphics and interactive techniques}, pages 303--312, 1996.

\bibitem{drost2010model}
Bertram Drost, Markus Ulrich, Nassir Navab, and Slobodan Ilic.
\newblock Model globally, match locally: Efficient and robust 3d object
  recognition.
\newblock In {\em 2010 IEEE computer society conference on computer vision and
  pattern recognition}, pages 998--1005. Ieee, 2010.

\bibitem{eppstein2010listing}
David Eppstein, Maarten L{\"o}ffler, and Darren Strash.
\newblock Listing all maximal cliques in sparse graphs in near-optimal time.
\newblock In {\em International Symposium on Algorithms and Computation}, pages
  403--414. Springer, 2010.

\bibitem{fischler1981random}
Martin~A Fischler and Robert~C Bolles.
\newblock Random sample consensus: a paradigm for model fitting with
  applications to image analysis and automated cartography.
\newblock {\em Communications of the ACM}, 24(6):381--395, 1981.

\bibitem{grant2014cvx}
Michael Grant and Stephen Boyd.
\newblock Cvx: Matlab software for disciplined convex programming, version 2.1,
  2014.

\bibitem{guo20143d}
Yulan Guo, Mohammed Bennamoun, Ferdous Sohel, Min Lu, and Jianwei Wan.
\newblock 3d object recognition in cluttered scenes with local surface
  features: A survey.
\newblock {\em IEEE Transactions on Pattern Analysis and Machine Intelligence},
  36(11):2270--2287, 2014.

\bibitem{hartley2013rotation}
Richard Hartley, Jochen Trumpf, Yuchao Dai, and Hongdong Li.
\newblock Rotation averaging.
\newblock {\em International journal of computer vision}, 103(3):267--305,
  2013.

\bibitem{henrion2009gloptipoly}
Didier Henrion, Jean-Bernard Lasserre, and Johan L{\"o}fberg.
\newblock Gloptipoly 3: moments, optimization and semidefinite programming.
\newblock {\em Optimization Methods \& Software}, 24(4-5):761--779, 2009.

\bibitem{henry2012rgb}
Peter Henry, Michael Krainin, Evan Herbst, Xiaofeng Ren, and Dieter Fox.
\newblock Rgb-d mapping: Using kinect-style depth cameras for dense 3d modeling
  of indoor environments.
\newblock {\em The International Journal of Robotics Research}, 31(5):647--663,
  2012.

\bibitem{horn1987closed}
Berthold~KP Horn.
\newblock Closed-form solution of absolute orientation using unit quaternions.
\newblock {\em Josa a}, 4(4):629--642, 1987.

\bibitem{lai2011large}
Kevin Lai, Liefeng Bo, Xiaofeng Ren, and Dieter Fox.
\newblock A large-scale hierarchical multi-view rgb-d object dataset.
\newblock In {\em 2011 IEEE international conference on robotics and
  automation}, pages 1817--1824. IEEE, 2011.

\bibitem{lasserre2001global}
Jean~B Lasserre.
\newblock Global optimization with polynomials and the problem of moments.
\newblock {\em SIAM Journal on optimization}, 11(3):796--817, 2001.

\bibitem{leclerc1989constructing}
Yvan~G Leclerc.
\newblock Constructing simple stable descriptions for image partitioning.
\newblock {\em International journal of computer vision}, 3(1):73--102, 1989.

\bibitem{li2021point}
Jiayuan Li, Qingwu Hu, and Mingyao Ai.
\newblock Point cloud registration based on one-point ransac and
  scale-annealing biweight estimation.
\newblock {\em IEEE Transactions on Geoscience and Remote Sensing}, 2021.

\bibitem{mosek2015mosek}
ApS Mosek.
\newblock The mosek optimization toolbox for matlab manual, 2015.

\bibitem{nie2014optimality}
Jiawang Nie.
\newblock Optimality conditions and finite convergence of lasserre's hierarchy.
\newblock {\em Mathematical programming}, 146(1):97--121, 2014.

\bibitem{olsson2008branch}
Carl Olsson, Fredrik Kahl, and Magnus Oskarsson.
\newblock Branch-and-bound methods for euclidean registration problems.
\newblock {\em IEEE Transactions on Pattern Analysis and Machine Intelligence},
  31(5):783--794, 2008.

\bibitem{papazov2012rigid}
Chavdar Papazov, Sami Haddadin, Sven Parusel, Kai Krieger, and Darius Burschka.
\newblock Rigid 3d geometry matching for grasping of known objects in cluttered
  scenes.
\newblock {\em The International Journal of Robotics Research}, 31(4):538--553,
  2012.

\bibitem{parra2014fast}
Alvaro Parra~Bustos, Tat-Jun Chin, and David Suter.
\newblock Fast rotation search with stereographic projections for 3d
  registration.
\newblock In {\em Proceedings of the IEEE conference on computer vision and
  pattern recognition}, pages 3930--3937, 2014.

\bibitem{putinar1993positive}
Mihai Putinar.
\newblock Positive polynomials on compact semi-algebraic sets.
\newblock {\em Indiana University Mathematics Journal}, 42(3):969--984, 1993.

\bibitem{rusu2009fast}
Radu~Bogdan Rusu, Nico Blodow, and Michael Beetz.
\newblock Fast point feature histograms (fpfh) for 3d registration.
\newblock In {\em 2009 IEEE international conference on robotics and
  automation}, pages 3212--3217. IEEE, 2009.

\bibitem{rusu2008aligning}
Radu~Bogdan Rusu, Nico Blodow, Zoltan~Csaba Marton, and Michael Beetz.
\newblock Aligning point cloud views using persistent feature histograms.
\newblock In {\em 2008 IEEE/RSJ international conference on intelligent robots
  and systems}, pages 3384--3391. IEEE, 2008.

\bibitem{shi2020robin}
Jingnan Shi, Heng Yang, and Luca Carlone.
\newblock Robin: a graph-theoretic approach to reject outliers in robust
  estimation using invariants.
\newblock {\em arXiv preprint arXiv:2011.03659}, 2020.

\bibitem{shuster1993survey}
Malcolm~D Shuster et~al.
\newblock A survey of attitude representations.
\newblock {\em Navigation}, 8(9):439--517, 1993.

\bibitem{sturm1999using}
Jos~F Sturm.
\newblock Using sedumi 1.02, a matlab toolbox for optimization over symmetric
  cones.
\newblock {\em Optimization methods and software}, 11(1-4):625--653, 1999.

\bibitem{tam2012registration}
Gary~KL Tam, Zhi-Quan Cheng, Yu-Kun Lai, Frank~C Langbein, Yonghuai Liu, David
  Marshall, Ralph~R Martin, Xian-Fang Sun, and Paul~L Rosin.
\newblock Registration of 3d point clouds and meshes: a survey from rigid to
  nonrigid.
\newblock {\em IEEE transactions on visualization and computer graphics},
  19(7):1199--1217, 2012.

\bibitem{tombari2013performance}
Federico Tombari, Samuele Salti, and Luigi Di~Stefano.
\newblock Performance evaluation of 3d keypoint detectors.
\newblock {\em International Journal of Computer Vision}, 102(1-3):198--220,
  2013.

\bibitem{tron2015inclusion}
Roberto Tron, David~M Rosen, and Luca Carlone.
\newblock On the inclusion of determinant constraints in lagrangian duality for
  3d slam.
\newblock In {\em Robotics: Science and Systems (RSS), Workshop "The problem of
  mobile sensors: Setting future goals and indicators of progress for SLAM},
  volume~4, 2015.

\bibitem{tzoumas2019outlier}
Vasileios Tzoumas, Pasquale Antonante, and Luca Carlone.
\newblock Outlier-robust spatial perception: Hardness, general-purpose
  algorithms, and guarantees.
\newblock In {\em 2019 IEEE/RSJ International Conference on Intelligent Robots
  and Systems (IROS)}, pages 5383--5390. IEEE, 2019.

\bibitem{yang2020graduated}
Heng Yang, Pasquale Antonante, Vasileios Tzoumas, and Luca Carlone.
\newblock Graduated non-convexity for robust spatial perception: From
  non-minimal solvers to global outlier rejection.
\newblock {\em IEEE Robotics and Automation Letters}, 5(2):1127--1134, 2020.

\bibitem{yang2019polynomial}
Heng Yang and Luca Carlone.
\newblock A polynomial-time solution for robust registration with extreme
  outlier rates.
\newblock In {\em Robotics: Science and Systems}, 2019.

\bibitem{yang2019quaternion}
Heng Yang and Luca Carlone.
\newblock A quaternion-based certifiably optimal solution to the wahba problem
  with outliers.
\newblock In {\em Proceedings of the IEEE/CVF International Conference on
  Computer Vision}, pages 1665--1674, 2019.

\bibitem{yang2020perfect}
Heng Yang and Luca Carlone.
\newblock In perfect shape: Certifiably optimal 3d shape reconstruction from 2d
  landmarks.
\newblock In {\em Proceedings of the IEEE/CVF Conference on Computer Vision and
  Pattern Recognition}, pages 621--630, 2020.

\bibitem{yang2020teaser}
Heng Yang, Jingnan Shi, and Luca Carlone.
\newblock Teaser: Fast and certifiable point cloud registration.
\newblock {\em IEEE Transactions on Robotics}, 2020.

\bibitem{zhang2014loam}
Ji Zhang and Sanjiv Singh.
\newblock Loam: Lidar odometry and mapping in real-time.
\newblock In {\em Robotics: Science and Systems}, volume~2, 2014.

\bibitem{zhou2016fast}
Qian-Yi Zhou, Jaesik Park, and Vladlen Koltun.
\newblock Fast global registration.
\newblock In {\em European Conference on Computer Vision}, pages 766--782.
  Springer, 2016.

\end{thebibliography}
}

\newpage

\section{Supplementary Material}

This supplementary material provides proof and derivations, a remark on RANSIC, the parameter setup for IRON, and some additional experiments.

\subsection{Proof and Derivations}

\begin{proof}[Proof of Closed-form Rotation Computation]
As discussed in~\cite{horn1987closed}, 3 point-to-point correspondences are sufficient to compute the rotation by creating a triad, based on which the rotation can be obtained linearly. We first obtain two pairs of unit vectors with 3 point correspondences such that
\begin{equation}\label{RANSIC-10}
\begin{gathered}
\mathbf{m}_1=\frac{\mathbf{P}_2-\mathbf{P}_1}{||\mathbf{P}_2-\mathbf{P}_1||}, \,\,
\mathbf{n}_1=\frac{\mathbf{Q}_2-\mathbf{Q}_1}{||\mathbf{Q}_2-\mathbf{Q}_1||}, \\
\mathbf{u}_1=\frac{\mathbf{P}_3-\mathbf{P}_1}{||\mathbf{P}_3-\mathbf{P}_1||}, \,\,
\mathbf{v}_1=\frac{\mathbf{Q}_3-\mathbf{Q}_1}{||\mathbf{Q}_3-\mathbf{Q}_1||},
\end{gathered}
\end{equation}
then we compute the normals of the planes defined by $\mathbf{m}_1$ and $\mathbf{u}_1$, and $\mathbf{n}_1$ and $\mathbf{v}_1$, respectively, such that
\begin{equation}\label{RANSIC-11}
\begin{gathered}
\mathbf{m}_2=\mathbf{m}_1 \times \mathbf{u}_1, \,\,\mathbf{n}_2=\mathbf{n}_1 \times \mathbf{v}_1,
\end{gathered}
\end{equation}
where $\mathbf{m}_1$ and $\mathbf{m}_2$, and $\mathbf{n}_1$ and $\mathbf{n}_2$ represent the X-axis and Y-axis in the respective coordinate systems. Lastly, we compute the corresponding Z-axis as
\begin{equation}\label{RANSIC-12}
\begin{gathered}
\mathbf{m}_3=\mathbf{m}_1 \times \mathbf{m}_2, \,\,\mathbf{n}_3=\mathbf{n}_1 \times \mathbf{n}_2,
\end{gathered}
\end{equation}
so that the rotation can be minimally achieved as
\begin{equation}\label{RANSIC-13}
\boldsymbol{\widetilde{R}}=\left[\begin{array}{ccc}
\mathbf{n}_1& \mathbf{n}_2& \mathbf{n}_3
\end{array}\right]\left[\begin{array}{ccc}
\mathbf{m}_1& \mathbf{m}_2& \mathbf{m}_3
\end{array}\right]^{\top}.
\end{equation}
\end{proof}

\begin{proof}[Proof of Theorem 1] First, it is easy to observe that the minimization problem: $\underset{\boldsymbol{x}\in\mathcal{K}}{\min} \boldsymbol{f}(\boldsymbol{x})$ is equivalent to the maximization problem: $\underset{\boldsymbol{x}\in\mathcal{K}}{\max \tau}, s.t.\, \boldsymbol{f}(\boldsymbol{x})-\tau\geq0$. 

Based on Putinar’s Positivstellensatz~\cite{putinar1993positive} and as stated in~\cite{yang2020perfect}, if $\boldsymbol{f}(\boldsymbol{x})-\tau\geq0$ over the feasible set $\mathcal{K}$ and $\langle \boldsymbol{h}\rangle_{2\theta}$ is Archimedean, then $\boldsymbol{f}(\boldsymbol{x})-\tau \in \phi_0(\boldsymbol{x}) + \langle \boldsymbol{h}\rangle_{2\theta}$. And according to Lasserre's Hierarchy~\cite{lasserre2001global}, the SOS Relaxation in Theorem 1 can be derived.

In addition, Nie~\cite{nie2014optimality} has proved that this hierarchy should have finite convergence under Archimedeanness for some finite order $\theta$, which depends on the degrees of ${f}(\boldsymbol{x})$ and the equality constraints ${h}_i(\boldsymbol{x})$ in set $\mathcal{K}$.
\end{proof}

\begin{proof}[Proof of Proposition 6] The proof is similar to that in~\cite{shi2020robin}. Following Proposition 6, given the two rotation-based invariant functions $\{\boldsymbol{r}_k(\mathbf{P}_i, \mathbf{Q}_i)\}_{k=1}^2$, we can have
\begin{equation}
\begin{gathered}
\boldsymbol{r}_1(\mathbf{P}_i, \mathbf{Q}_i)^{\top}\boldsymbol{r}_2(\mathbf{P}_i, \mathbf{Q}_i) \\
=\mathbf{Exp}(\boldsymbol{\eta}_2)^{\top}\boldsymbol{{R}}^{\top} \boldsymbol{{R}} \mathbf{Exp}(\boldsymbol{\eta}_1)\\
=\mathbf{Exp}(\boldsymbol{\eta}_2)^{\top} \mathbf{Exp}(\boldsymbol{\eta}_1).
\end{gathered}
\end{equation}
Then we introduce the geodesic error~\cite{hartley2013rotation} that can describe the geometric distance between two rotation matrices (e.g. $\boldsymbol{R}_1$ and $\boldsymbol{R}_2$) as
\begin{equation}
\boldsymbol{E}_{SO3}(\boldsymbol{R}_1, \boldsymbol{R}_2)=arccos \left(\frac{trace(\boldsymbol{R}_1^{\top}\boldsymbol{R}_2)-1}{2}\right).
\end{equation}
Thus, if we set an upper-bound to the geodesic error between $\mathbf{Exp}(\boldsymbol{\eta}_1)$ and $\mathbf{Exp}(\boldsymbol{\eta}_2)$ such that 
\begin{equation}
\boldsymbol{E}_{SO3}(\mathbf{Exp}(\boldsymbol{\eta}_1), \mathbf{Exp}(\boldsymbol{\eta}_2))\leq\gamma^{*},
\end{equation}
then it is apparent to have a corresponding compatibility condition w.r.t. $\{\boldsymbol{r}_k(\mathbf{P}_i, \mathbf{Q}_i)\}_{i=1}^2$ such that
\begin{equation}
\begin{gathered}
trace(\boldsymbol{r}_1(\mathbf{P}_i, \mathbf{Q}_i)^{\top}\boldsymbol{r}_2(\mathbf{P}_i, \mathbf{Q}_i))\\=trace(\mathbf{Exp}(\boldsymbol{\eta}_1)^{\top} \mathbf{Exp}(\boldsymbol{\eta}_2))\leq\gamma.
\end{gathered}
\end{equation}
The proof of the second part of Proposition 6 is omitted here, as it is easy to derive based on Proposition 2 and 4.
\end{proof}

\begin{proof}[Proof of Theorem 2] Following Problem 1, we endow each correspondence, say the $i_{th}$ correspondence, with a certain weight $\omega_i$ that should be within $(0,1]$ (typically 1) if it is an inlier and equal to 0 if it is an outlier, so that we can rewrite the objective function as the following weighted formulation:
\begin{equation}\label{obj}
\omega_i||\mathit{s}\boldsymbol{R}\mathbf{P}_i+\boldsymbol{t}-\mathbf{Q}_i||^2=\omega_i\left(\mathit{\hat{s}}^{2}\mathbf{\widetilde{P}}_i^{\top}\mathbf{\widetilde{P}}_i+\mathbf{\widetilde{Q}}_i^{\top}\mathbf{\widetilde{Q}}_i-2\mathit{\hat{s}}\mathbf{\widetilde{Q}}_i^{\top}\boldsymbol{R}\mathbf{\widetilde{P}}_i\right),
\end{equation}
and accordingly, by analogy, the centroids before and after transformation should be computed using the weights as 
\begin{equation}
\mathbf{\bar{P}}=\frac{\sum_{i=1}^N(\omega_i \mathbf{P}_i)}{\sum_{i=1}^N\omega_i},
\,\,\mathbf{\bar{Q}}=\frac{\sum_{i=1}^N(\omega_i \mathbf{Q}_i)}{\sum_{i=1}^N\omega_i}.
\end{equation}
\end{proof}

\begin{proof}[Proof of Proposition 7] Following the optimization problem given in Proposition 7, based on the property of unit quaternions \cite{breckenridge1999quaternions,shuster1993survey}, similar to the derivation of Problem 1 in \cite{yang2019quaternion}, we can derive that
\begin{equation}
\left[\begin{array}{c} \boldsymbol{R}\mathbf{\widetilde{P}}_i \\ 0\end{array}\right]=\mathbf{\Pi}_2(\boldsymbol{q})^{\top}\mathbf{\Pi}_1(\boldsymbol{q})\mathbf{\widetilde{P}}_i^*,
\end{equation}
where $\mathbf{\widetilde{P}}_i^*=[\mathbf{\widetilde{P}}_i^{\top} , 0]^{\top}$, $\boldsymbol{q}=[{q}_1,  {q}_2, {q}_3,  {q}_4 ]^{\top}$, and
\begin{equation}
\begin{gathered}
\mathbf{\Pi}_1(\boldsymbol{q})=\left[\begin{array}{cccc} q_4 & -q_3 & q_2 &q_1 \\ q_3 & q_4 & -q_1 &q_2 \\ -q_2 & q_1 & q_4 &q_3 \\ -q_1 & -q_2 & -q_3 &q_4 \end{array}\right], \\
\mathbf{\Pi}_2(\boldsymbol{q})=\left[\begin{array}{cccc} q_4 & q_3 & -q_2 &q_1 \\ -q_3 & q_4 & q_1 &q_2 \\ q_2 & -q_1 & q_4 &q_3 \\ -q_1 & -q_2 & -q_3 &q_4 \end{array}\right].
\end{gathered}
\end{equation}
Thus, letting $\mathbf{\widetilde{Q}}_i^{*}=[\mathbf{\widetilde{Q}}_i^{\top} , 0]^{\top}$, we can have
\begin{equation}
\begin{gathered}
\mathbf{\widetilde{Q}}_i^{\top}\boldsymbol{R}\mathbf{\widetilde{P}}_i=\mathbf{\widetilde{Q}}_i^{*\top}\left[\begin{array}{c} \boldsymbol{R}\mathbf{\widetilde{P}}_i \\ 0\end{array}\right]\\
={\mathbf{\widetilde{Q}}_i^{*\top}}\mathbf{\Pi}_2(\boldsymbol{q})^{\top}\mathbf{\Pi}_1(\boldsymbol{q})\mathbf{\widetilde{P}}_i^*\\
=\boldsymbol{q}^{\top} \underbrace{\mathbf{\Pi}_1(\mathbf{\widetilde{Q}}_i^*)^{\top}\mathbf{\Pi}_2(\mathbf{\widetilde{P}}_i^*) }_{\mathbf{C}^{\#}_i}\boldsymbol{q},
\end{gathered}
\end{equation}
Since the first two terms in \eqref{obj} are constants, minimizing objective \eqref{obj} is equivalent to minimizing 
\begin{equation}\label{QCQP-}
-\mathit{\hat{s}}\sum_{i=1}^N\omega_i\boldsymbol{q}^{\top}\mathbf{C}^{\#}_i\boldsymbol{q}=\boldsymbol{q}^{\top}\mathbf{C}\boldsymbol{q},
\end{equation}
where $\mathbf{C}=-\mathit{\hat{s}}\sum_{i=1}^N \omega_i \mathbf{\Pi}_1(\mathbf{\widetilde{P}}_i^*)^{\top}\mathbf{\Pi}_2(\mathbf{\widetilde{Q}}_i^*)$.
\end{proof}

\begin{proof}[Proof of Proposition 8]
In the point cloud registration problem, due to the existence of noise, the SOS objective function in Proposition 8 should be positive in practice. And apparently, $\langle \boldsymbol{h}\rangle_{2 \theta}$ is Archimedean with $\theta=1$, since $1-(q_1^2+q_2^2+q_3^2+q_4^2) = h(\boldsymbol{q})$ where $h(\boldsymbol{q})=1-\boldsymbol{q}^{\top}\boldsymbol{q}$. 

According to Putinar's Positivstellensatz~\cite{putinar1993positive}, we can derive the following condition: $(\boldsymbol{f}(\boldsymbol{q})+\mathit{D})-(\tau+\mathit{D}) \in \phi_0(\boldsymbol{q})+\langle \boldsymbol{h}\rangle_{2 \theta}$ where $\mathit{D}$ denotes the sum of all the constant terms in the objective and  $\phi_0(\boldsymbol{q})$ has degree no more than 2, so $\boldsymbol{f}(\boldsymbol{q})-\tau =\phi_0(\boldsymbol{q})+g(\boldsymbol{q})$ where $g(\boldsymbol{q})=\zeta \cdot h(\boldsymbol{q}) \in \langle \boldsymbol{h}\rangle_{2 \theta}$ is always satisfied. Since the registration problem is essentially equivalent to minimizing the QCQP \eqref{QCQP-}, the relaxation order $\theta$ should be 1. Therefore, the robust registration problem can be certifiably converted into the SOS program as written in Proposition 8.
\end{proof}

\begin{proof}[Proof of Lemma 1]
The general outlier process of the Leclerc cost function has been given in \cite{leclerc1989constructing,black1996unification} such that
\begin{equation}\label{Lclc}
\boldsymbol{\Psi}(z)=z\ln(z)-z+1.
\end{equation}
Now let us design a new function based on \eqref{Lclc} by replacing $z$ with the weight $\omega_i^t$ in the $t_{th}$ iteration and adopting the controlling parameter $\mu$ and the threshold $\bar{r}$ as variables, leading to a new outlier process for GNC:
\begin{equation}\label{GNC2}
\boldsymbol{\Psi}(\omega_i^{t},\mu^{t},\bar{r})=\mu^2\bar{r}^2\left(\omega_i^t\ln(\omega_i^t)-\omega_i^t+1\right).
\end{equation}
Then we define an objective function written as
\begin{equation}\label{Lc-function-original}
\boldsymbol{H}(r_i^t,\omega_i^{t},\mu^{t},\bar{r})=\frac{\omega_i^{t} {r_i^t}^2+\boldsymbol{\Psi}(\omega_i^{t},\mu^{t},\bar{r})}{{\mu^{t}}^2\bar{r}^2},
\end{equation}
and after its first order derivation w.r.t. $\omega_i^{t}$ we can have 
\begin{equation}
\frac{\partial}{\partial\omega_i^{t}}\boldsymbol{H}(r_i^t,\omega_i^{t},\mu^{t},\bar{r})=\frac{{r_i^t}^2+\mu^2\bar{r}^2\ln(\omega_i^t)}{{\mu^{t}}^2\bar{r}^2}=0,
\end{equation}
so that the weight in the next iteration can be updated in closed-form as
\begin{equation}
\omega_{i}^{t+1}=e^{-\frac{{r^t_{i}}^2}{{\mu^t}^2\bar{r}^2}},
\end{equation}
where according to \cite{black1996unification}, the cost function of \eqref{Lc-function-original} can be derived as
\begin{equation}
\boldsymbol{\rho}(r_i^{t},\mu^{t},\bar{r})=1-e^{-\frac{r_i^2}{{\mu^{t}}^2 \bar{r}^2}}.
\end{equation}
Apparently, $\underset{\omega_i^{t}\in[0,1]}{\min} \sum_{i=1}^N \boldsymbol{H}(r_i^t,\omega_i^{t},\mu^{t},\bar{r})$ is equivalent to the minimization problem in Lemma 1 so that Lemma 1 can be derived.
\end{proof}

\subsection{A Remark on RANSIC}

RANSIC is our scale estimating and inlier searching method, rendered in the main article. Though we cannot guarantee that all the inlier 3-point sets that pass the two compatibility tests in RANSIC must be absolutely inliers, we empirically find that RANSIC is robust and effective even with 99\% outliers. 

For known-scale registration, in IRON, the RANSIC process is also required, but it simply aims at seeking inliers for our heuristic RT-GNC. During this process, we add another condition before our scale-based compatibility test, which is to check whether the rough scales computed from this randomly-generated 3-point set are close to 1. If so, we can continue our compatibility tests in RANSIC; if not, we directly skip this 3-point set and proceed to the next random sample. When the noise level is not high (e.g. $\sigma=0.01$), this condition could be:
\begin{equation}
0.95\leq s_i(\mathbf{\widetilde{P}}_i,\mathbf{\widetilde{Q}}_i)\leq 1.05.
\end{equation}

\subsection{Parameter Setup for IRON}

In Algorithm 1, we can set $\alpha=4.7\sigma-5.2\sigma$, $\beta=5.9\sigma-6.0\sigma$, and $\gamma=296\sigma-298\sigma$ (usually we just use the median values). When the outlier ratio is lower than 99\%, we can set $X=2$; when the outlier ratio is over 99\%,  we can set $X=2$ or $X=3$ (usually $X=2$ is already enough for over 80\% cases at 99\% outliers).

In Algorithm 2, we can set $\nu=0.4-0.7$ (the smaller the faster, usually we just use $\nu=0.45$), $\eta=1.05$ and $big\_num=150-300$ (usually we use $big\_num=200$).

\subsection{Supplementary Experiments}

\begin{figure*}[t]
\centering

\subfigure[Registration with Known Scale]{
\begin{minipage}[t]{0.245\linewidth}
\centering
\includegraphics[width=0.49\linewidth]{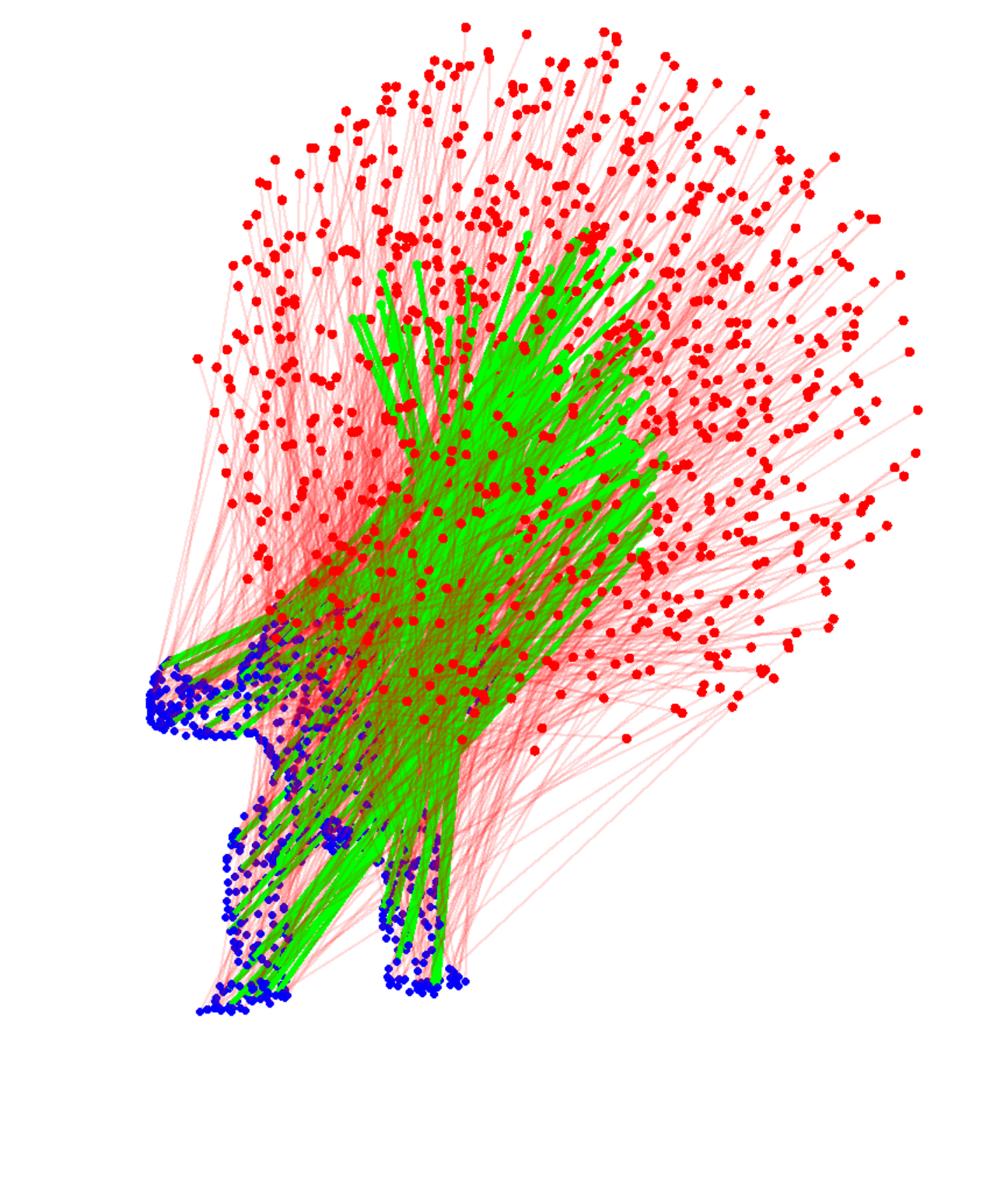}\hspace{-1mm}
\includegraphics[width=0.49\linewidth]{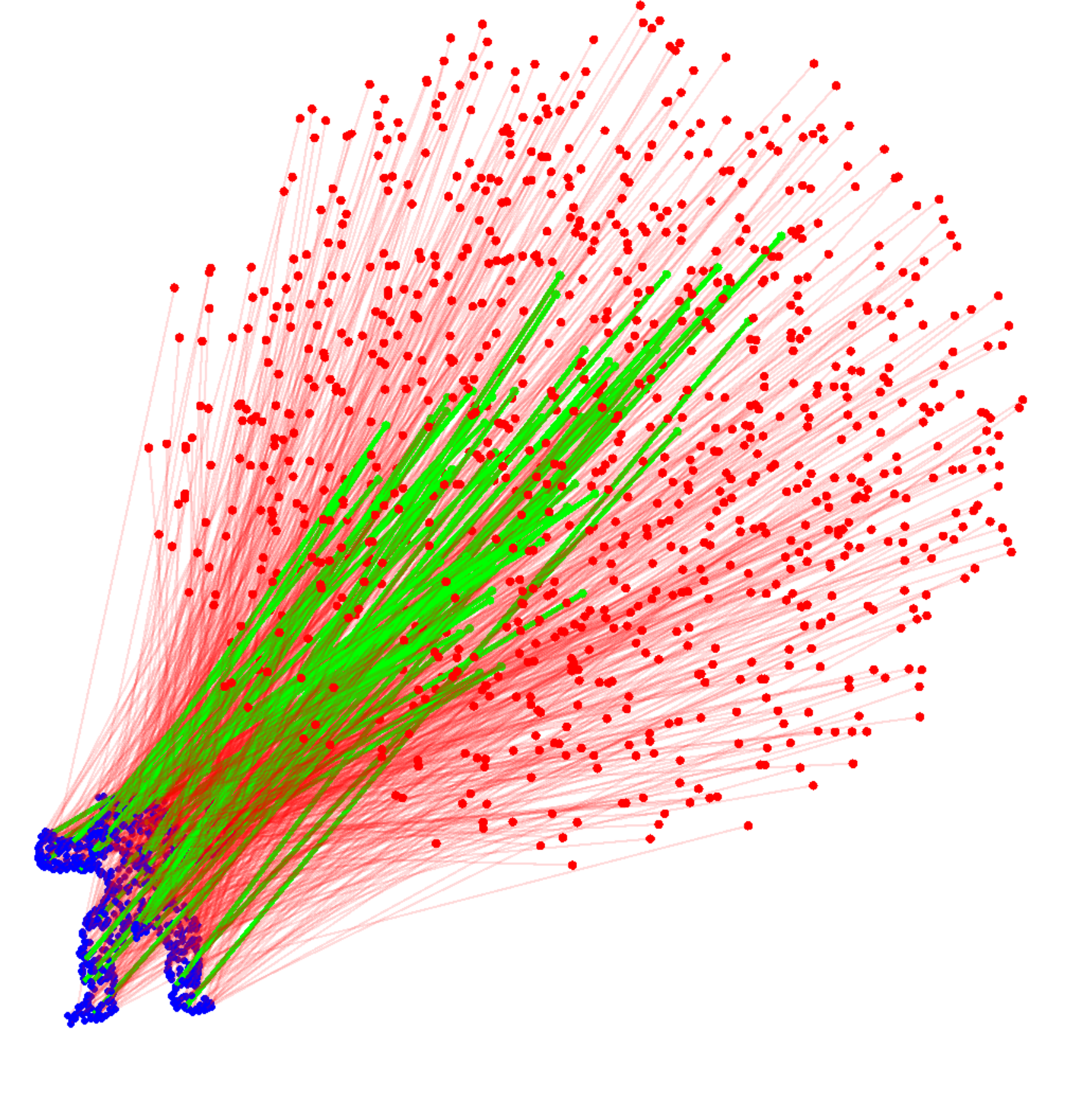}
\includegraphics[width=1\linewidth]{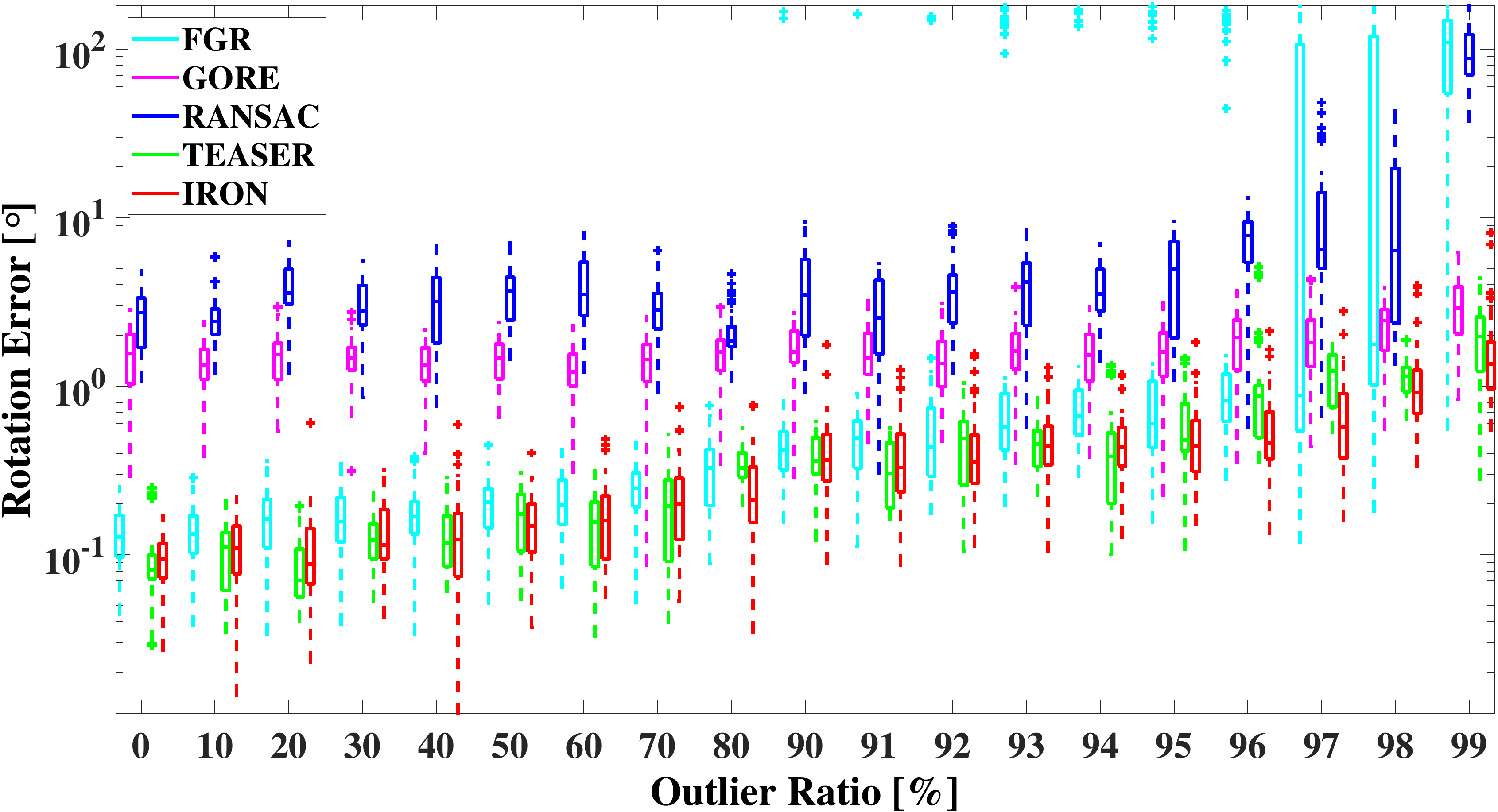}
\includegraphics[width=1\linewidth]{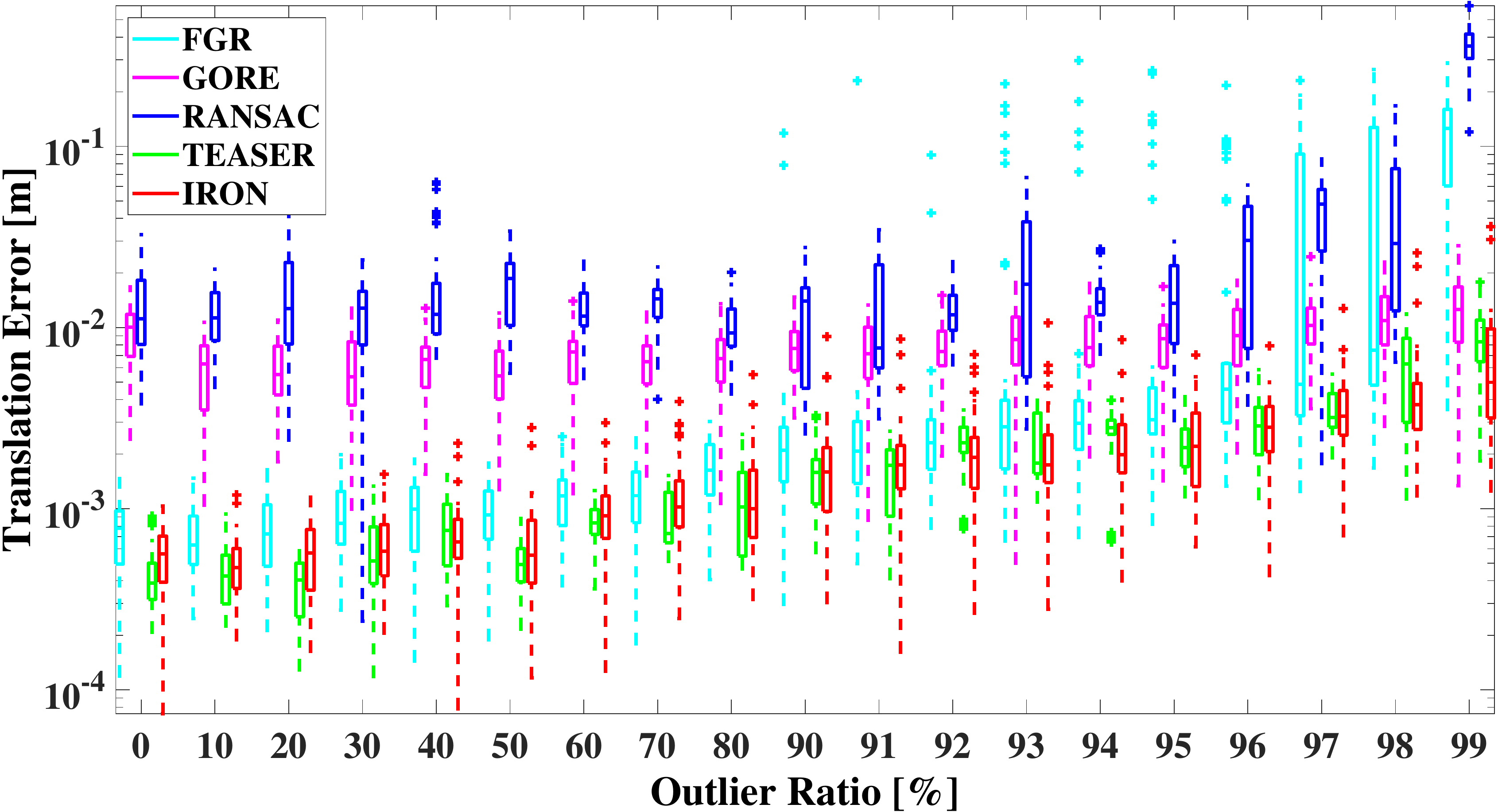}
\includegraphics[width=1\linewidth]{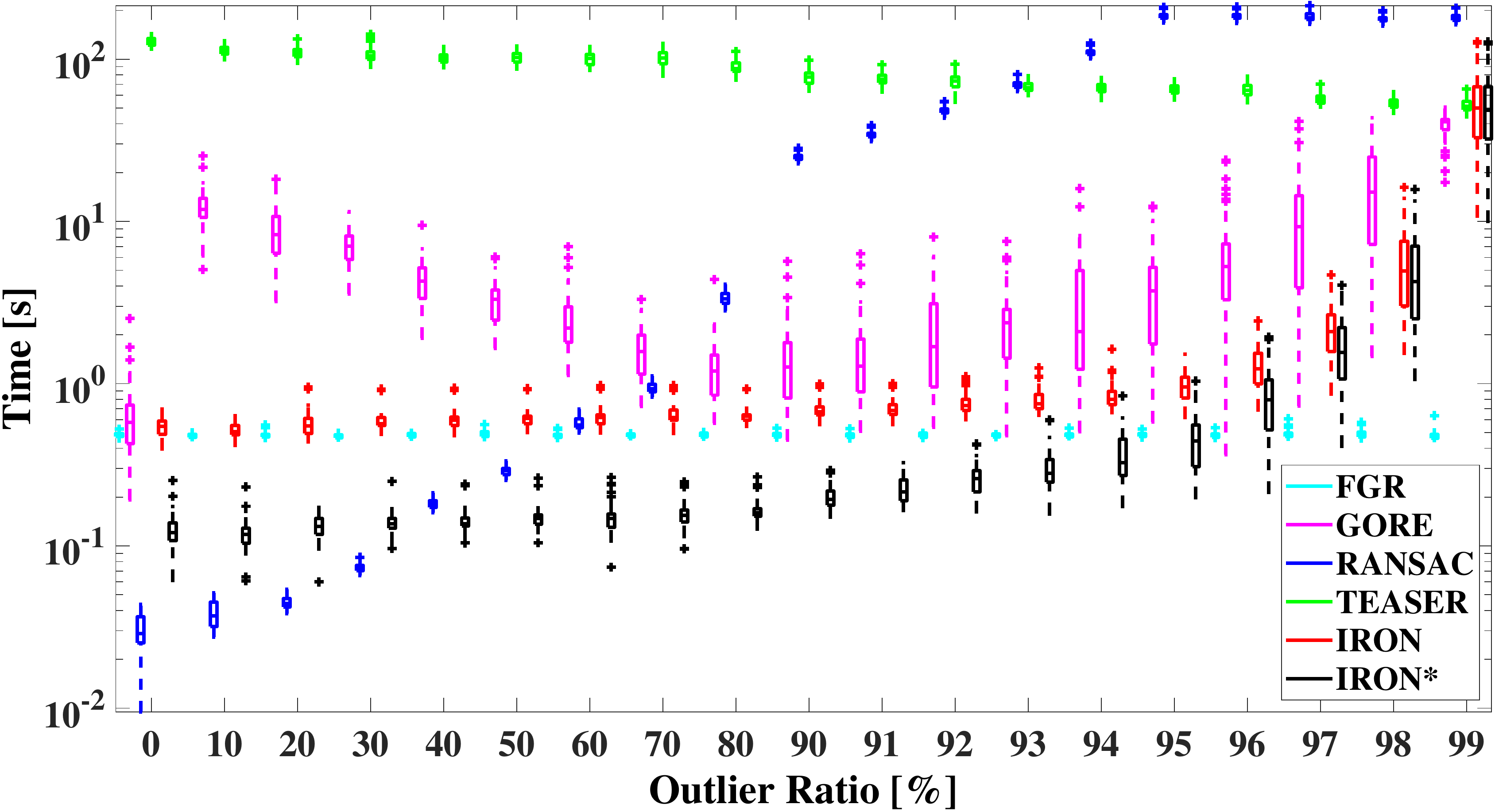}
\end{minipage}
}%
\subfigure[Registration with Unknown Scale]{
\begin{minipage}[t]{0.245\linewidth}
\centering
\includegraphics[width=1\linewidth]{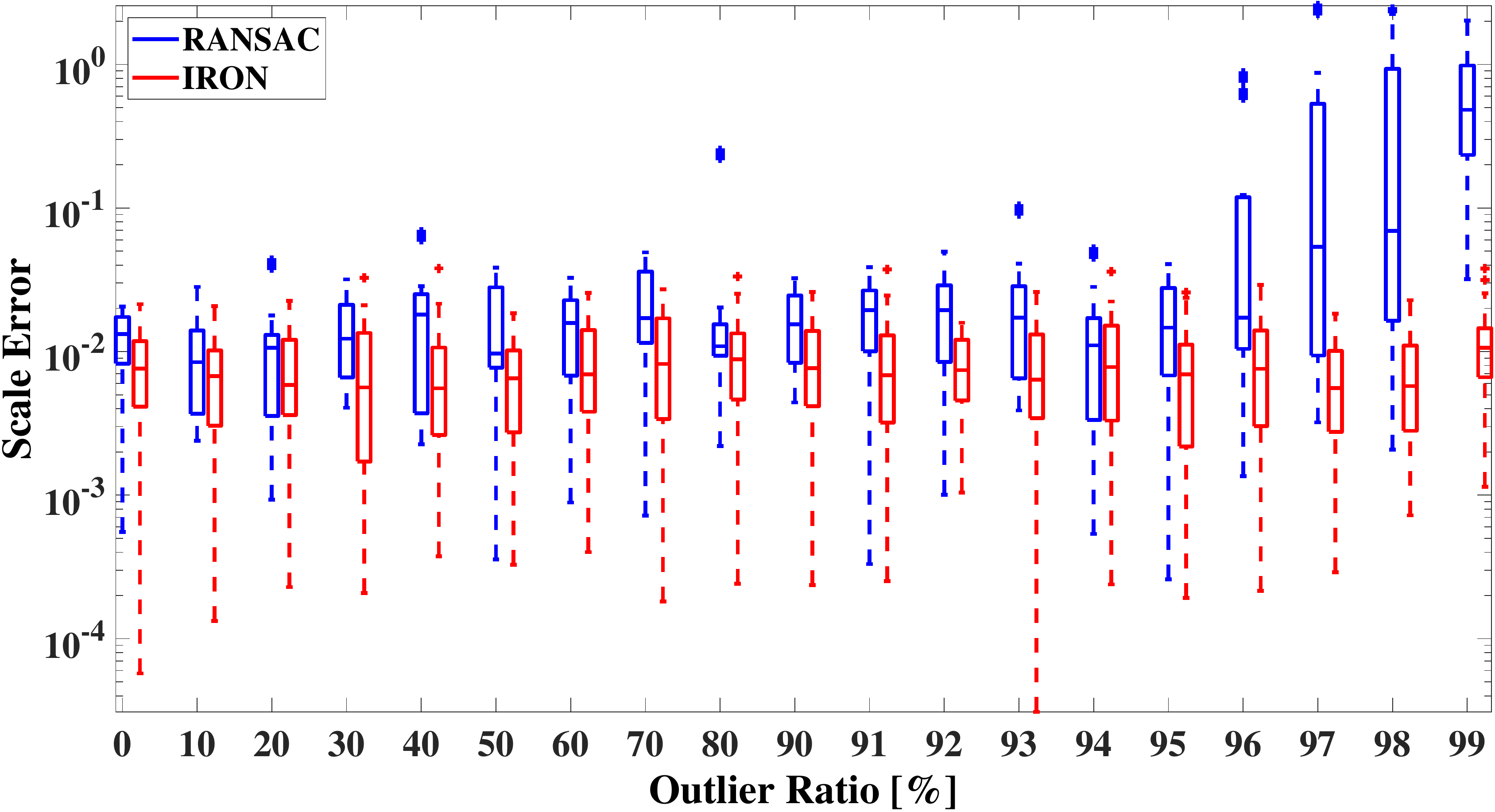}
\includegraphics[width=1\linewidth]{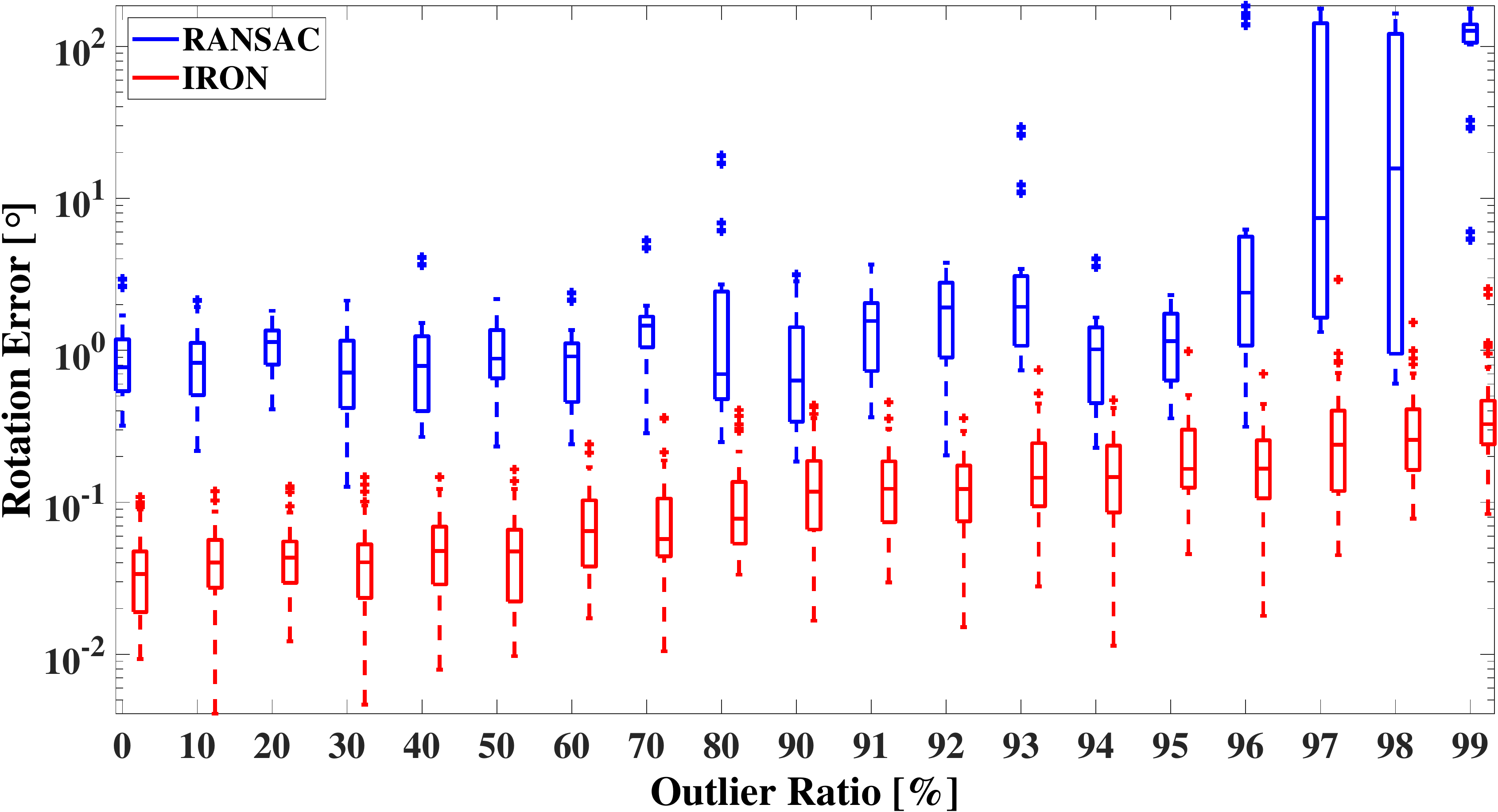}
\includegraphics[width=1\linewidth]{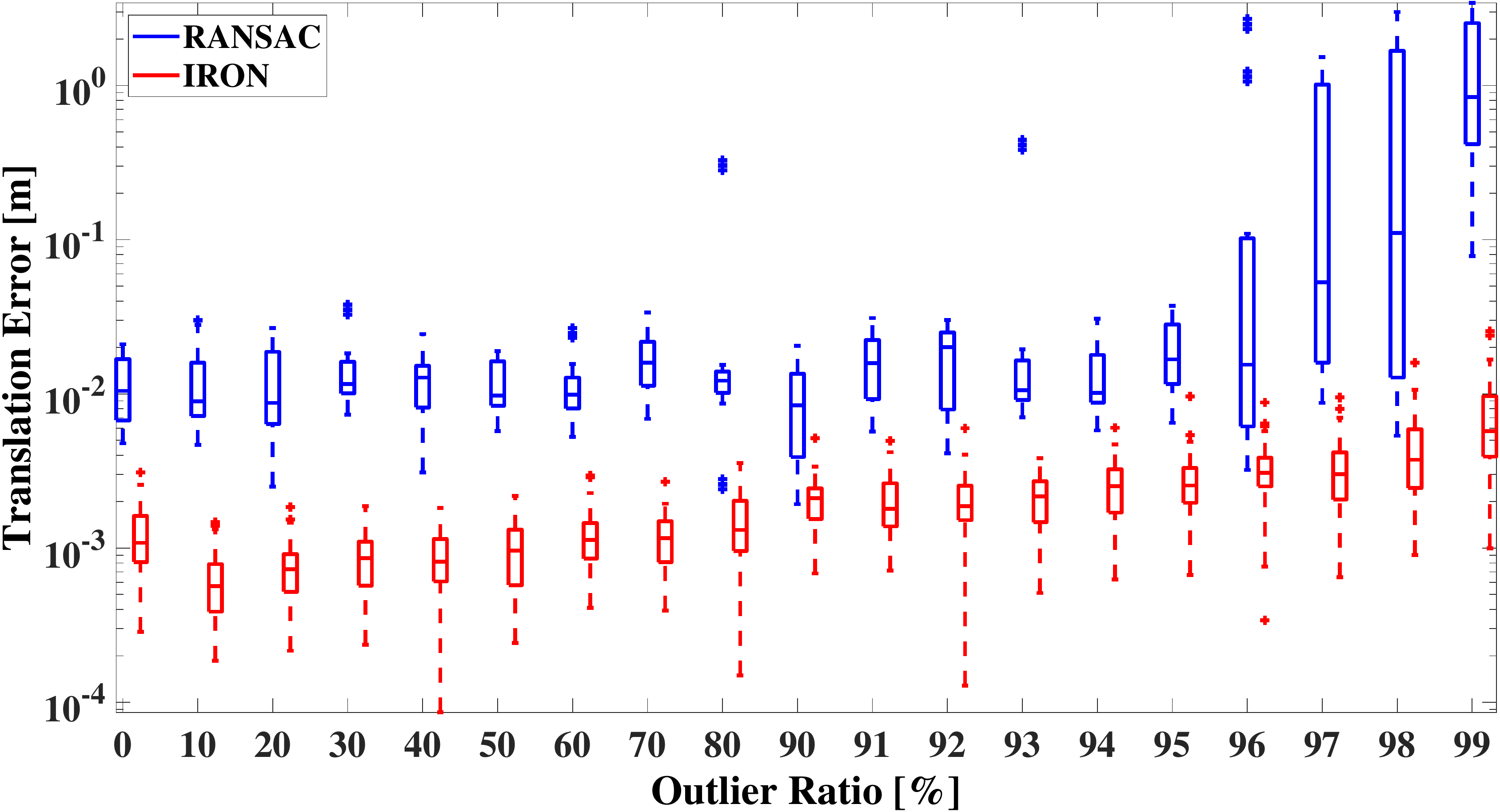}
\includegraphics[width=1\linewidth]{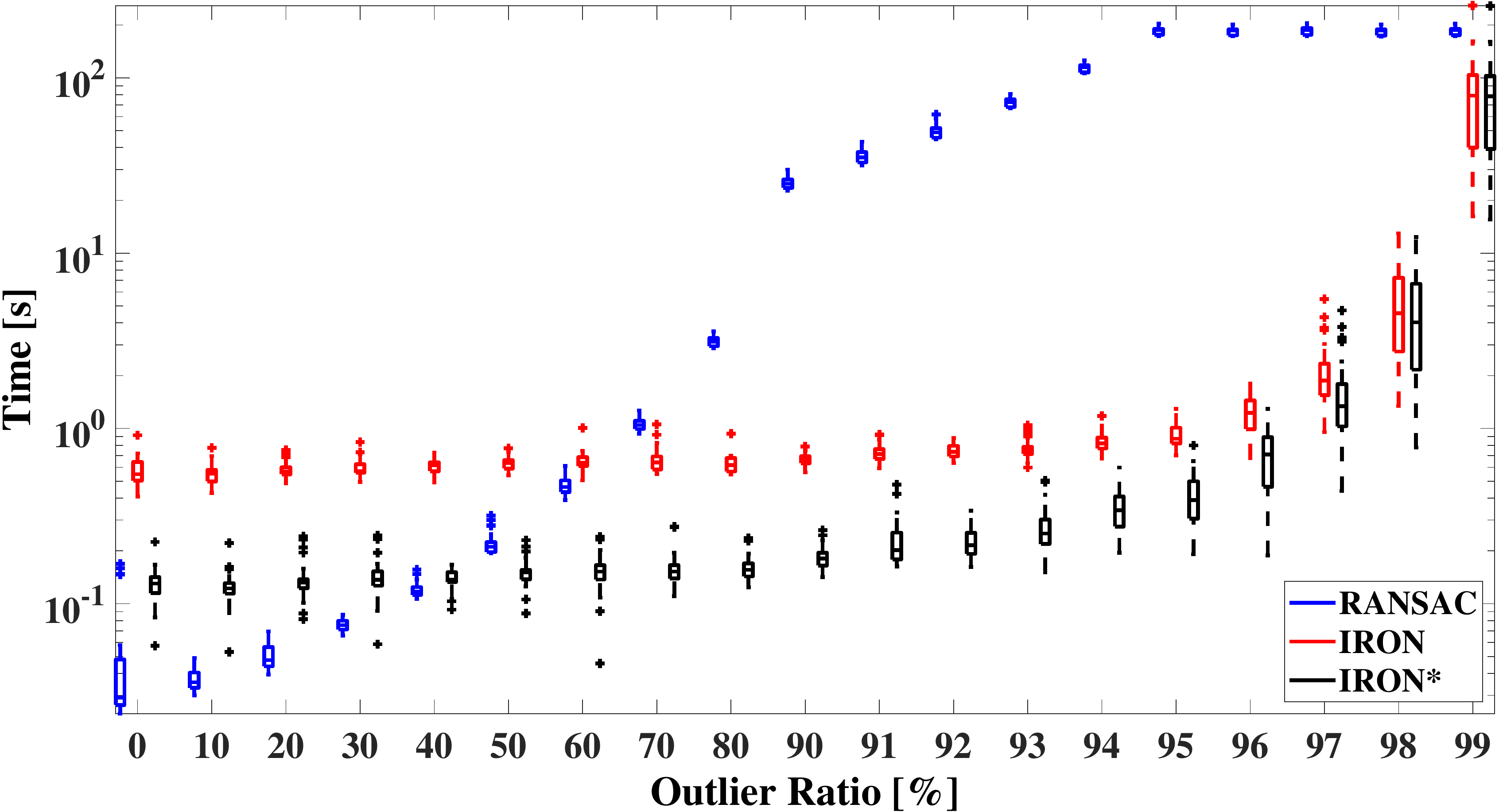}
\end{minipage}
}%
\subfigure[Registration with Known Scale]{
\begin{minipage}[t]{0.245\linewidth}
\centering
\includegraphics[width=0.49\linewidth]{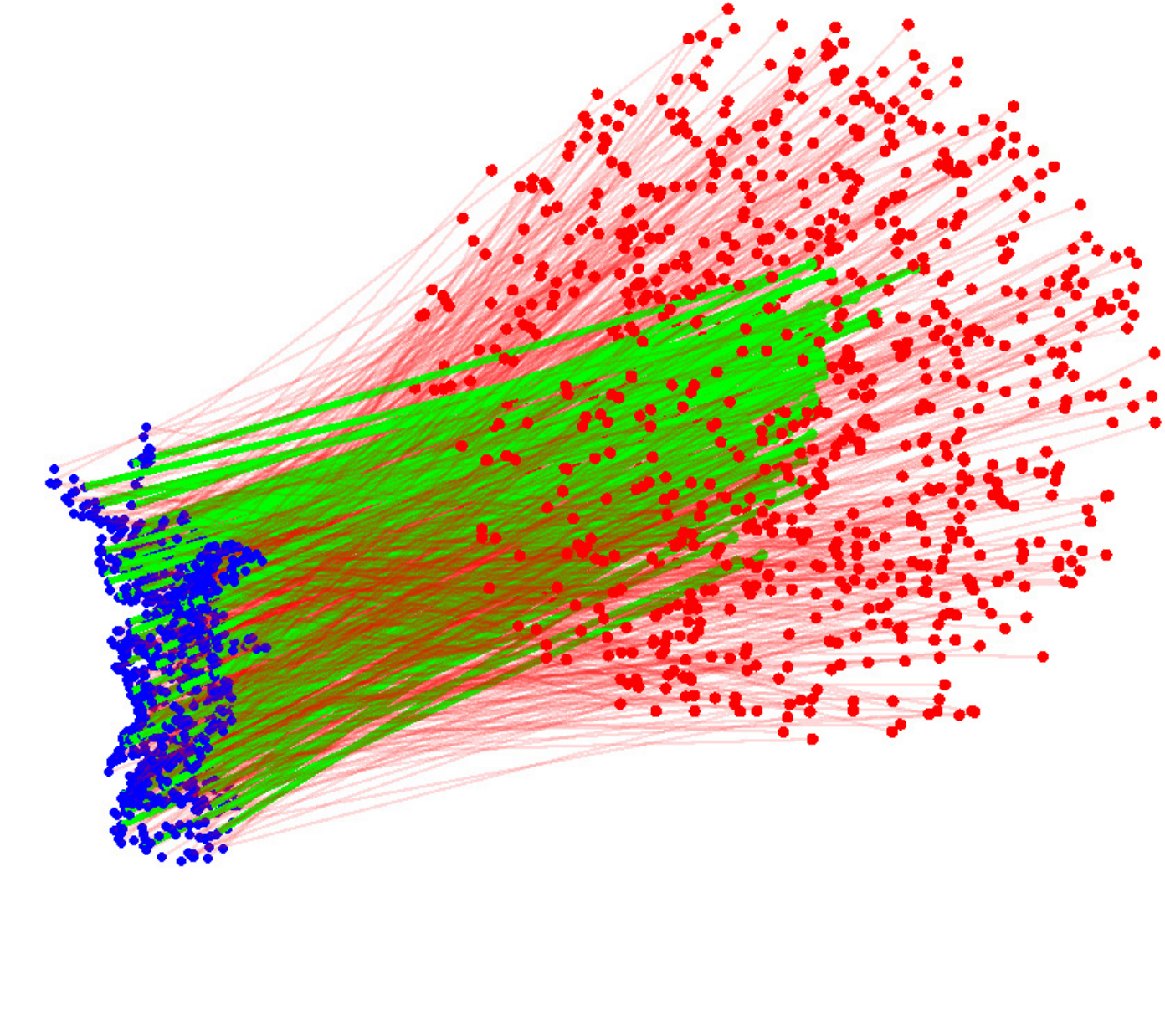}\hspace{-1mm}
\includegraphics[width=0.49\linewidth]{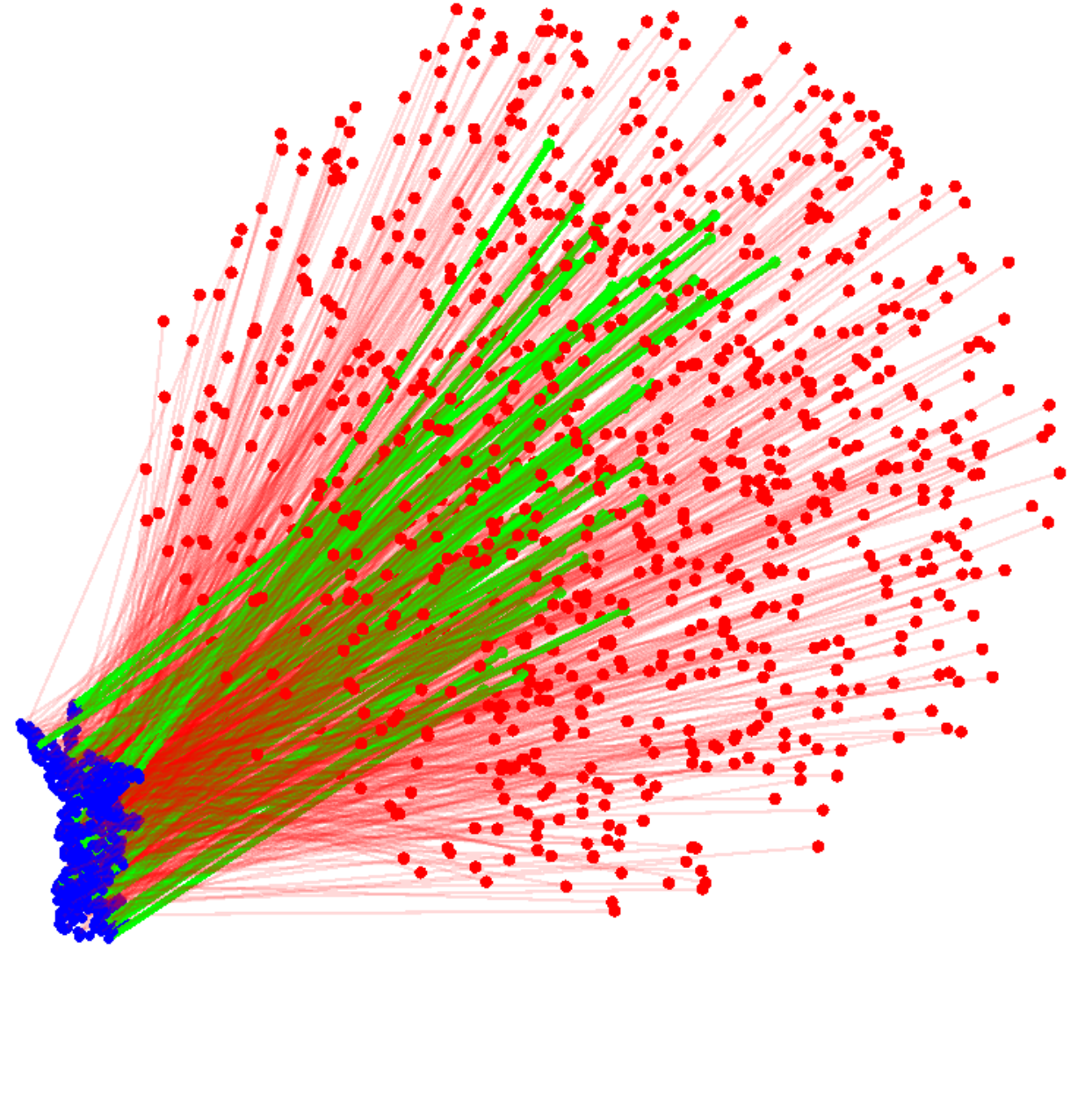}
\includegraphics[width=1\linewidth]{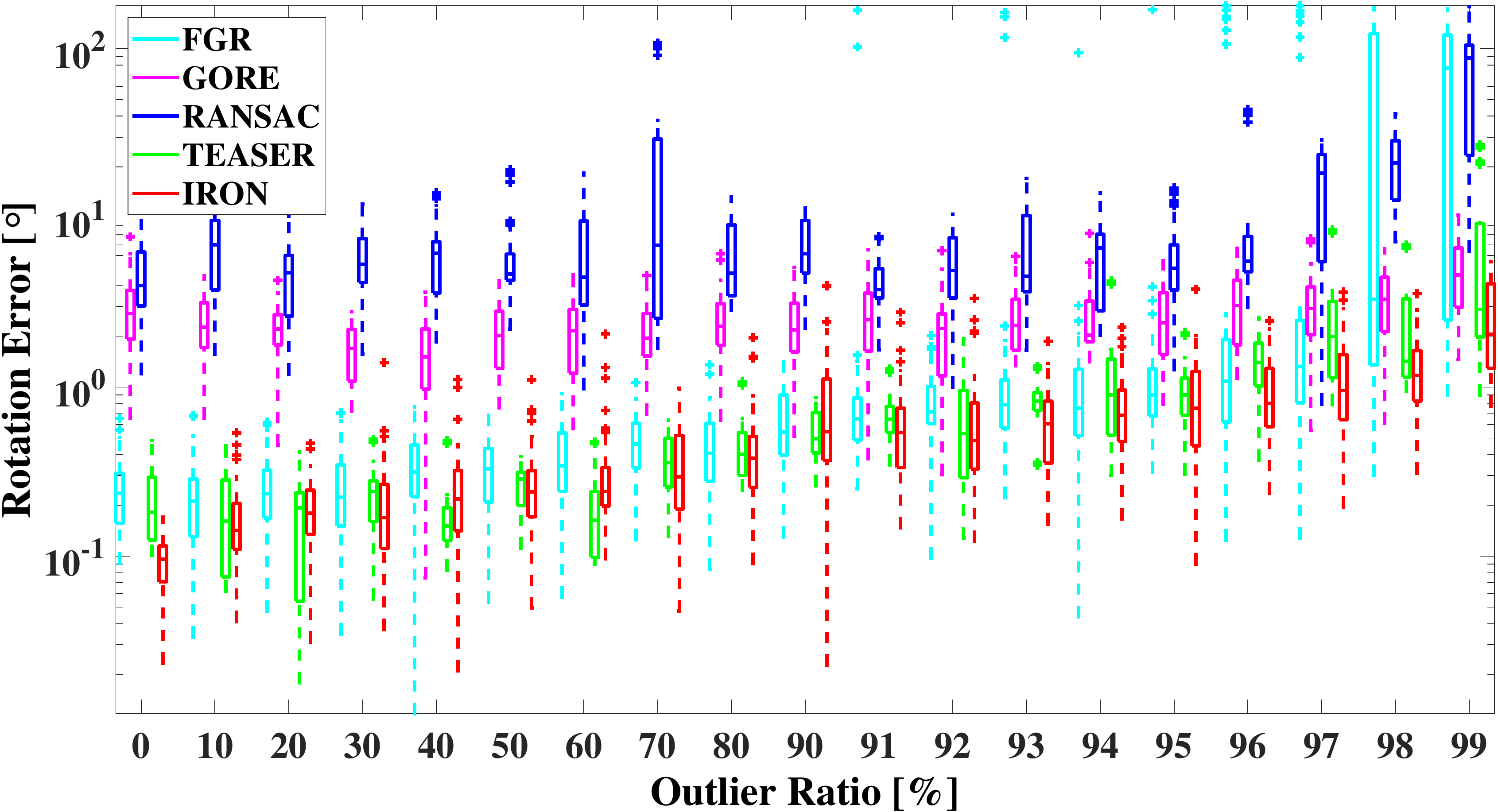}
\includegraphics[width=1\linewidth]{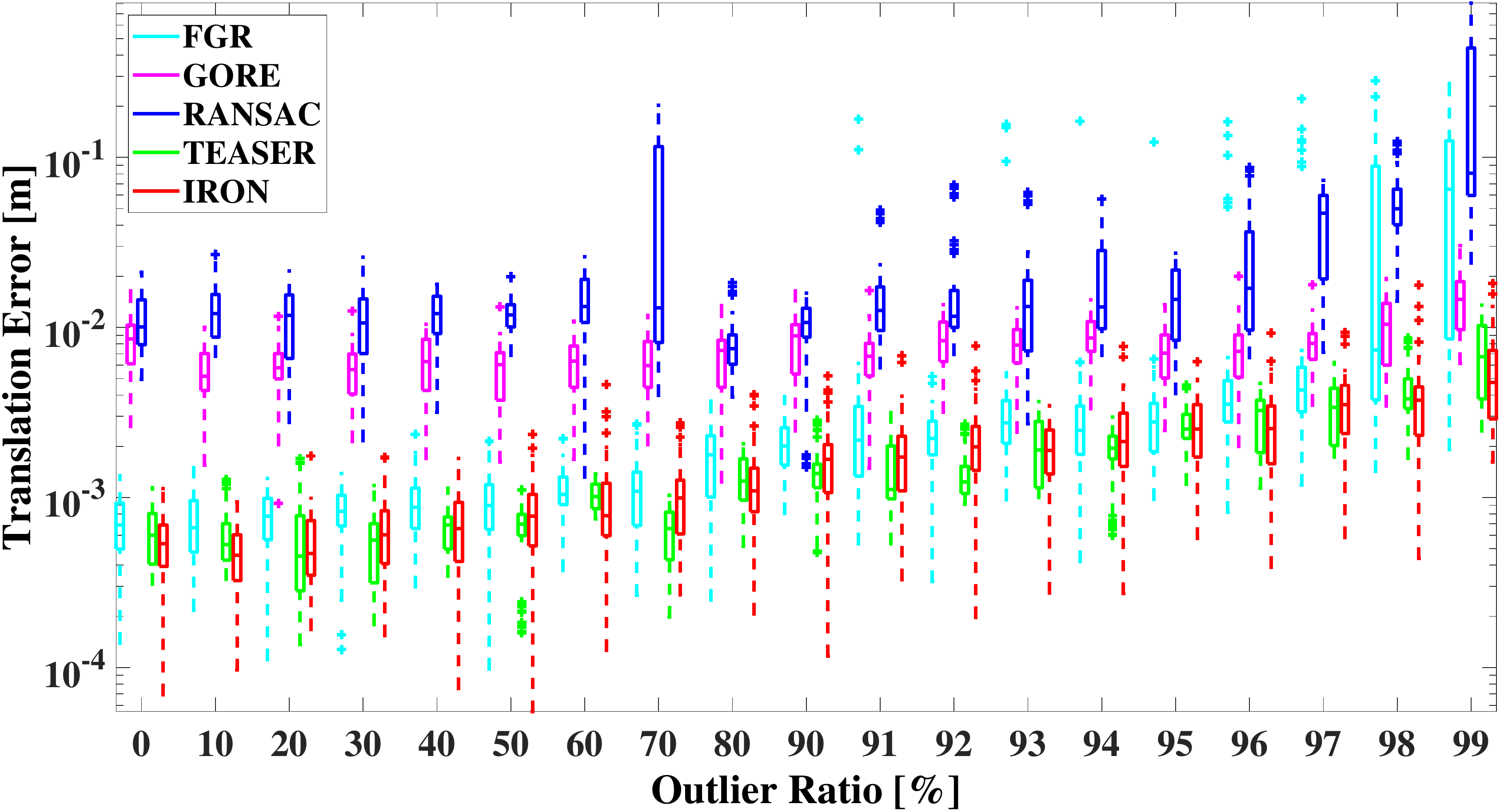}
\includegraphics[width=1\linewidth]{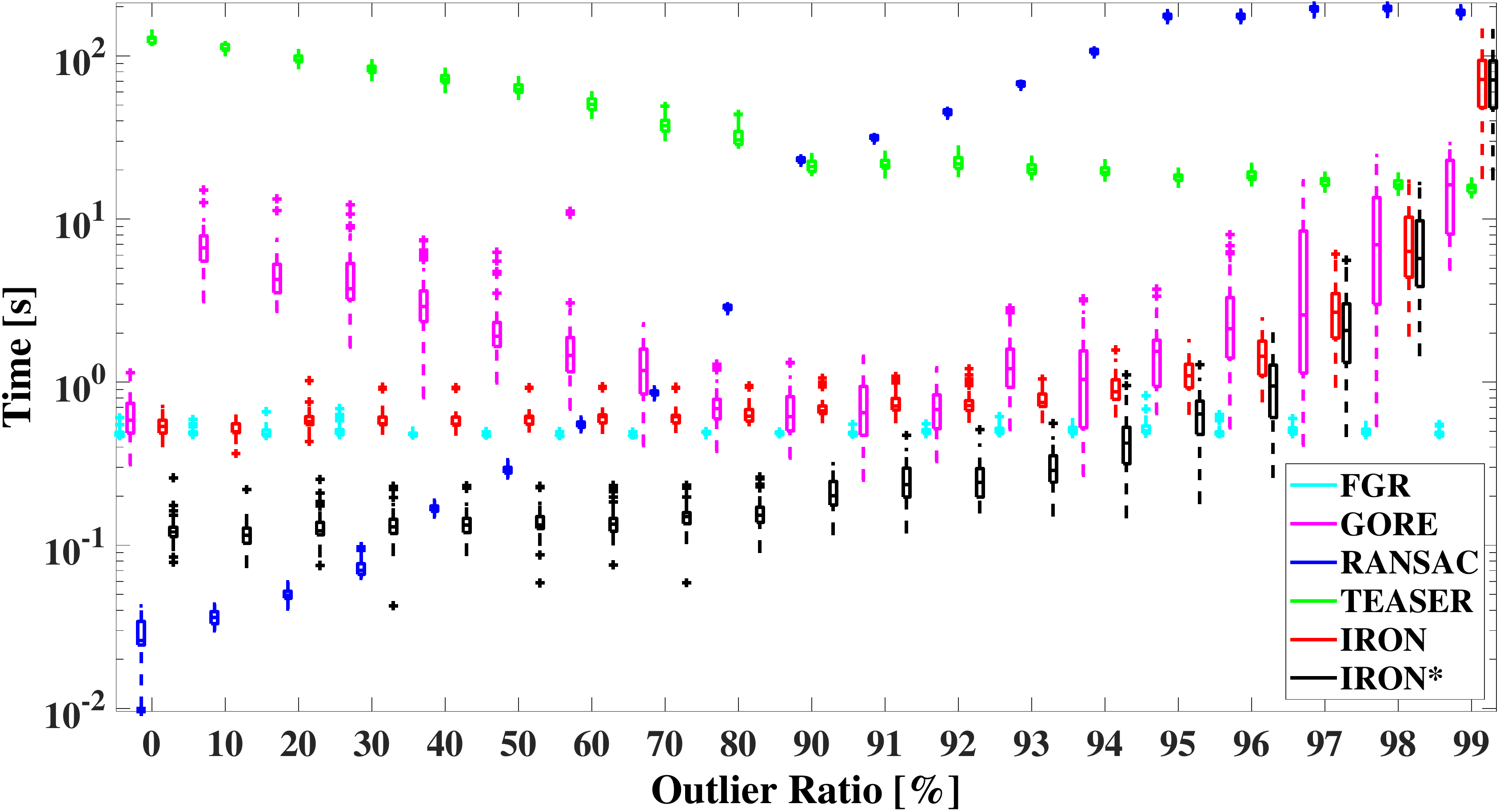}
\end{minipage}
}%
\subfigure[Registration with Unknown Scale]{
\begin{minipage}[t]{0.245\linewidth}
\centering
\includegraphics[width=1\linewidth]{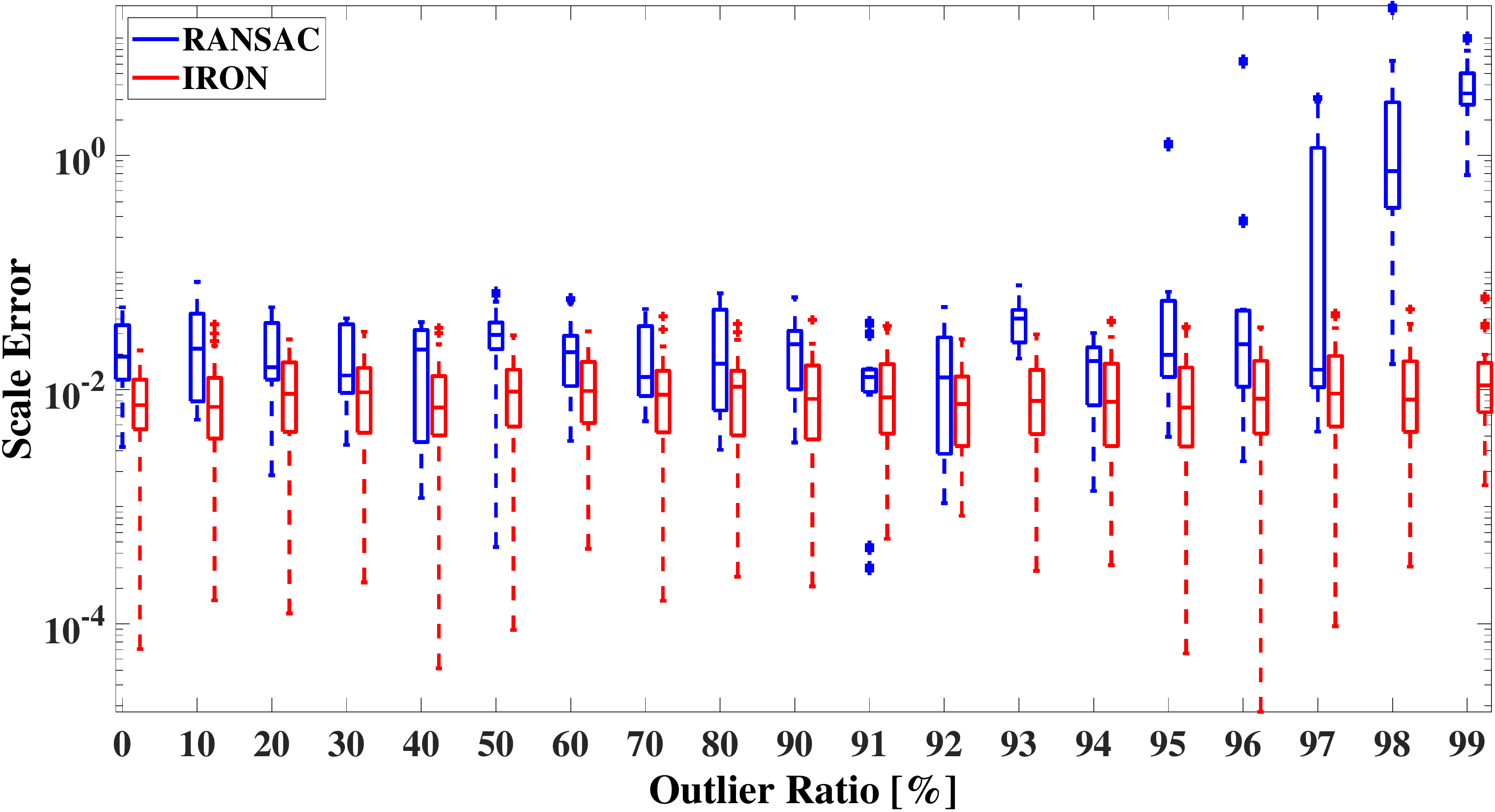}
\includegraphics[width=1\linewidth]{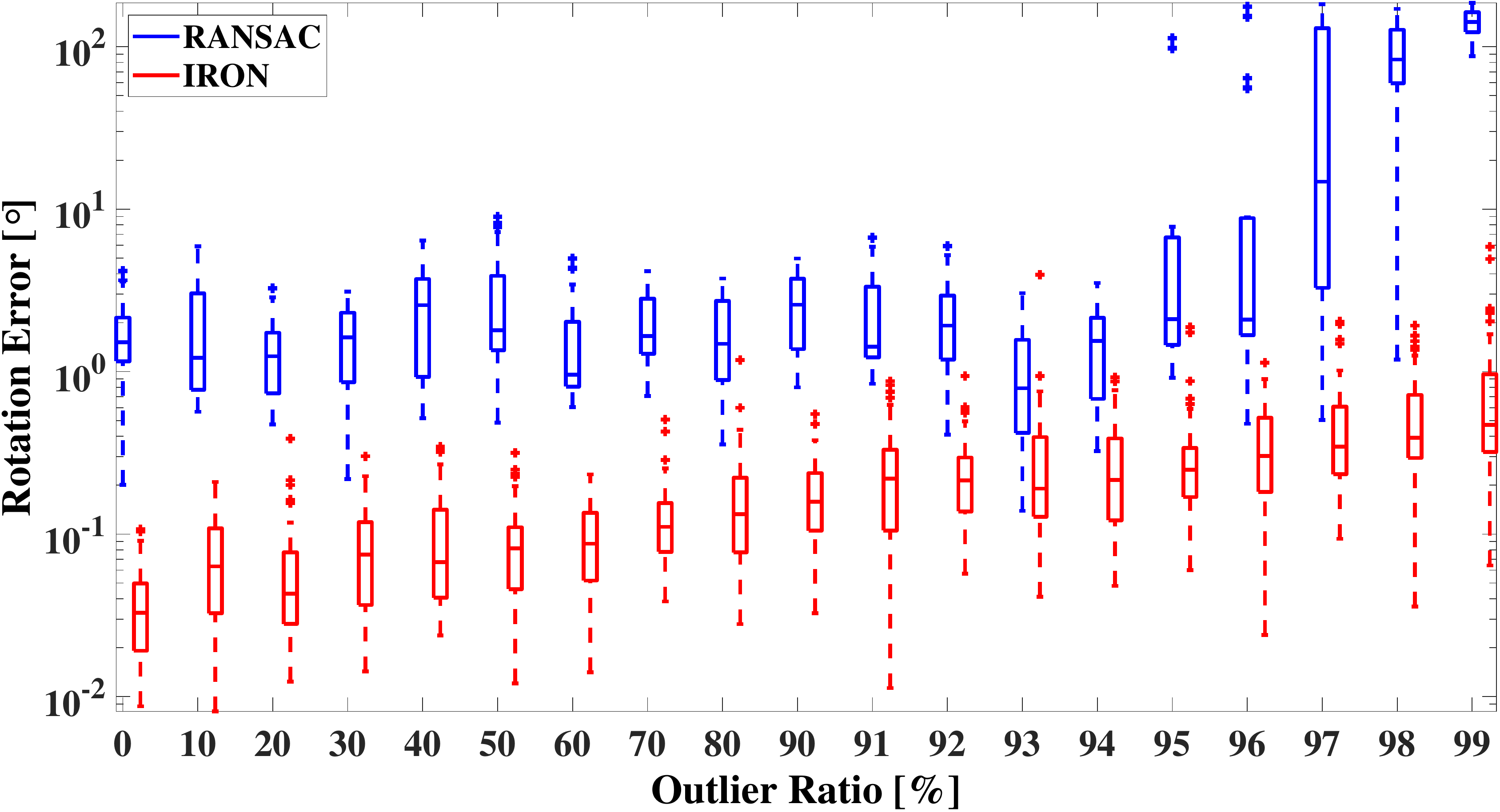}
\includegraphics[width=1\linewidth]{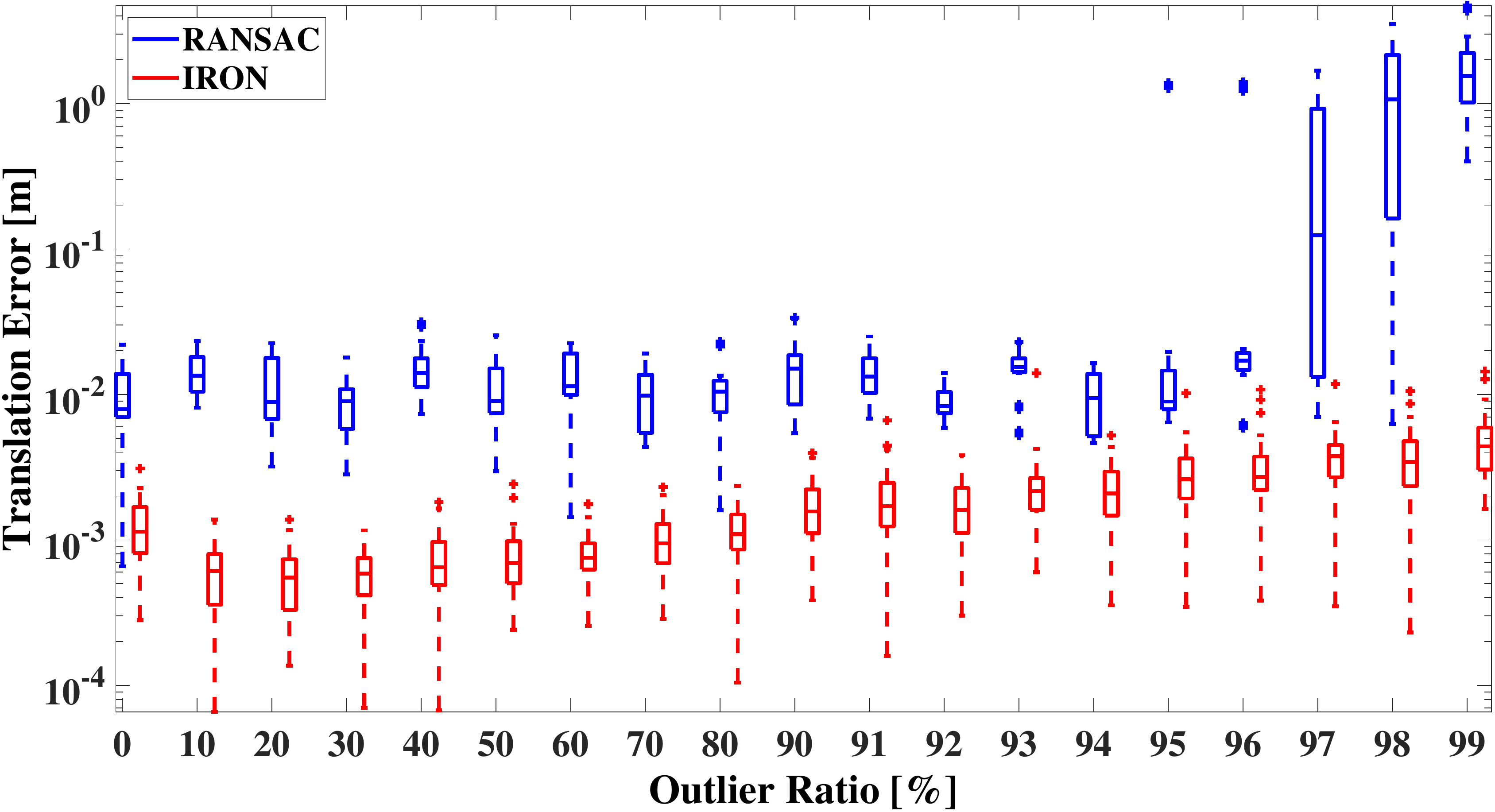}
\includegraphics[width=1\linewidth]{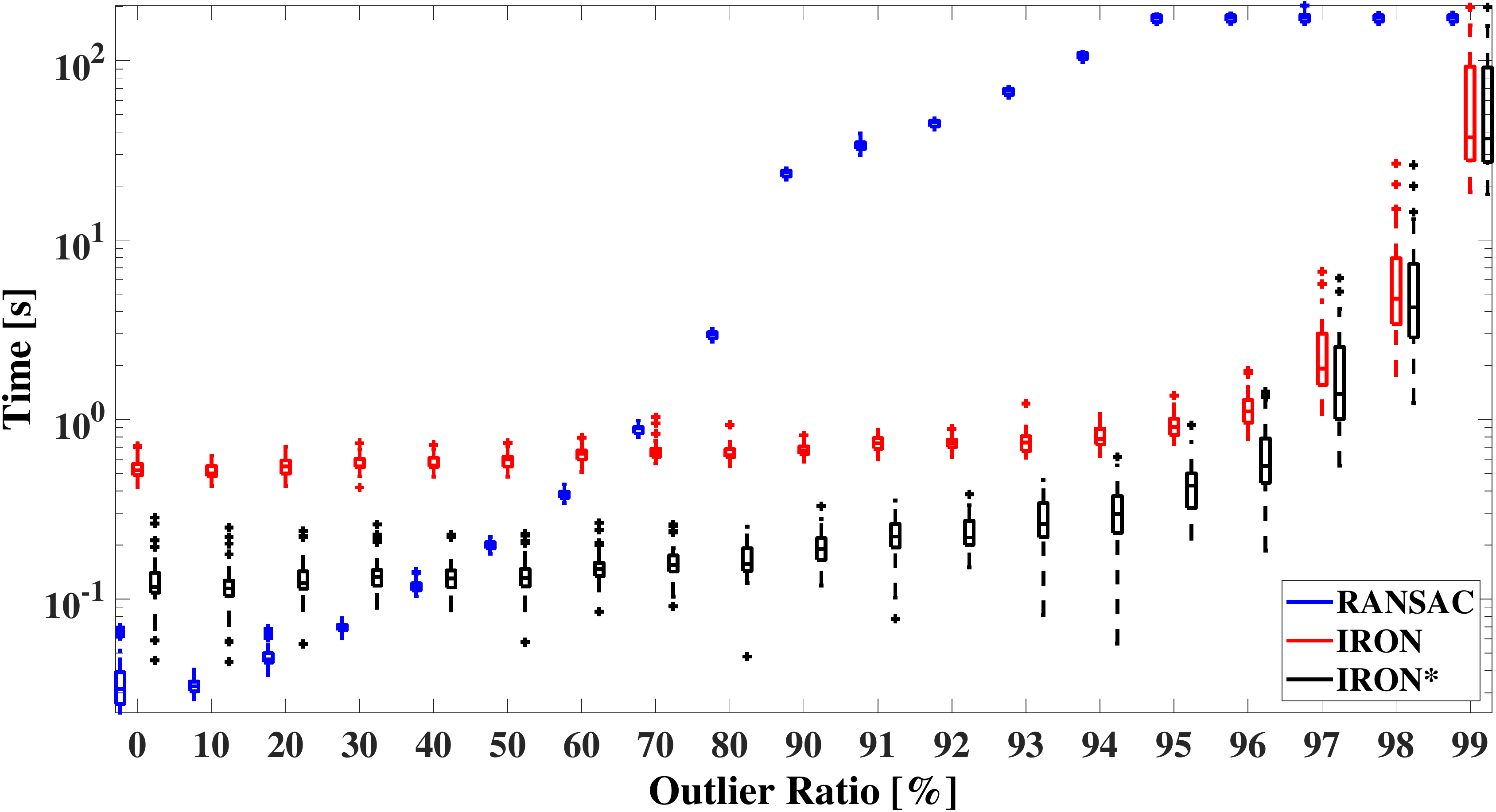}
\end{minipage}
}%

\centering
\caption{Benchmark on standard registration problems ('Armadillo' (a-b) and 'Lucy'  (c-d) in~\cite{curless1996volumetric}). \textbf{(a)} Estimation accuracy and runtime of FGR, GORE, RANSAC, TEASER and IRON with $\mathit{s}=1$. The top-row images demonstrate a known-scale registration problem with 90\% outliers and an unknown-scale registration problem with 95\% outliers.  \textbf{(b)} Estimation accuracy and runtime of RANSAC and IRON. \textbf{(c)} Estimation accuracy and runtime of FGR, GORE, RANSAC, TEASER and IRON with $\mathit{s}=1$.  The top-row images demonstrate a known-scale registration problem with 90\% outliers and an unknown-scale registration problem with 95\% outliers. \textbf{(d)} Estimation accuracy and runtime of RANSAC and IRON.}
\label{OB2}
\vspace{-9pt}
\end{figure*}

\begin{figure*}[t]
\centering

\subfigure[\textit{Scene-02}, Known Scale: ($817,\,96.94\%,\,{3.35\times10^{-5}}^{\circ},6.54\times10^{-7}m$), Unknown Scale: ($594,\,92.93\%,\,6.89\times10^{-7},{0.1158}^{\circ},0.0050 m$)]{
\begin{minipage}[t]{1\linewidth}
\centering
\includegraphics[width=0.24\linewidth]{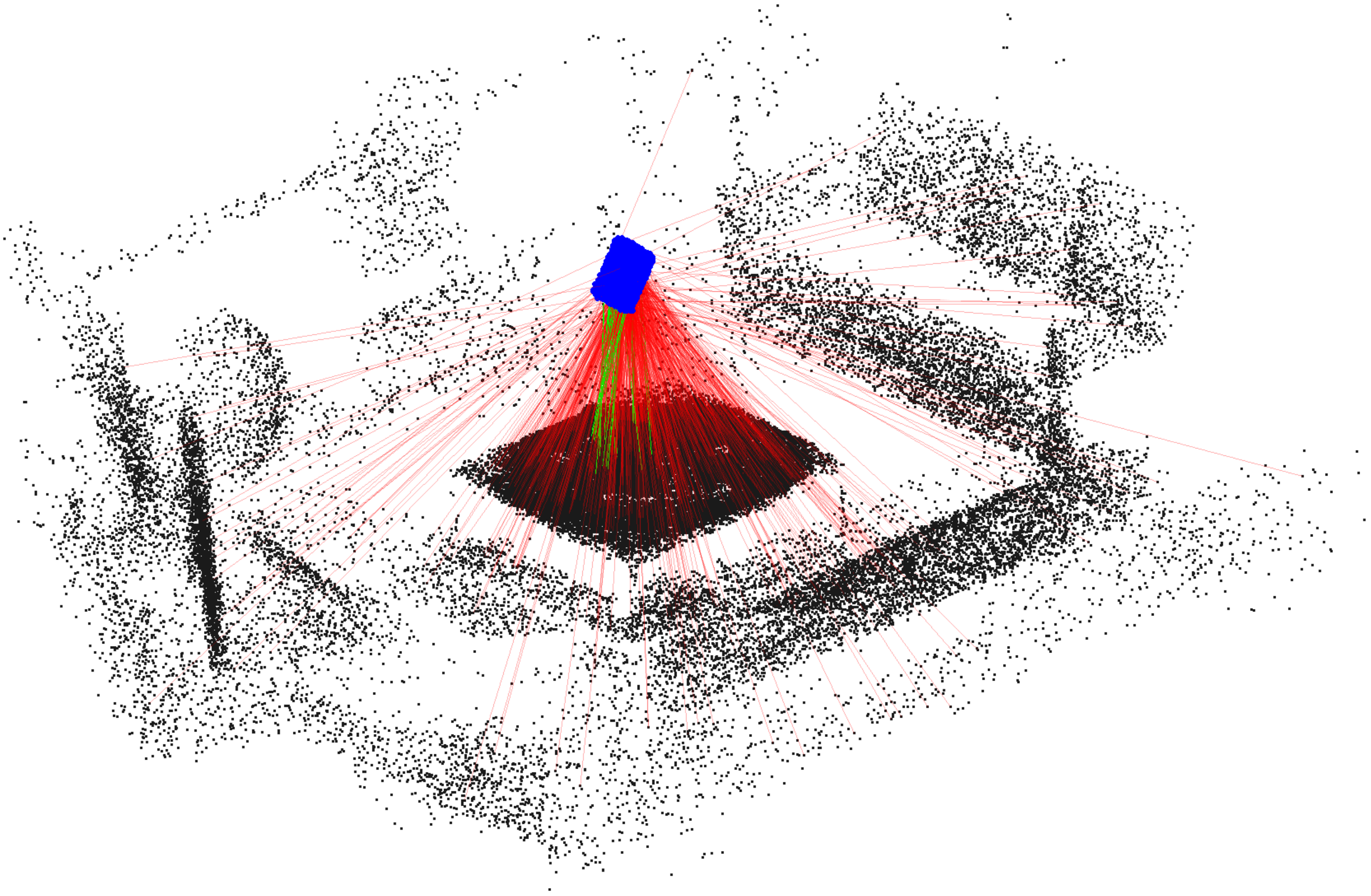}
\includegraphics[width=0.24\linewidth]{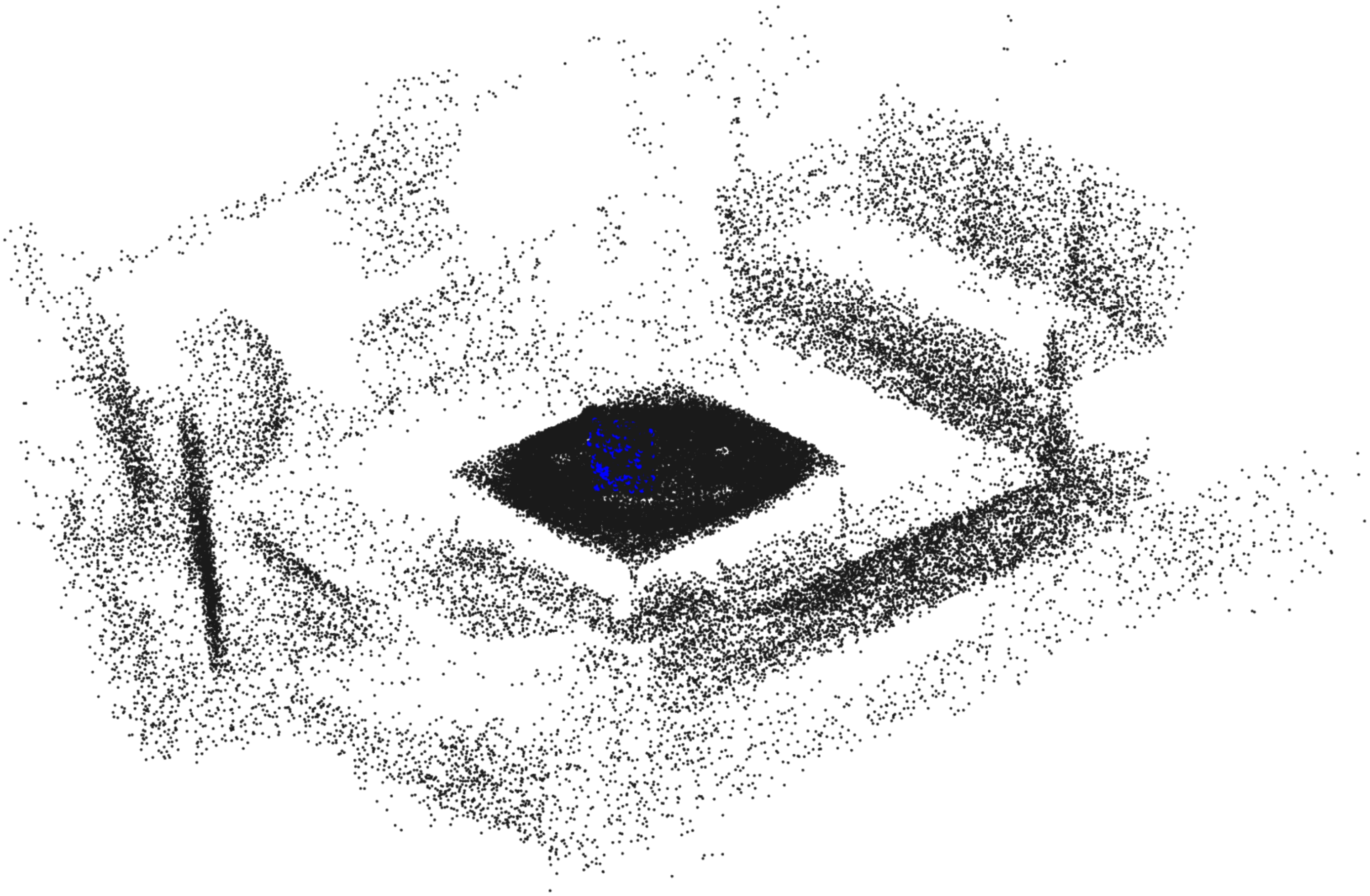}
\includegraphics[width=0.24\linewidth]{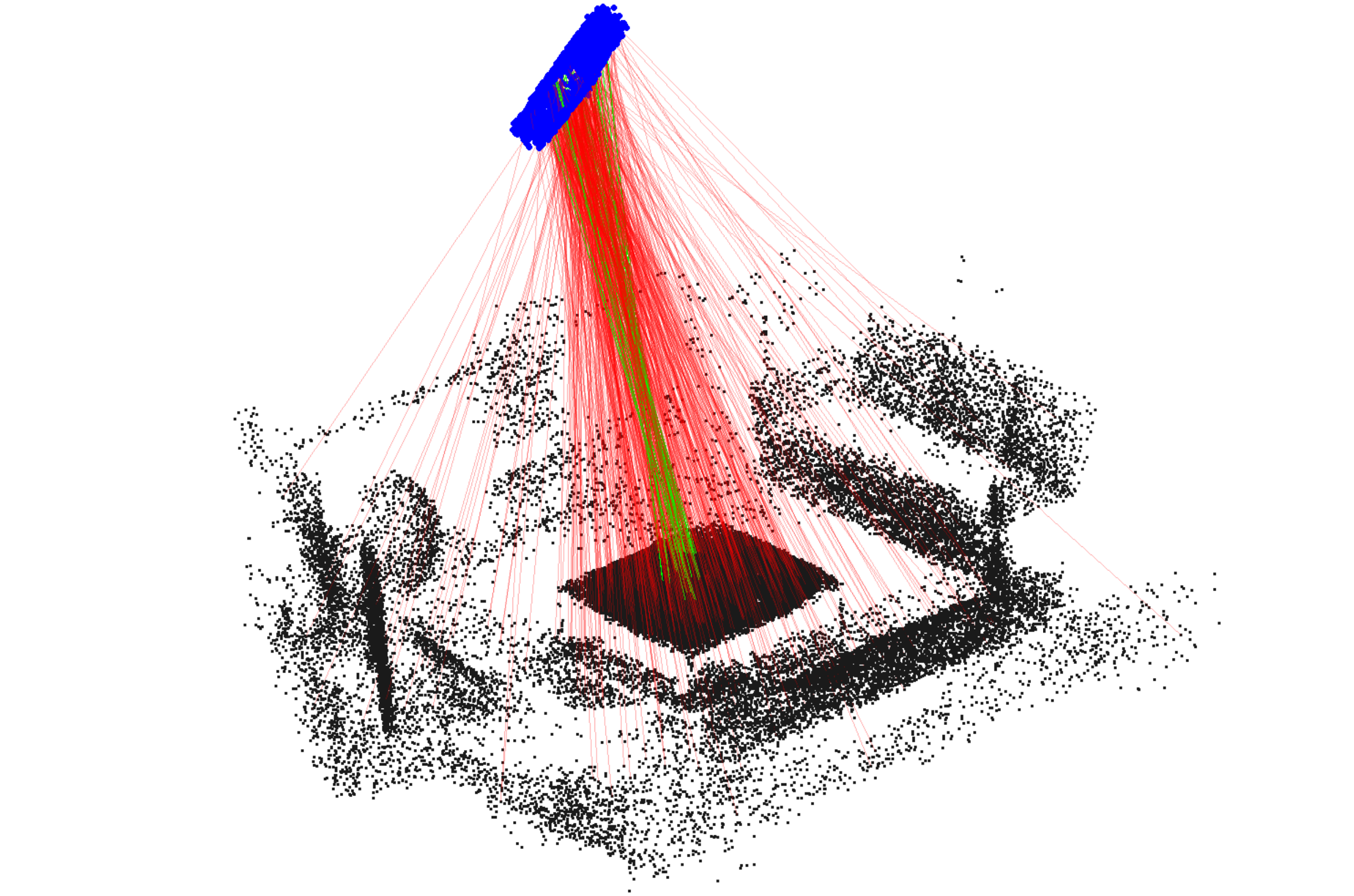}
\includegraphics[width=0.24\linewidth]{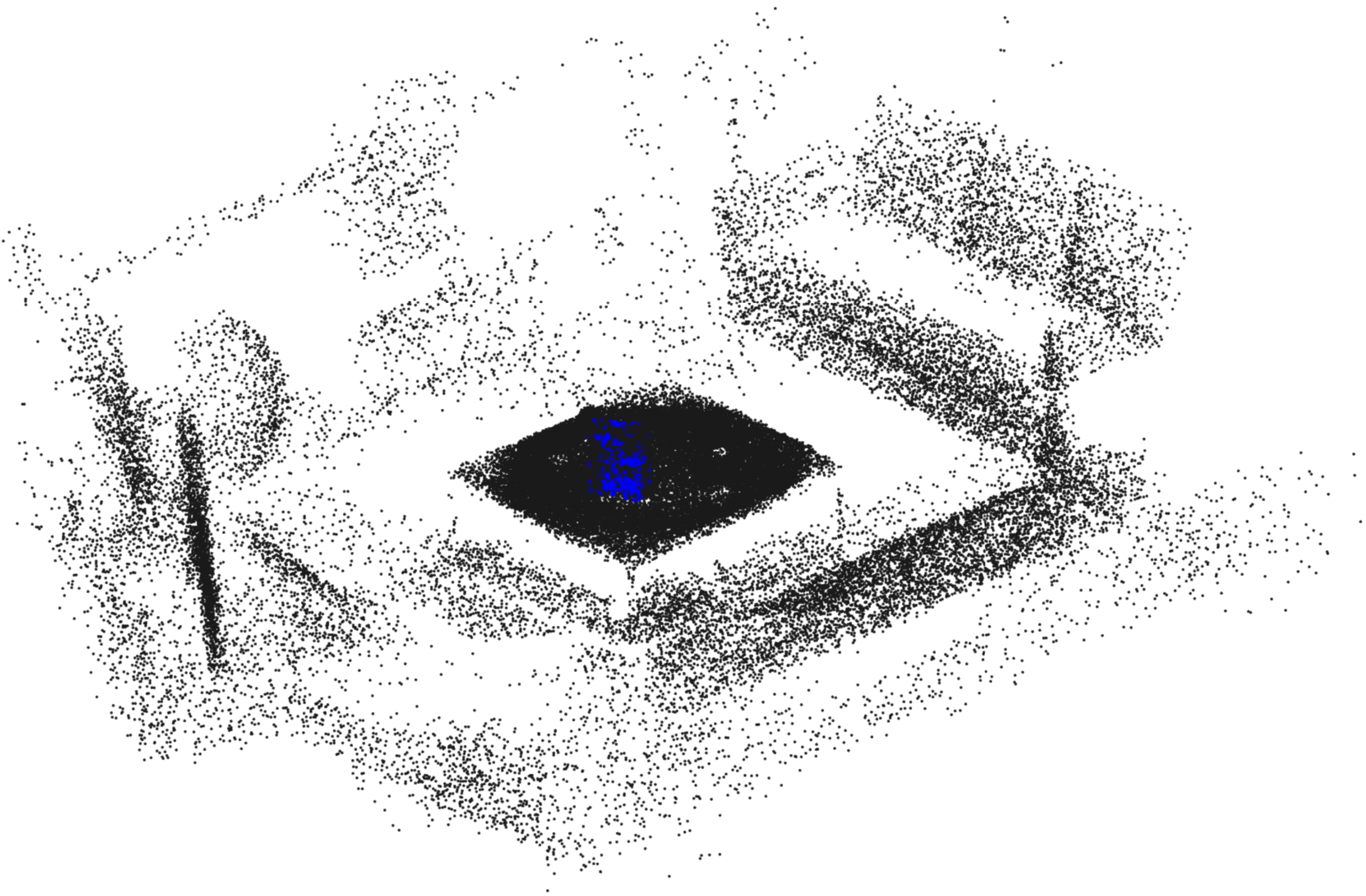}
\end{minipage}
}%

\subfigure[\textit{Scene-04}, Known Scale: ($396,\,92.68\%,\,{0.2679}^{\circ},0.0067m$), Unknown Scale: ($394,\,93.91\%,\,3.02\times10^{-7},{8.36\times10^{-6}}^{\circ},3.532\times10^{-7} m$)]{
\begin{minipage}[t]{1\linewidth}
\centering
\includegraphics[width=0.24\linewidth]{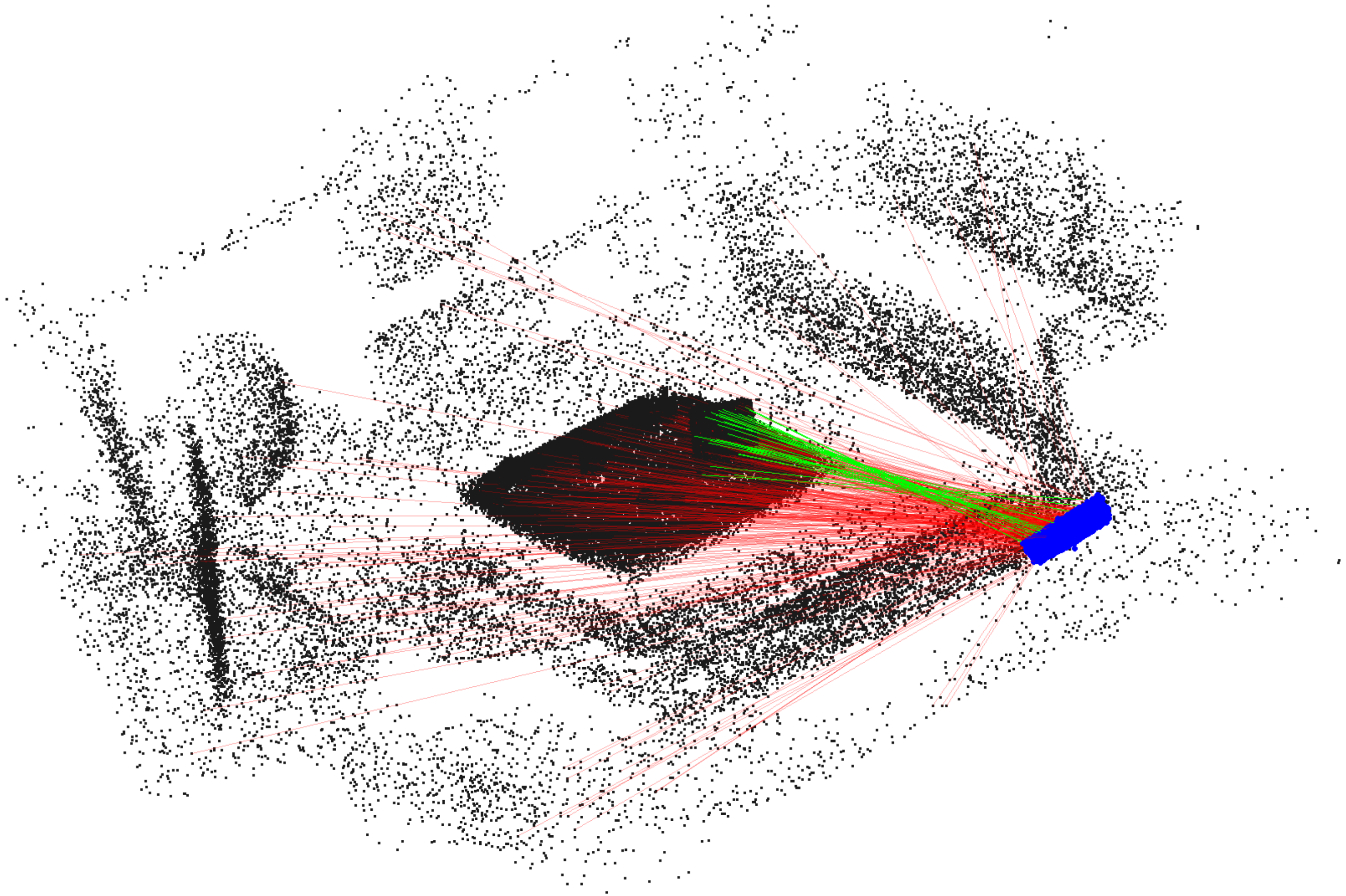}
\includegraphics[width=0.24\linewidth]{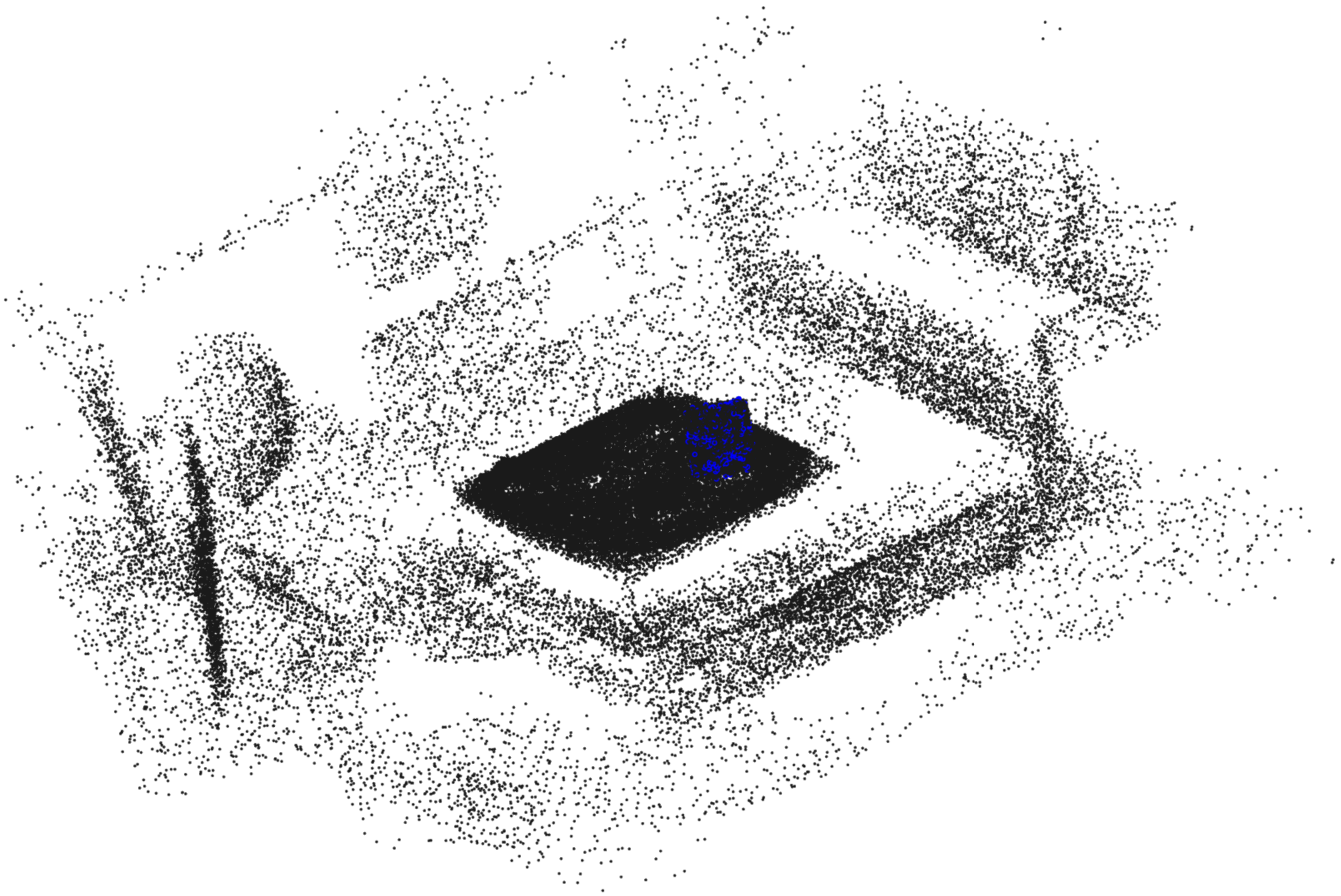}
\includegraphics[width=0.24\linewidth]{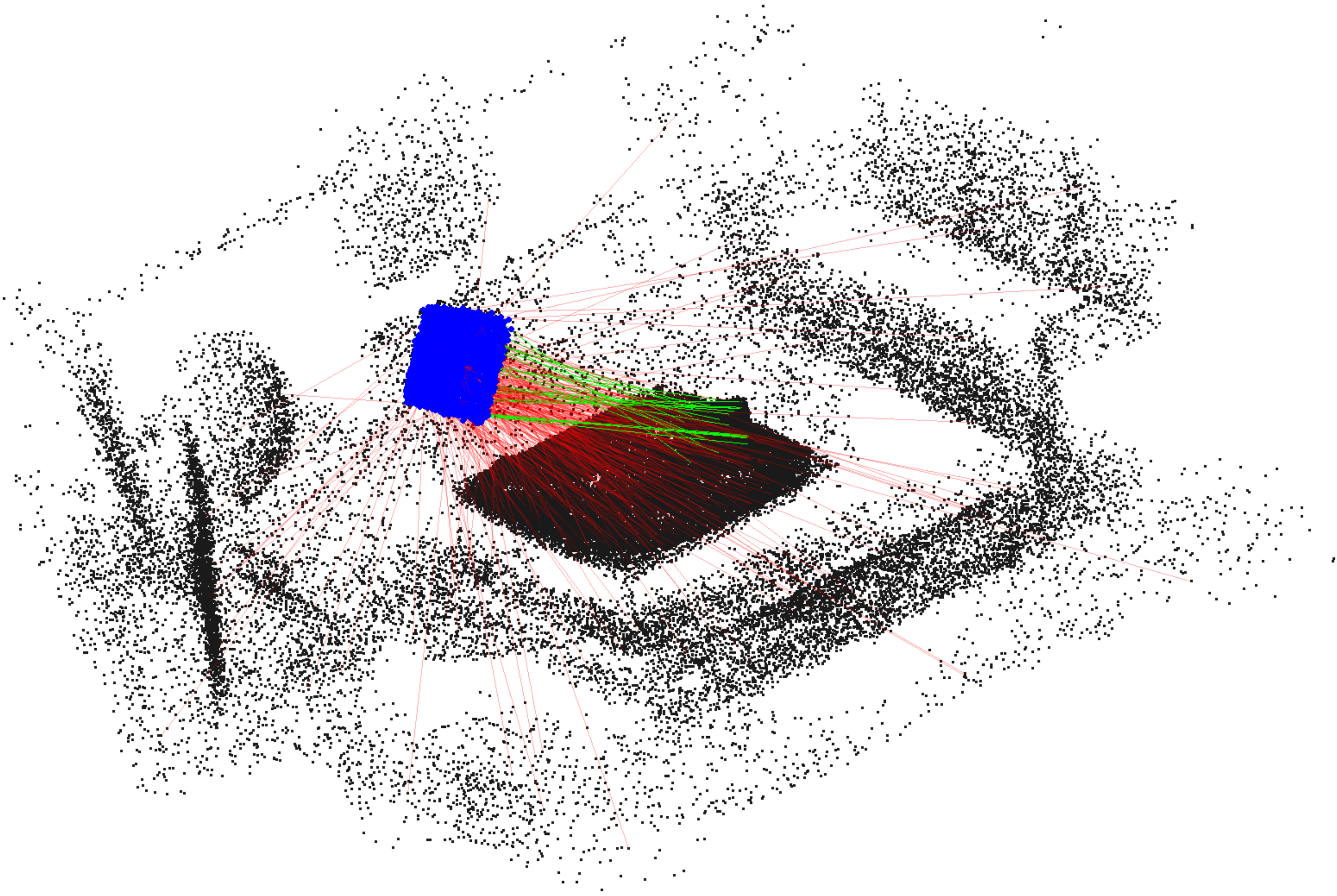}
\includegraphics[width=0.24\linewidth]{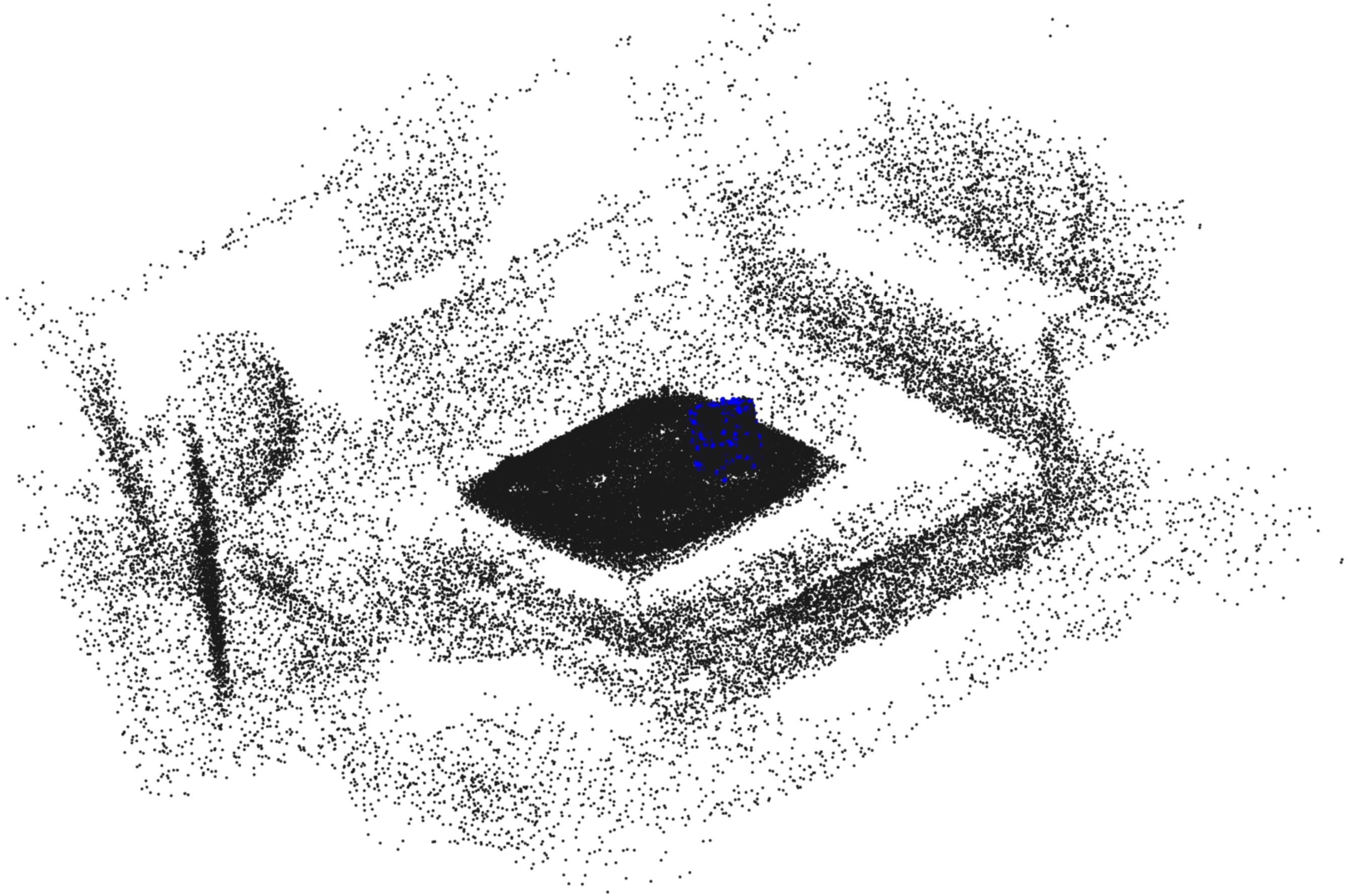}
\end{minipage}
}%

\subfigure[\textit{Scene-05}, Known Scale: ($305,\,95.41\%,\,{2.27\times10^{-5}}^{\circ},4.81\times10^{-7} m$), Unknown Scale: ($571,\,92.82\%,\,5.99\times10^{-7},{0.1161}^{\circ},0.0026 m$)]{
\begin{minipage}[t]{1\linewidth}
\centering
\includegraphics[width=0.24\linewidth]{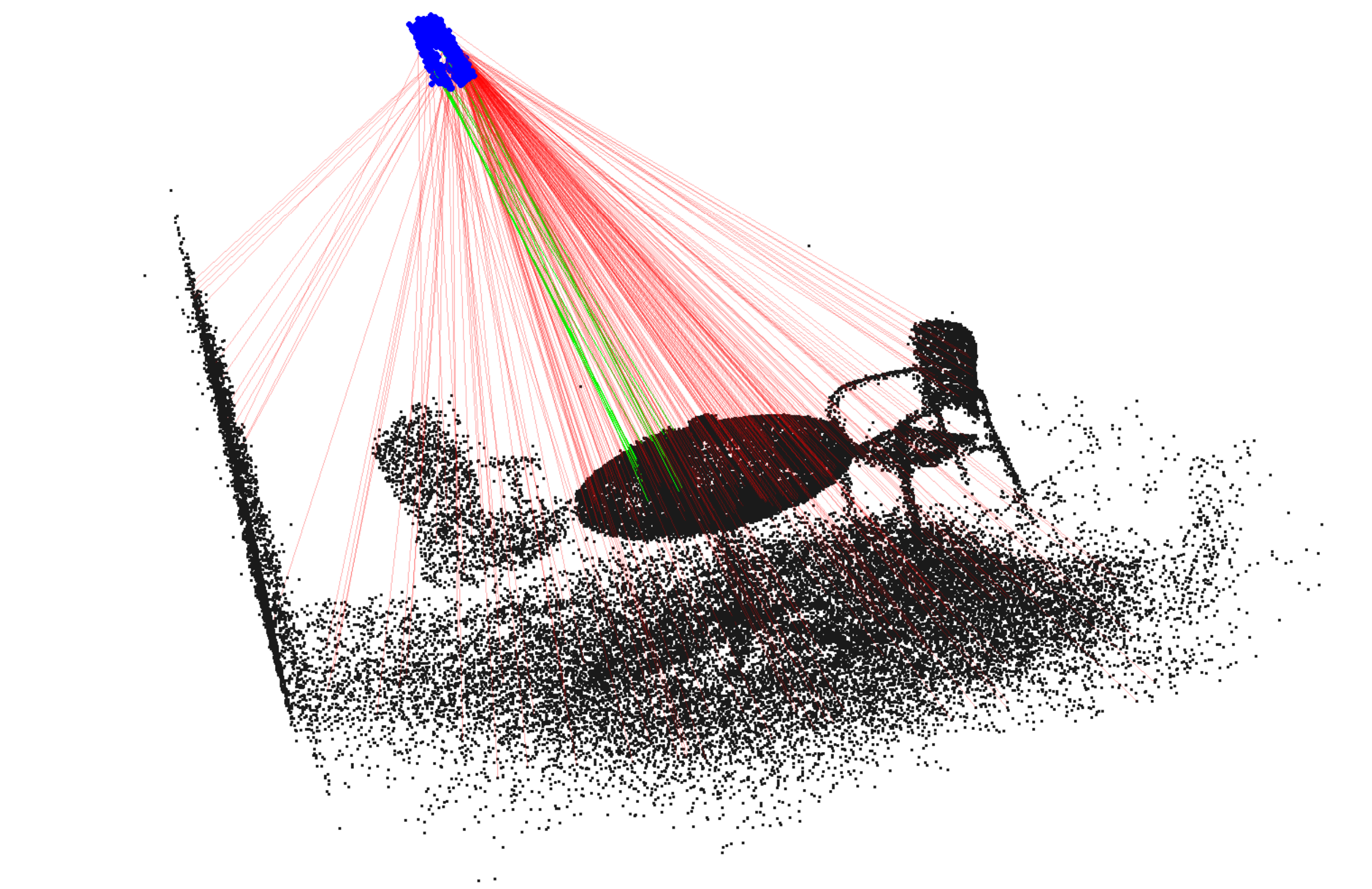}
\includegraphics[width=0.24\linewidth]{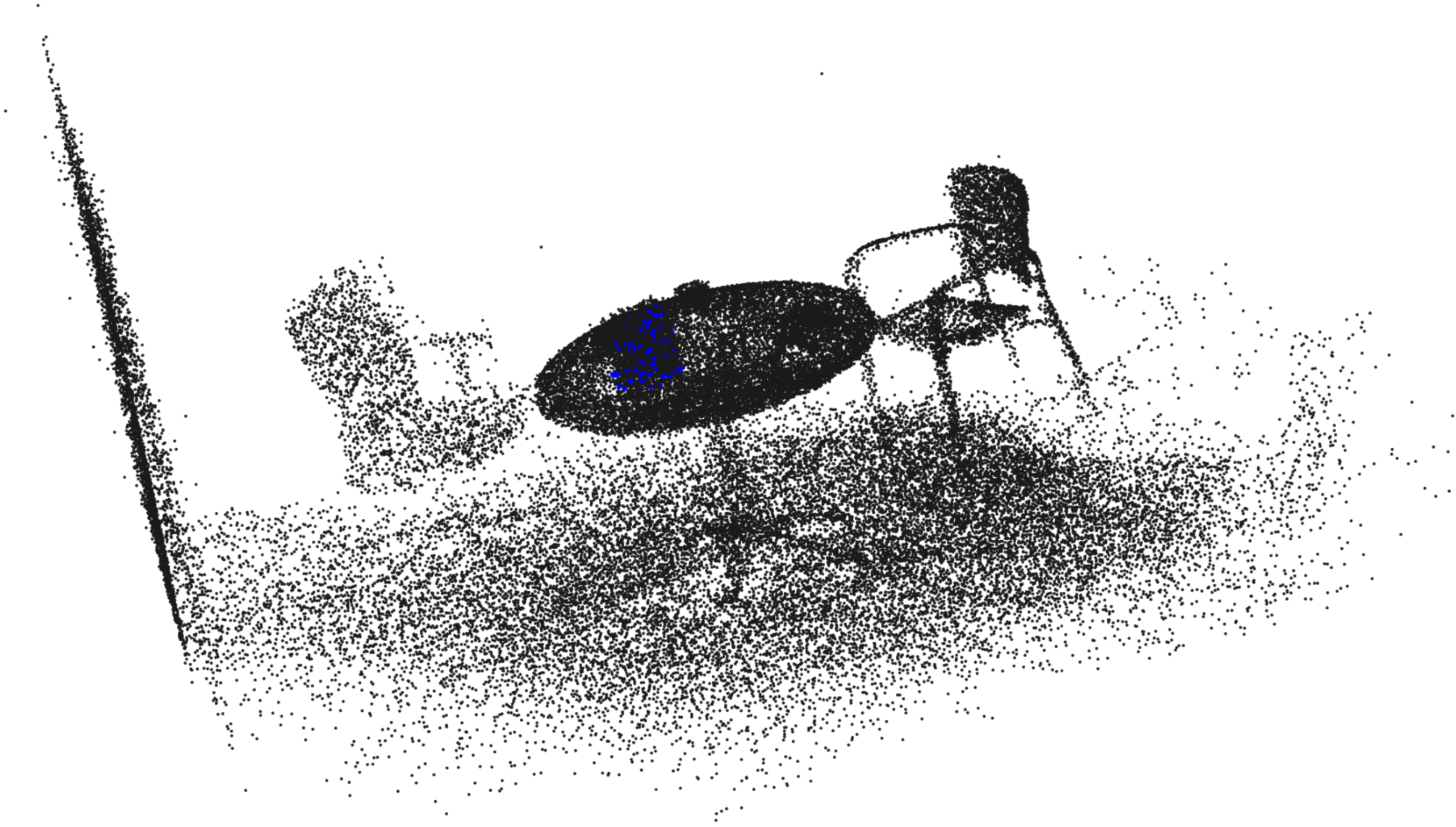}
\includegraphics[width=0.24\linewidth]{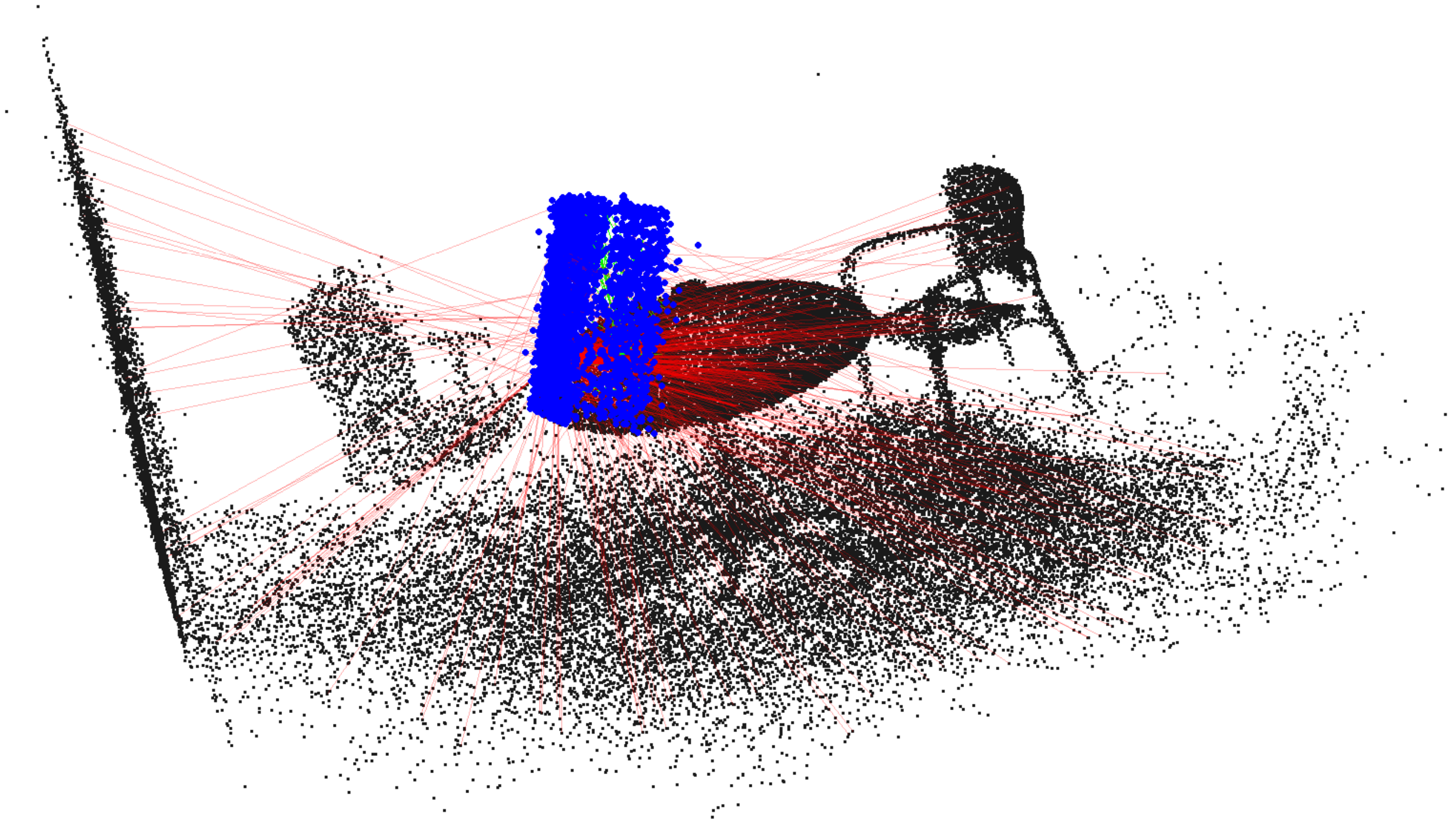}
\includegraphics[width=0.24\linewidth]{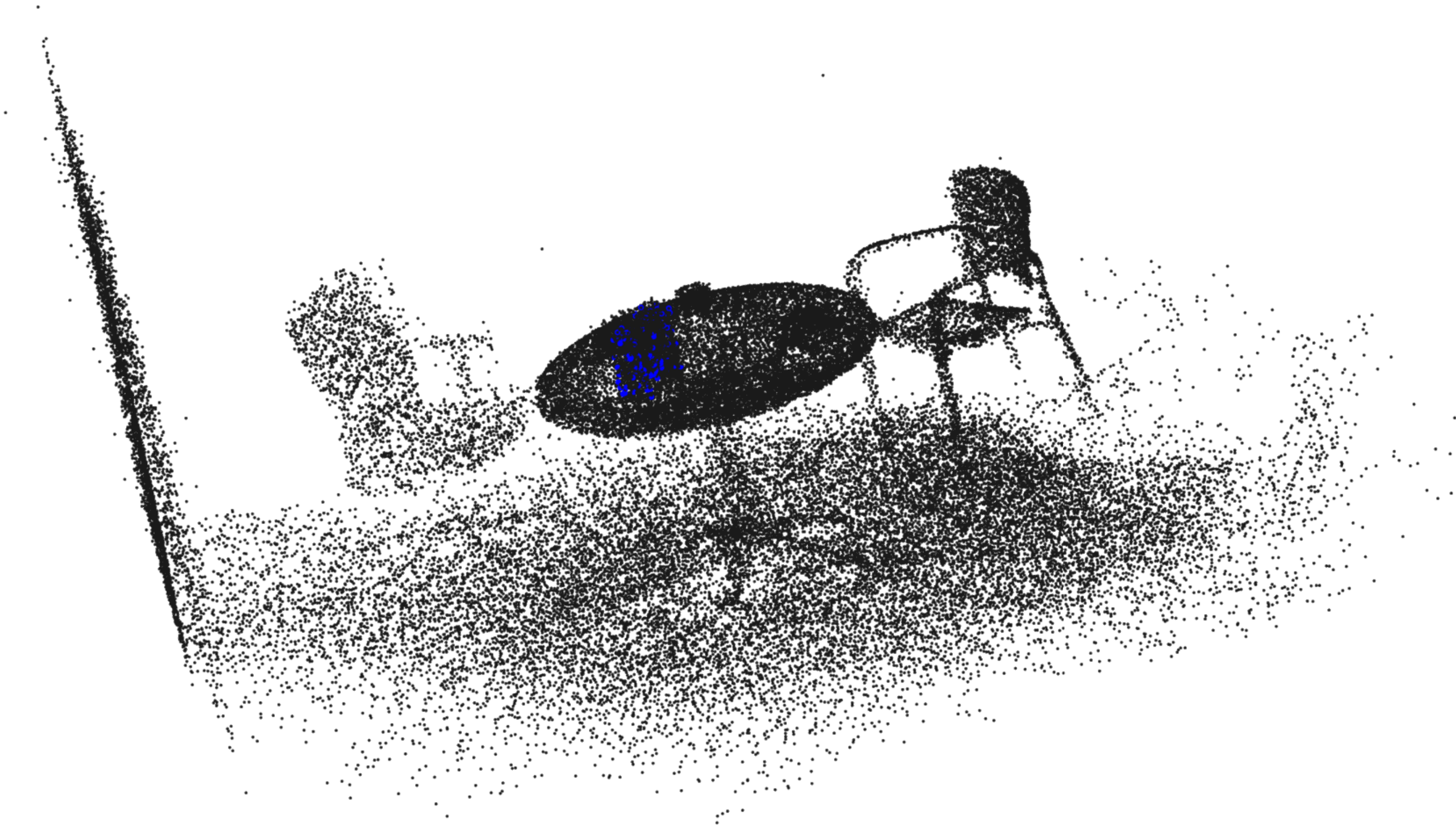}
\end{minipage}
}%

\subfigure[\textit{Scene-07}, Known Scale: ($206,\,93.69\%,\,{9.81\times10^{-6}}^{\circ},1.63\times10^{-7} m$), Unknown Scale: ($191,\,92.15\%,\,4.67\times10^{-7},{0.0800}^{\circ},0.0010 m$)]{
\begin{minipage}[t]{1\linewidth}
\centering
\includegraphics[width=0.24\linewidth]{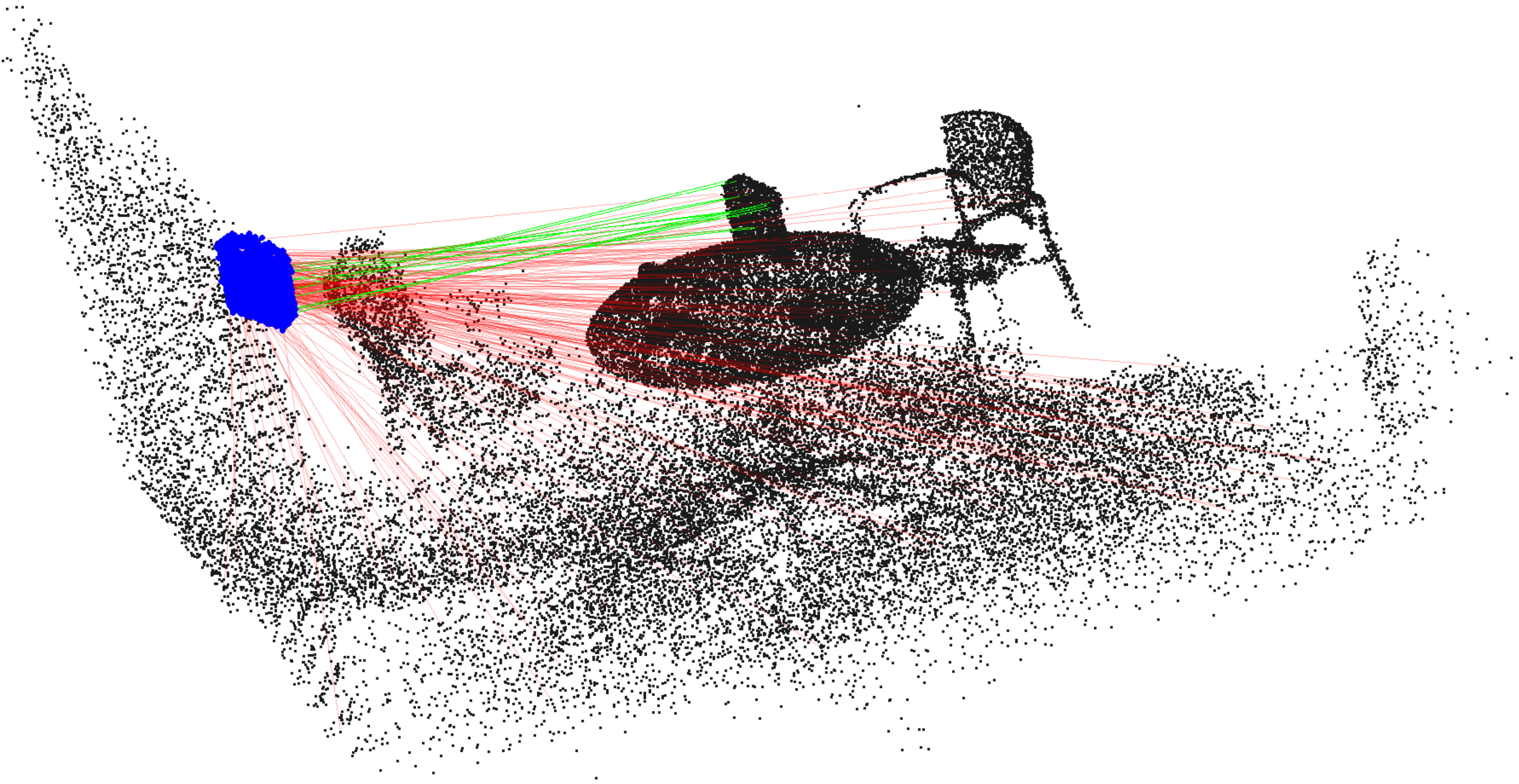}
\includegraphics[width=0.24\linewidth]{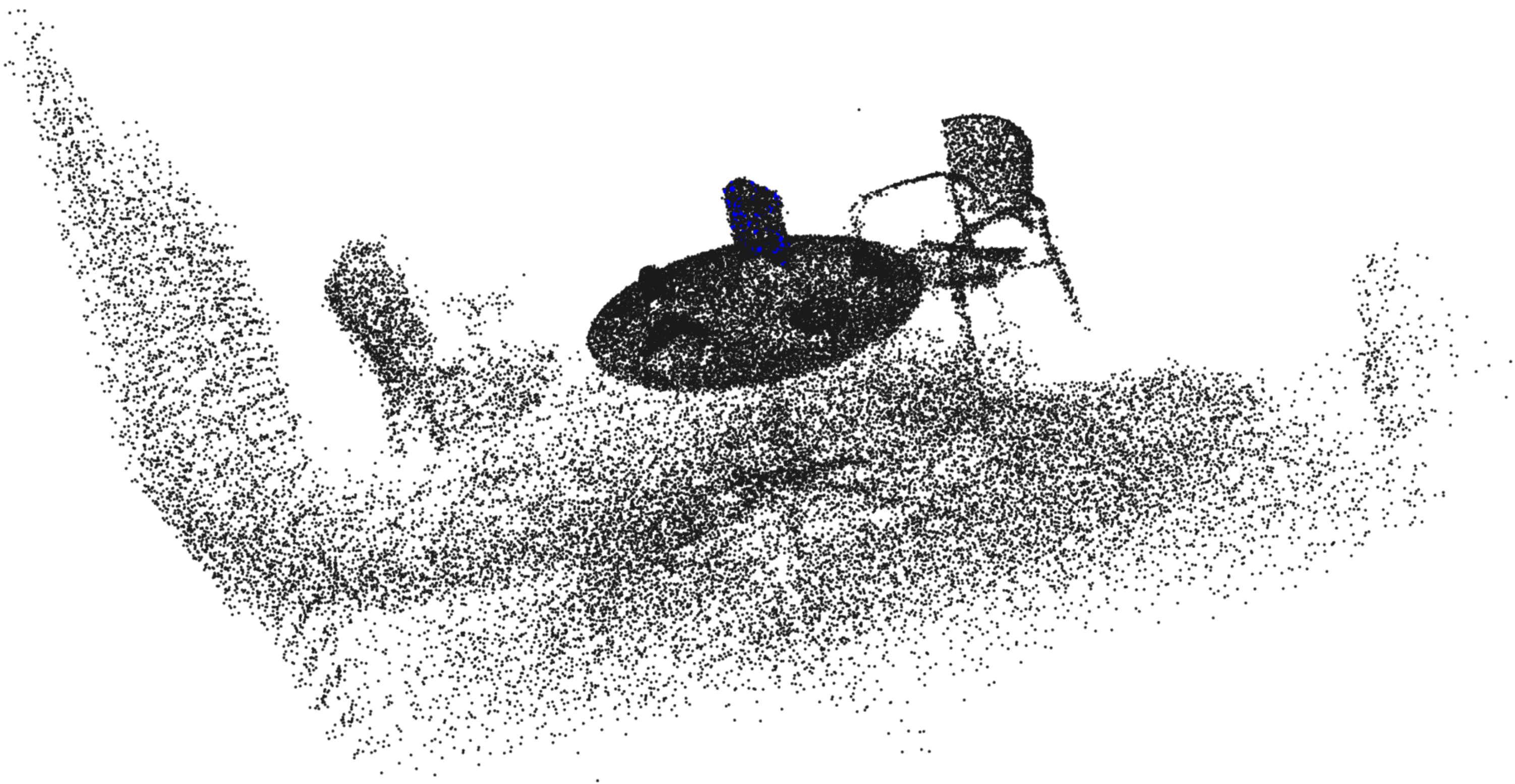}
\includegraphics[width=0.24\linewidth]{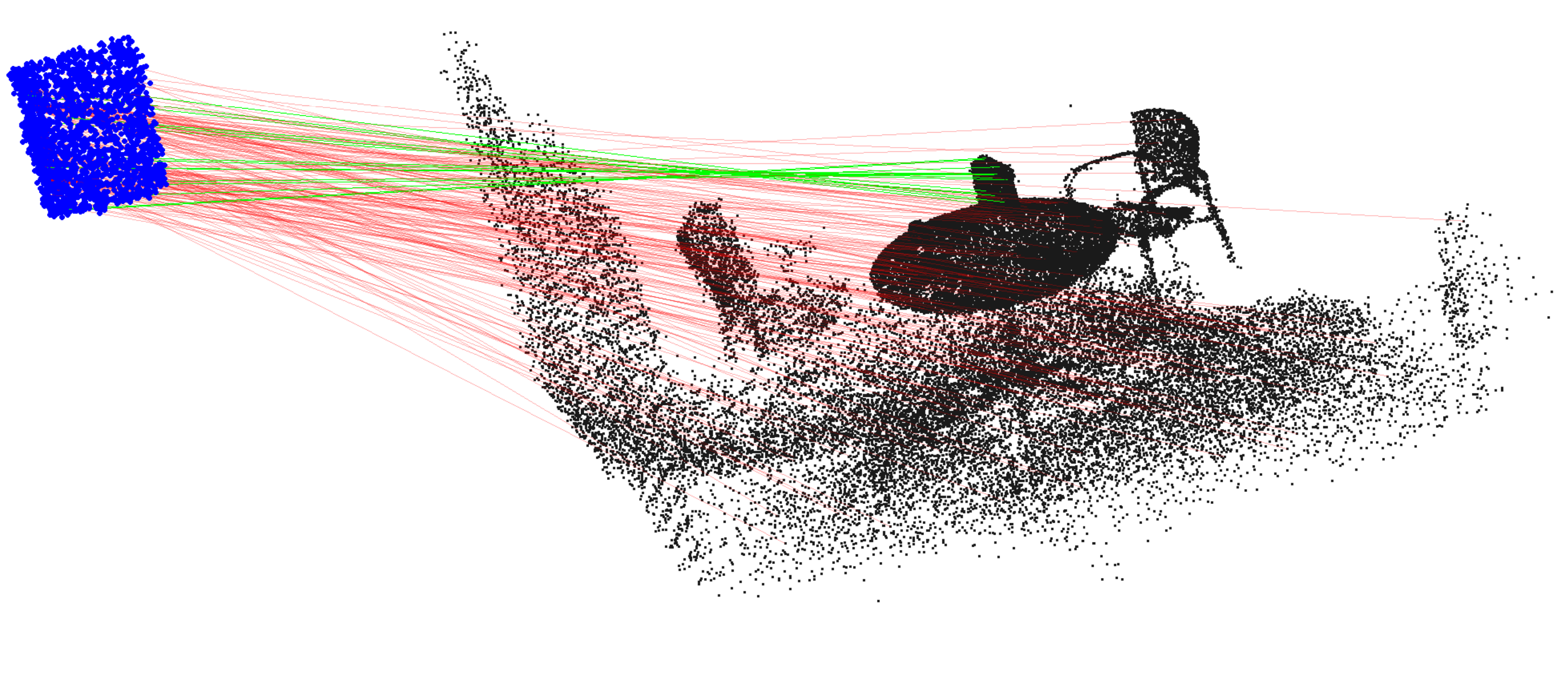}
\includegraphics[width=0.24\linewidth]{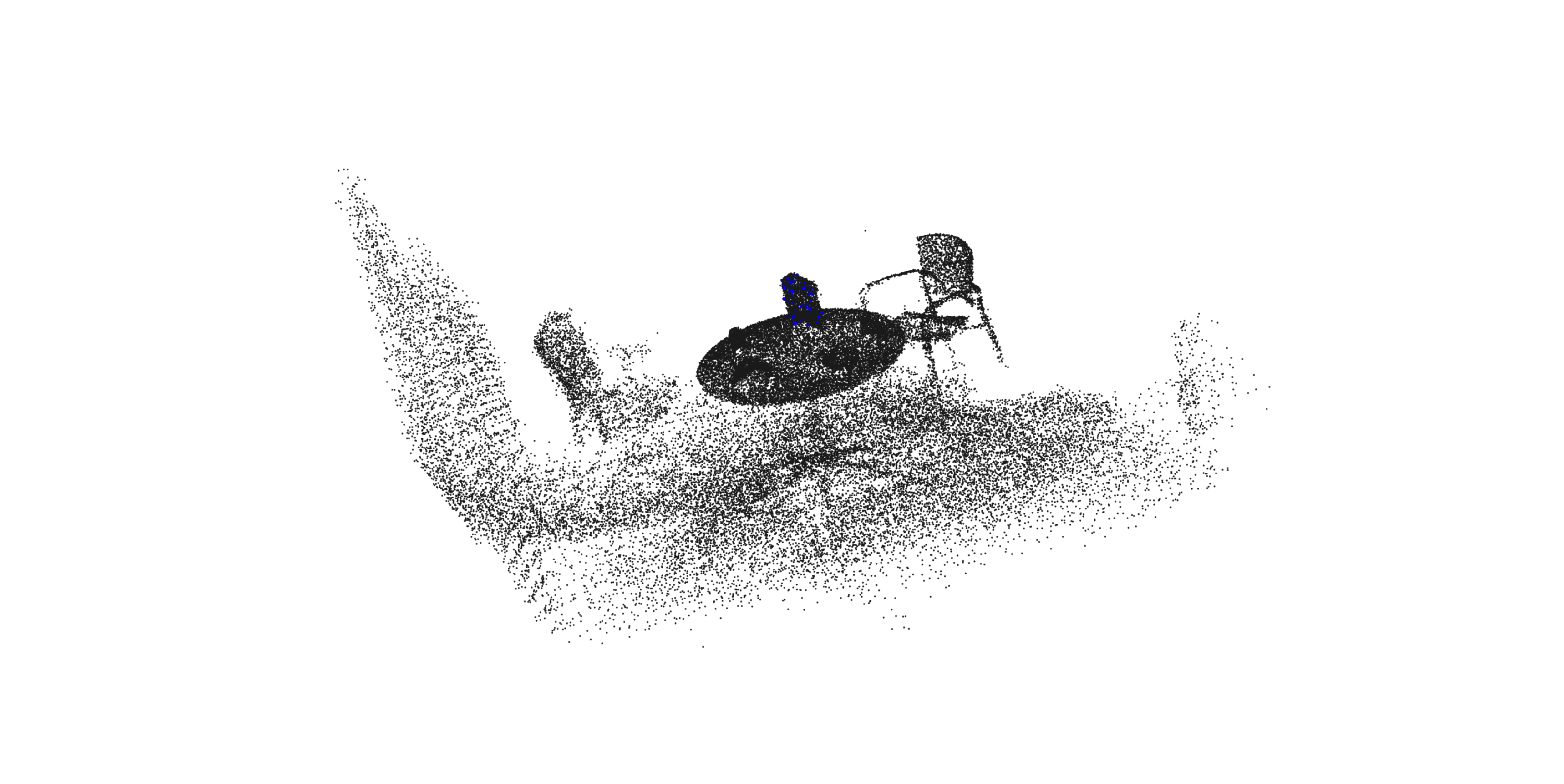}
\end{minipage}
}%

\subfigure[\textit{Scene-09}, Known Scale: ($577,\,90.29\%,\,{6.03\times10^{-6}}^{\circ},1.28\times10^{-7} m$), Unknown Scale: ($684,\,93.84\%,\,2.23\times10^{-7},{0.2366}^{\circ},0.0085 m$)]{
\begin{minipage}[t]{1\linewidth}
\centering
\includegraphics[width=0.24\linewidth]{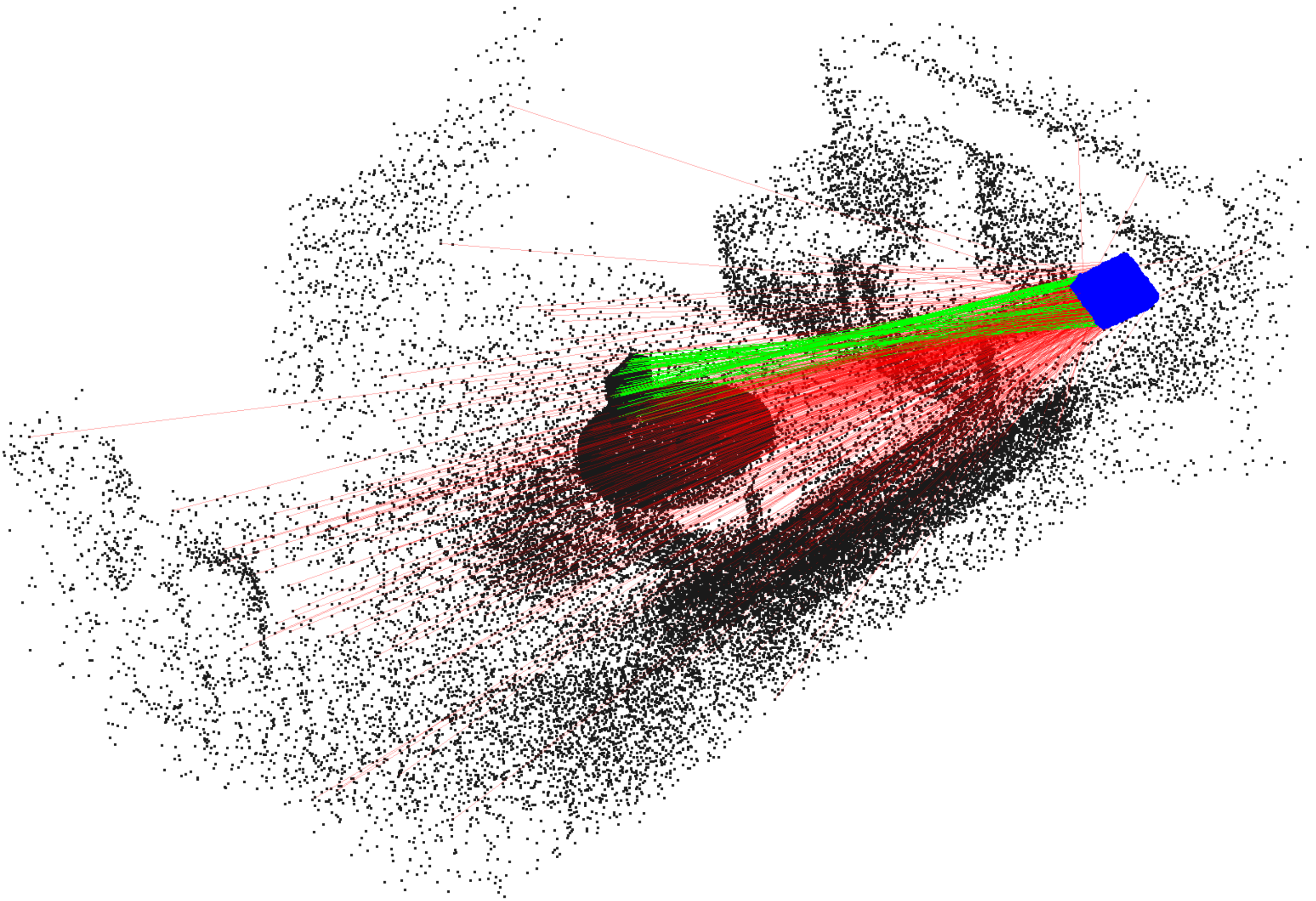}
\includegraphics[width=0.24\linewidth]{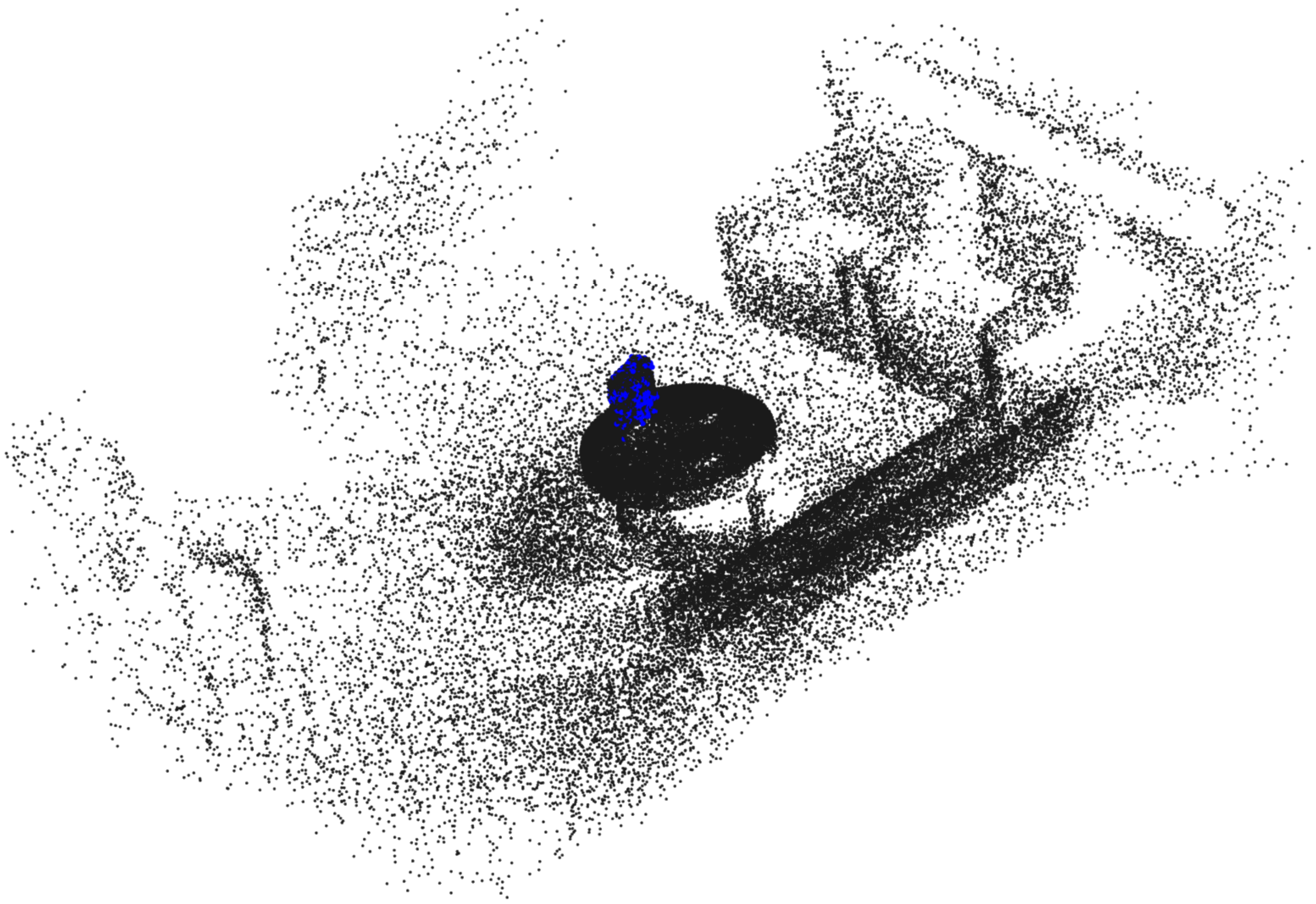}
\includegraphics[width=0.24\linewidth]{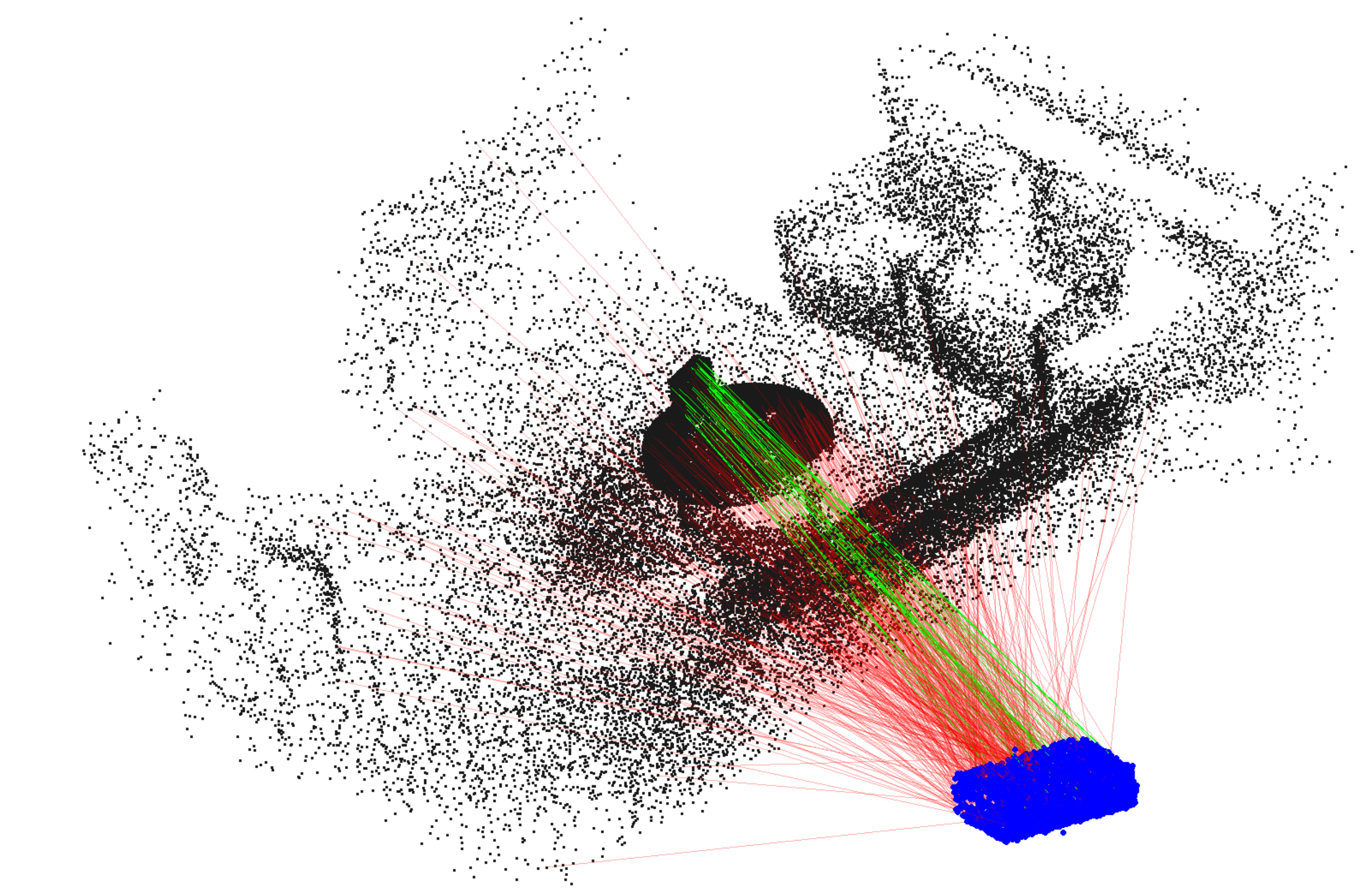}
\includegraphics[width=0.24\linewidth]{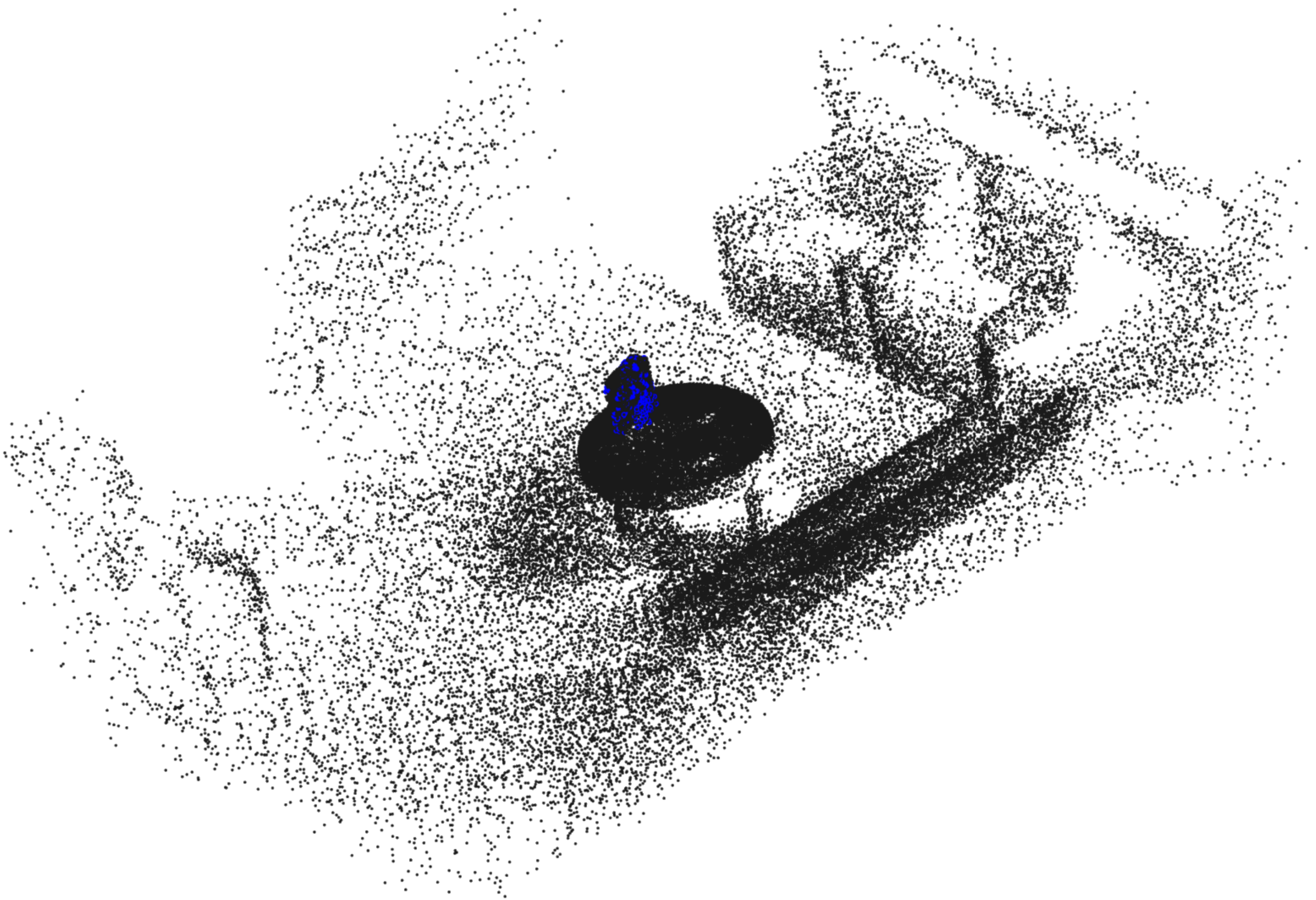}
\end{minipage}
}%

\subfigure[\textit{Scene-11}, Known Scale: ($678,98.53\%,{2.17\times10^{-5}}^{\circ},2.81\times10^{-7}m$), Unknown Scale: ($457,95.62\%,5.06\times10^{-8},{1.70\times10^{-5}}^{\circ},7.32\times10^{-7} m$)]{
\begin{minipage}[t]{1\linewidth}
\centering
\includegraphics[width=0.24\linewidth]{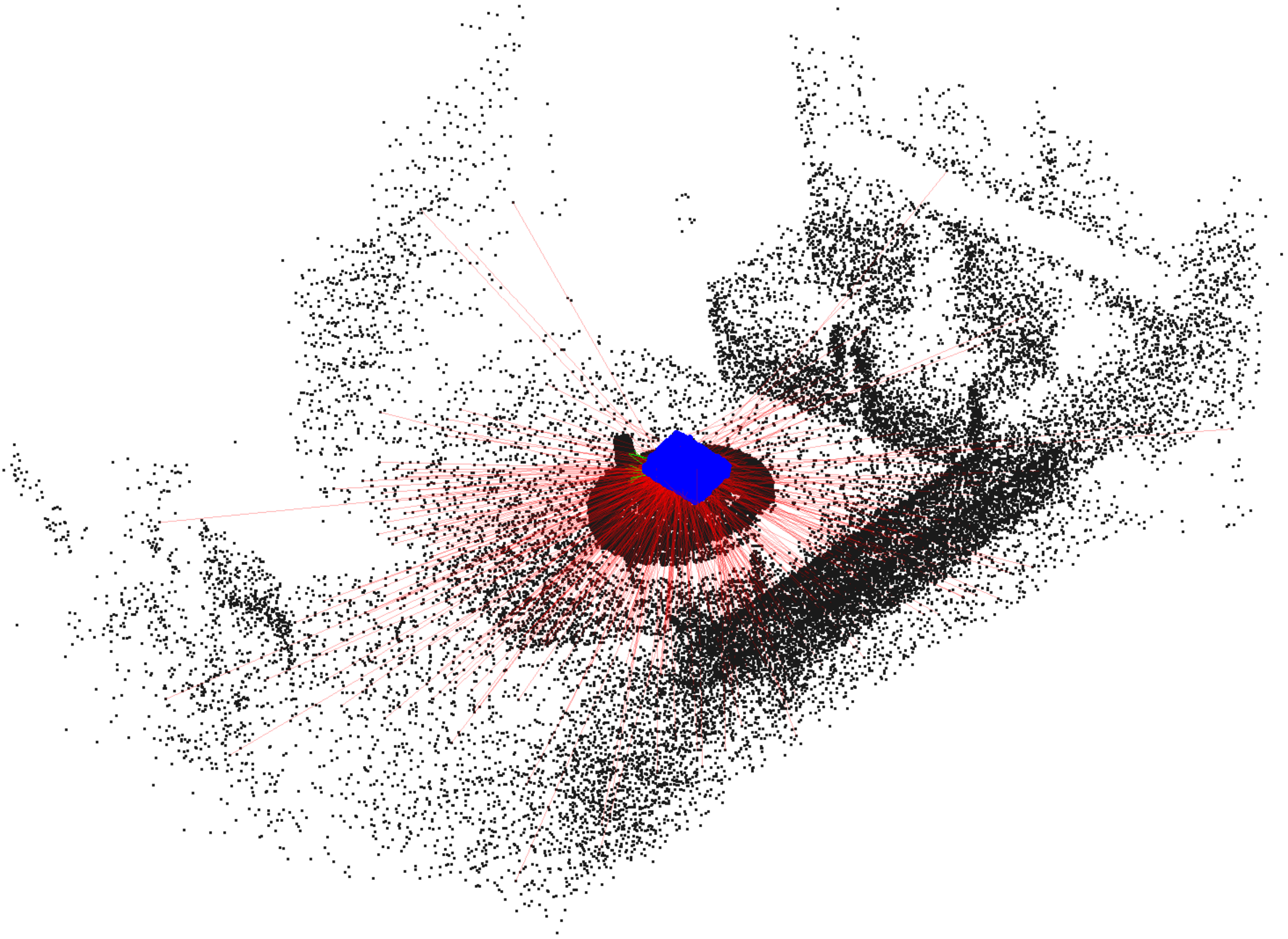}
\includegraphics[width=0.24\linewidth]{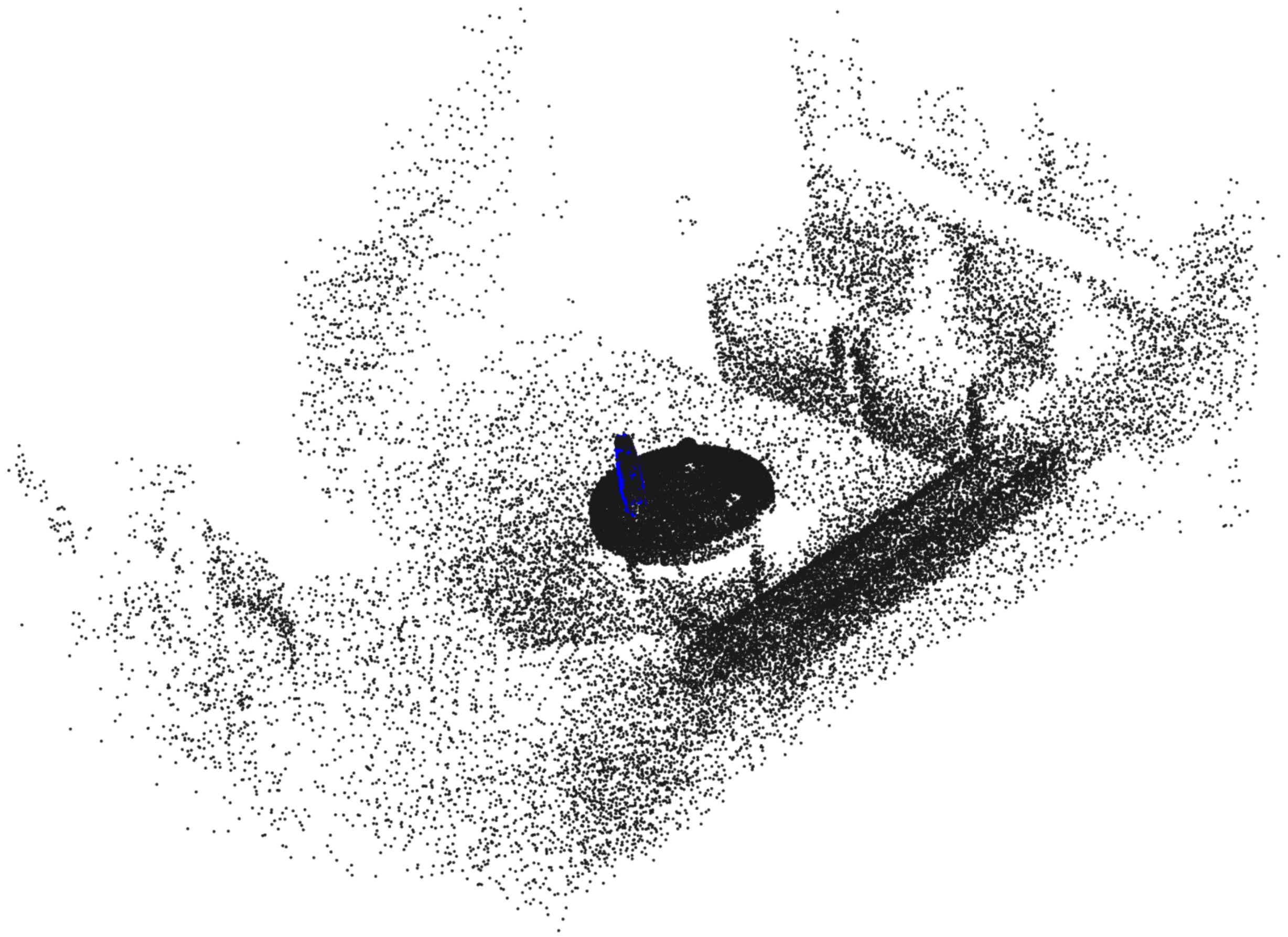}
\includegraphics[width=0.24\linewidth]{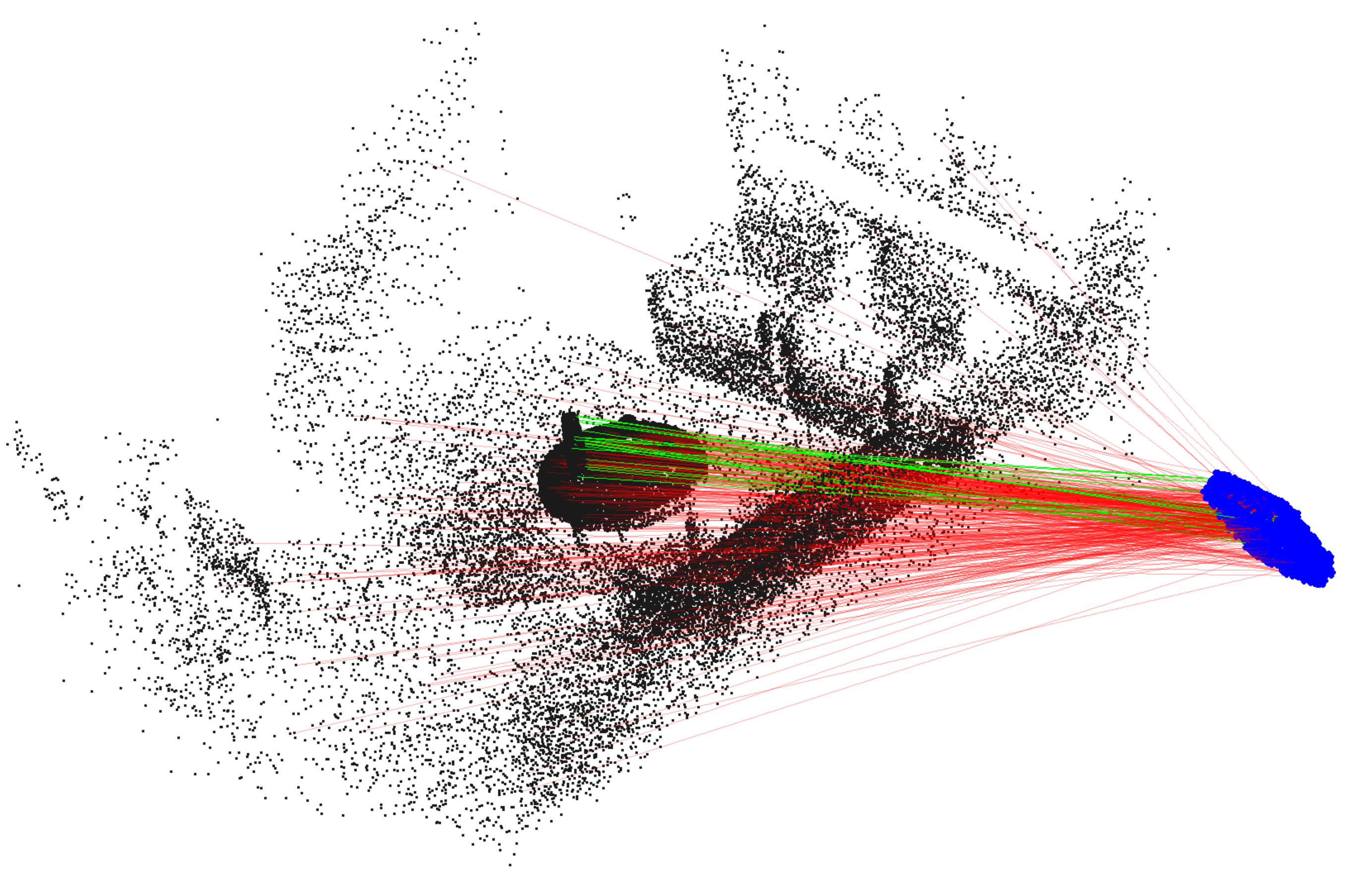}
\includegraphics[width=0.24\linewidth]{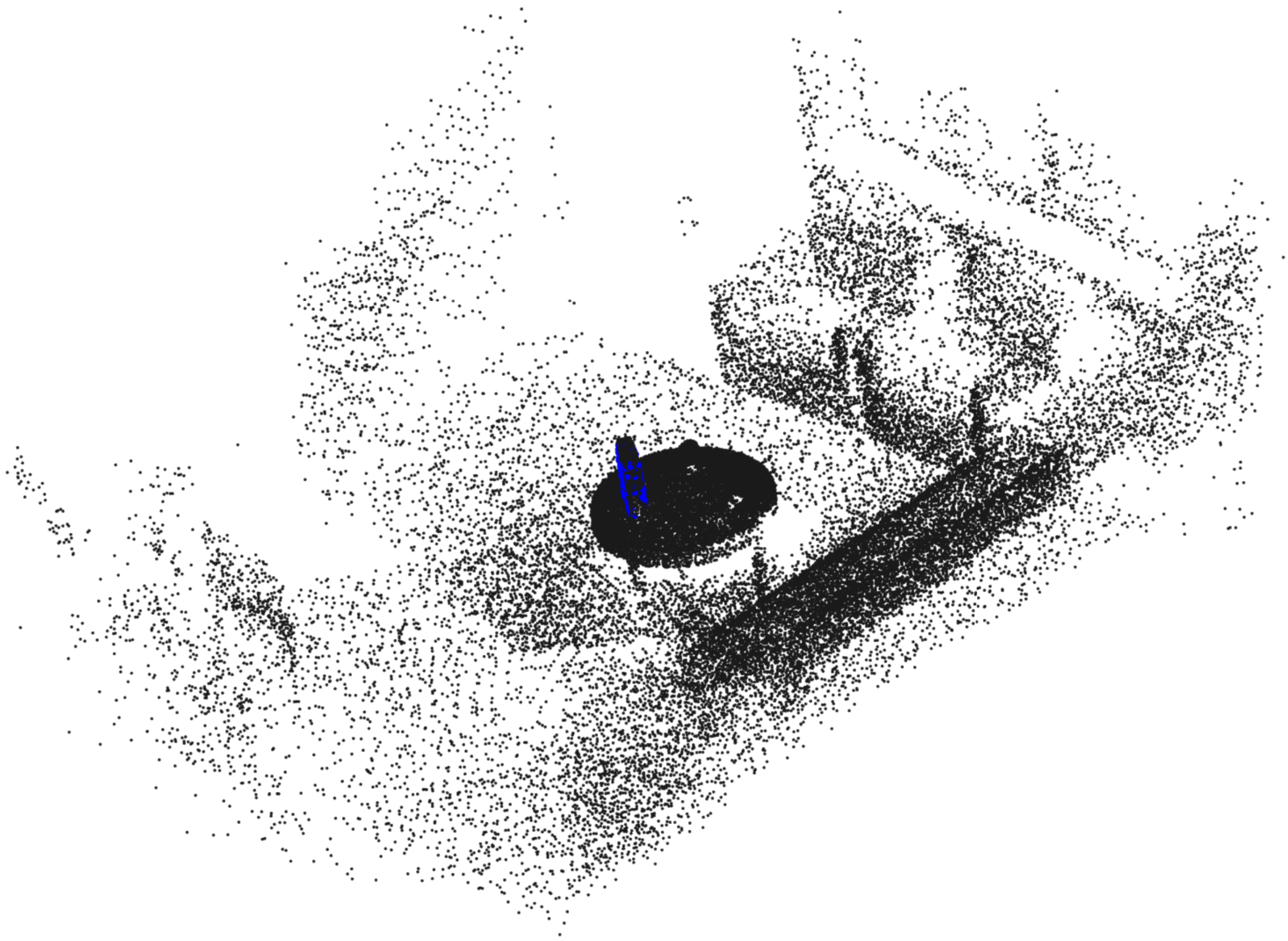}
\end{minipage}
}%

\centering
\caption{3D object localization of \textit{cereal box} over dataset~\cite{lai2011large}. In each scene from left to right, the images show: correspondences for known-scale registration, qualitative reprojection result of known-scale registration, correspondences for unknown-scale registration, and qualitative reprojection result of unknown-scale registration, and the data below denote: correspondence number, outlier ratio, (scale error), rotation error, and translation error.}
\label{CB}
\vspace{-9pt}
\end{figure*}

\begin{figure*}[t]
\centering

\subfigure[\textit{Scene-01}, Known Scale: ($417,92.09\%,{9.19\times10^{-6}}^{\circ},8.46\times10^{-8}m$), Unknown Scale: ($579,96.55\%,4.61\times10^{-7},{1.55\times10^{-5}}^{\circ},8.39\times10^{-7} m$)]{
\begin{minipage}[t]{1\linewidth}
\centering
\includegraphics[width=0.24\linewidth]{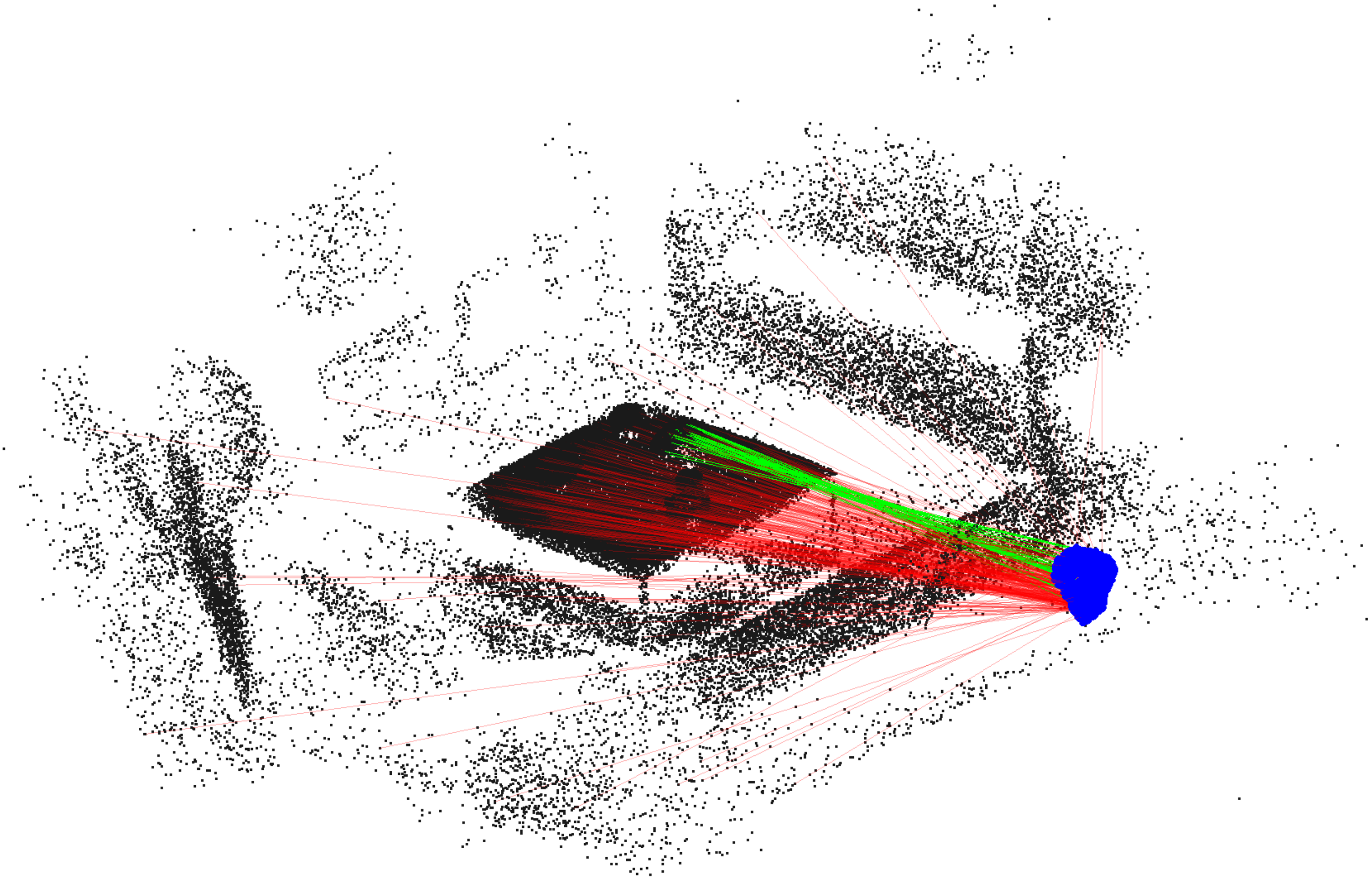}
\includegraphics[width=0.24\linewidth]{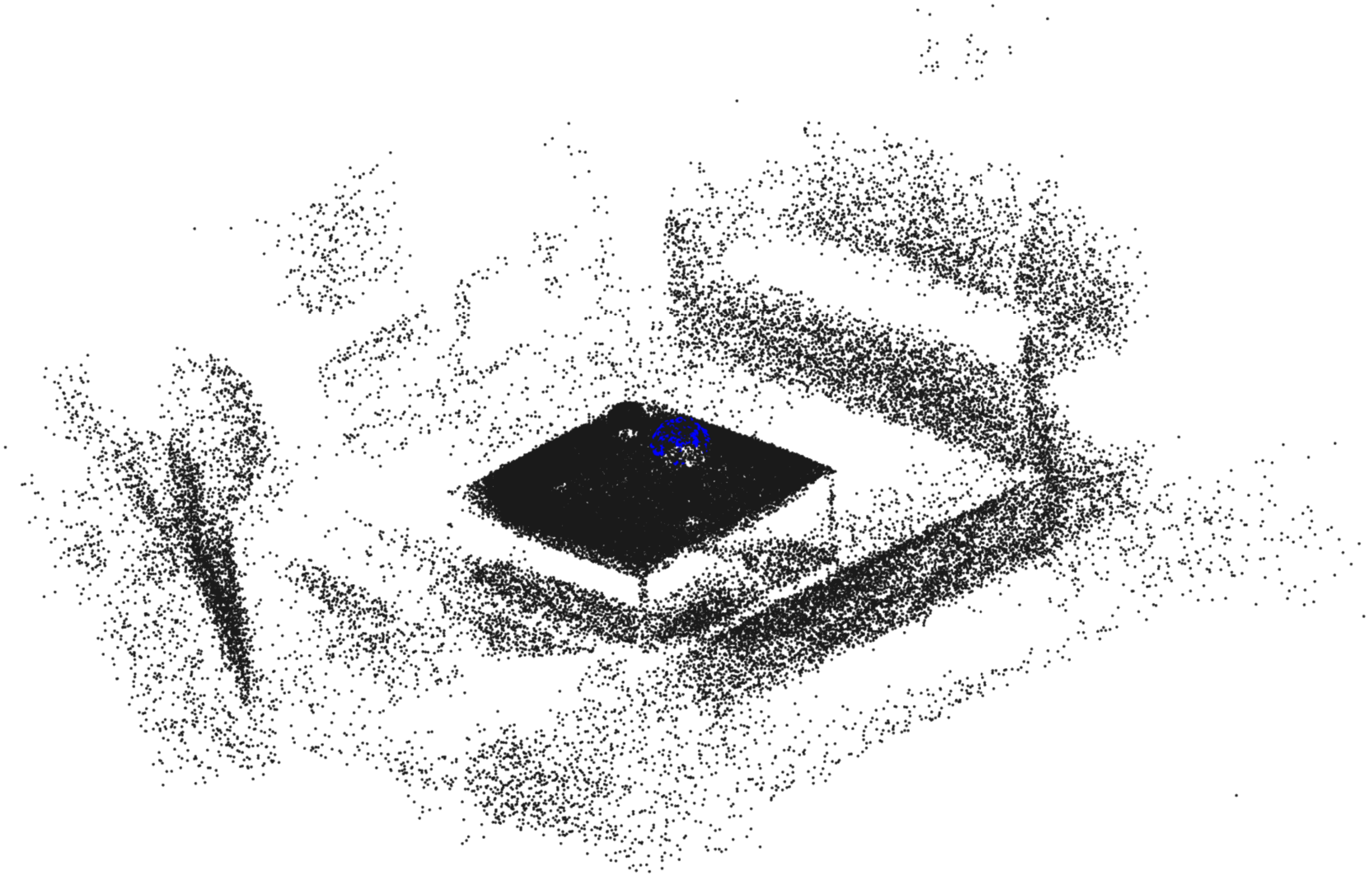}
\includegraphics[width=0.24\linewidth]{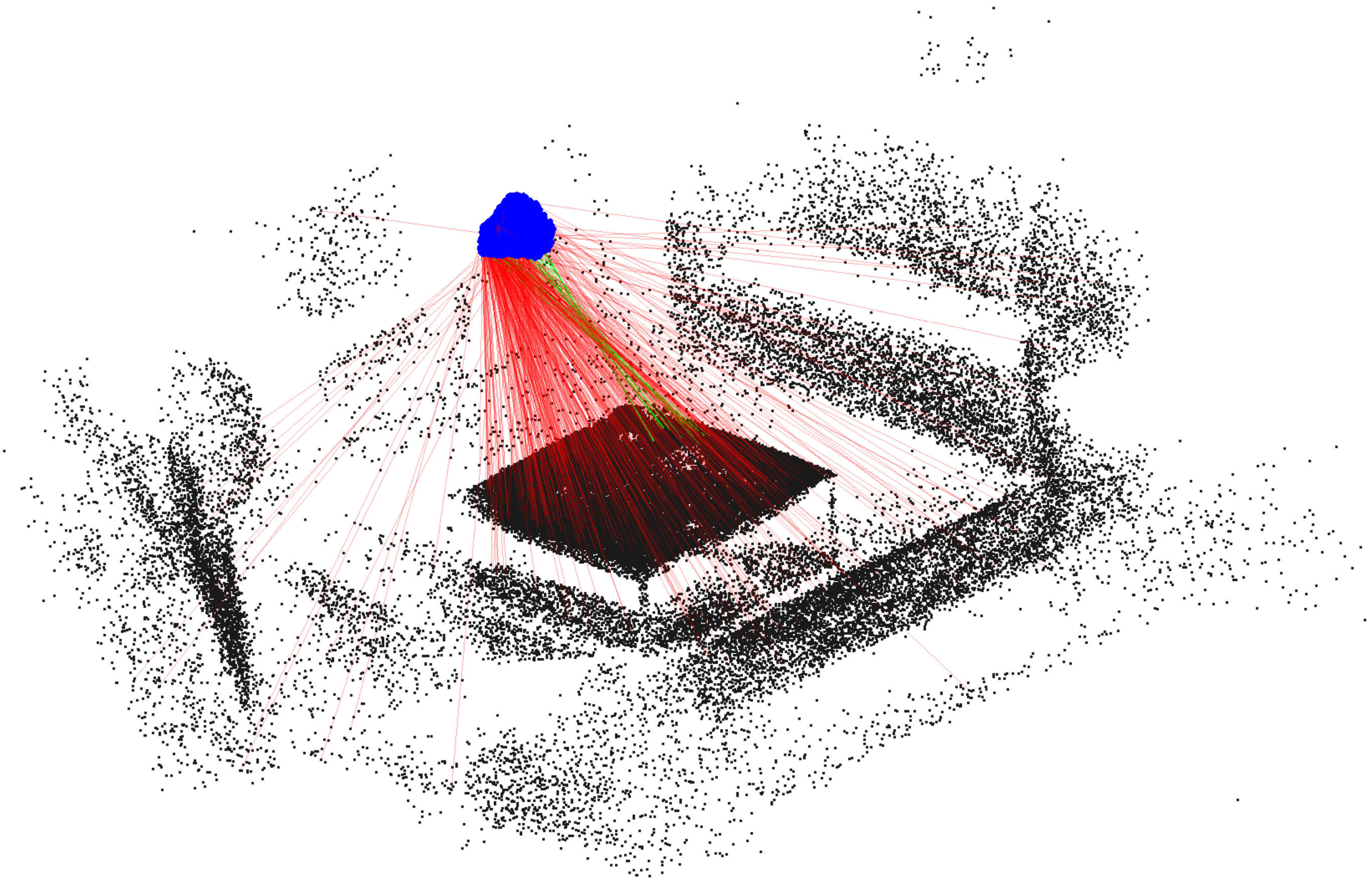}
\includegraphics[width=0.24\linewidth]{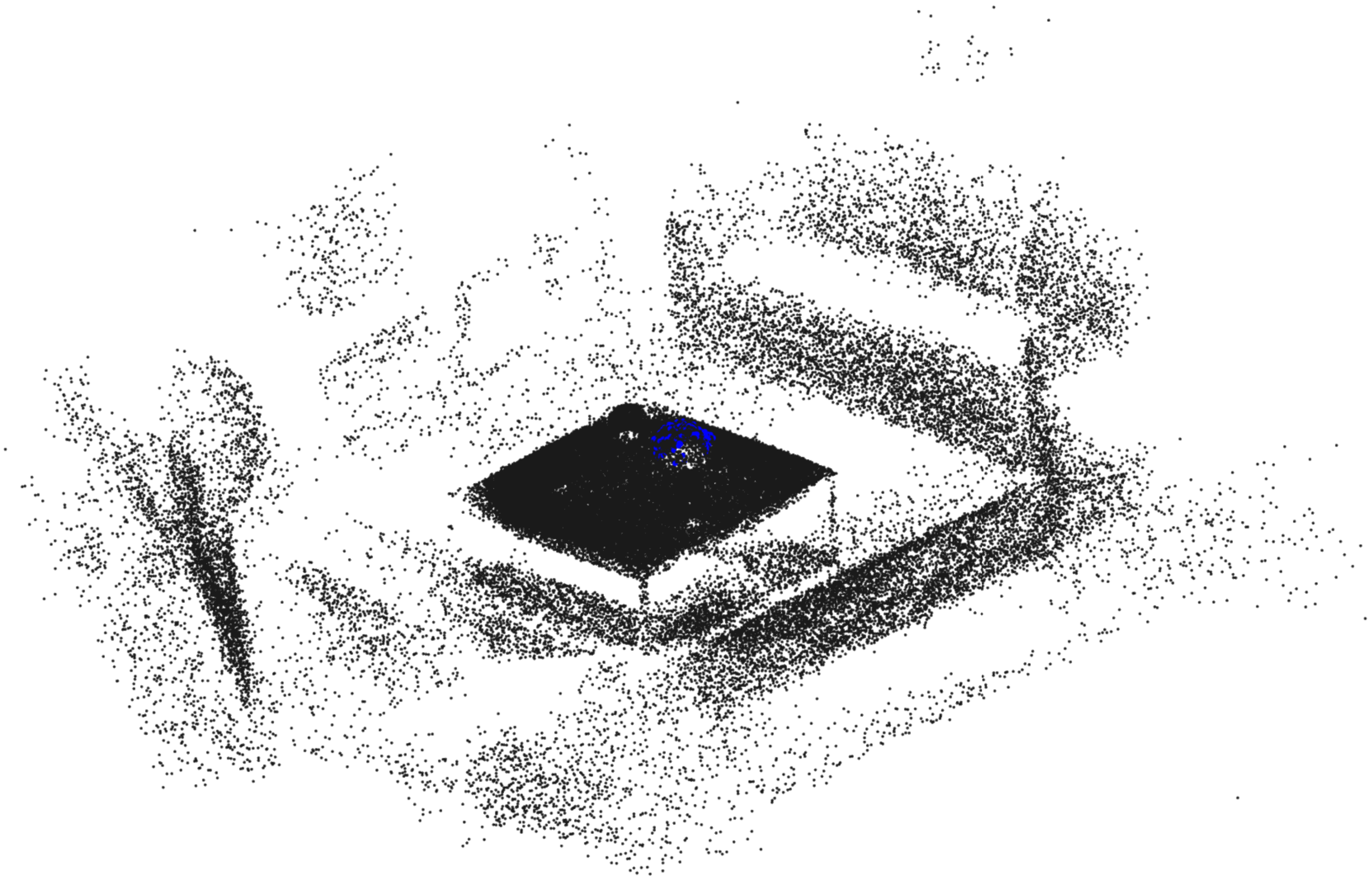}
\end{minipage}
}%

\subfigure[\textit{Scene-03}, Known Scale: ($260,\,91.54\%,\,{0.1950}^{\circ},0.0026 m$), Unknown Scale: ($249,\,91.97\%,\,6.23\times10^{-8},{1.42\times10^{-5}}^{\circ},3.74\times10^{-7} m$)]{
\begin{minipage}[t]{1\linewidth}
\centering
\includegraphics[width=0.24\linewidth]{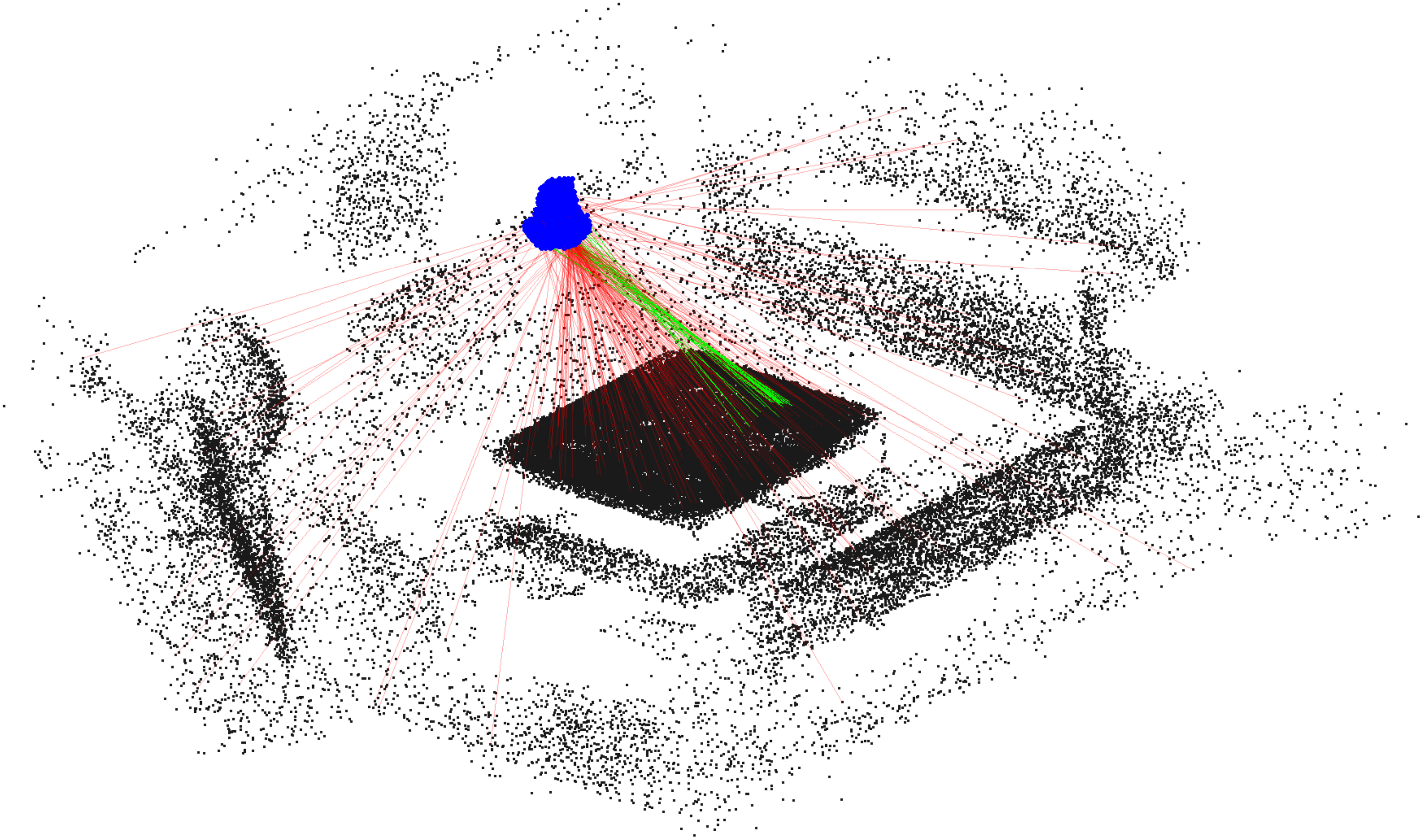}
\includegraphics[width=0.24\linewidth]{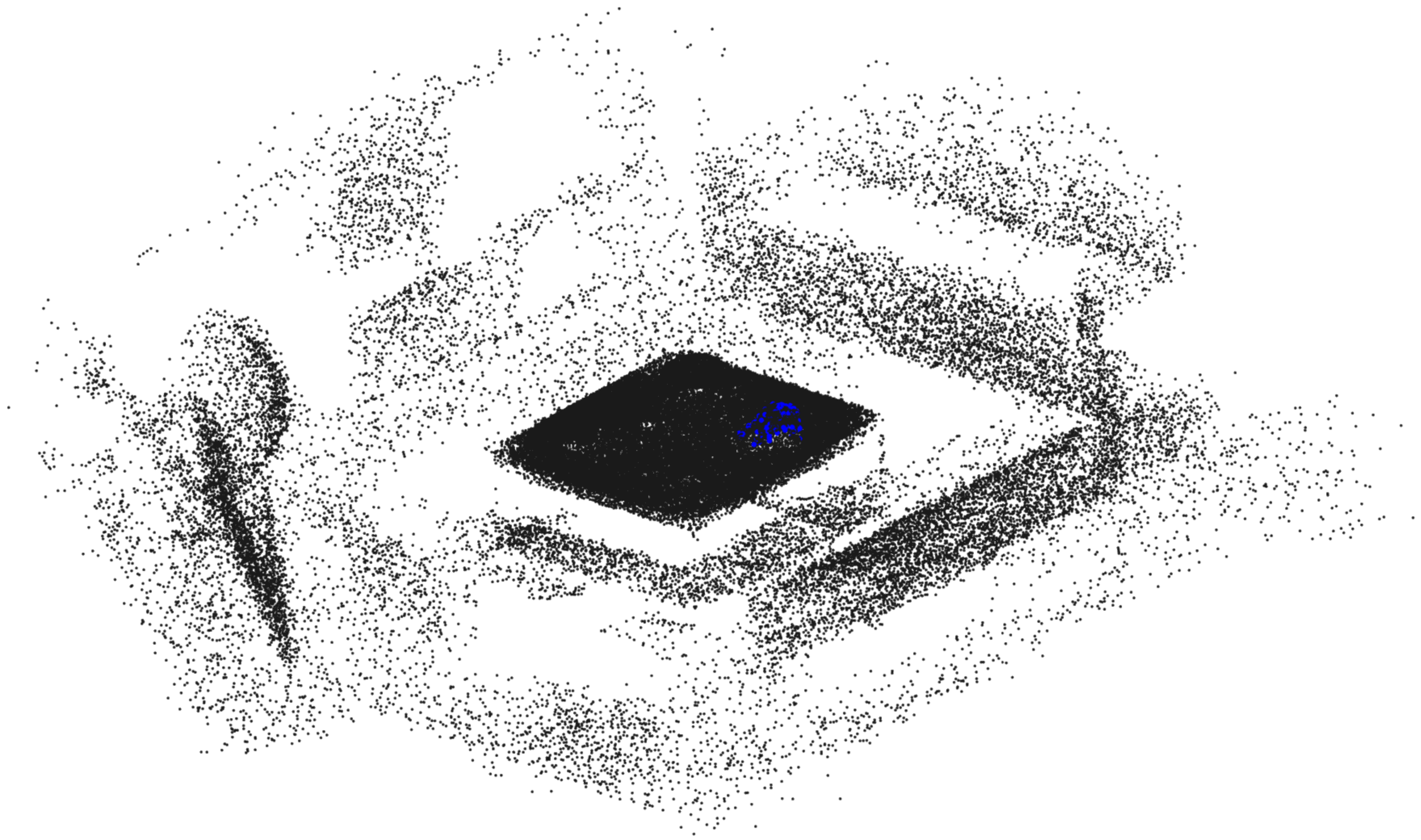}
\includegraphics[width=0.24\linewidth]{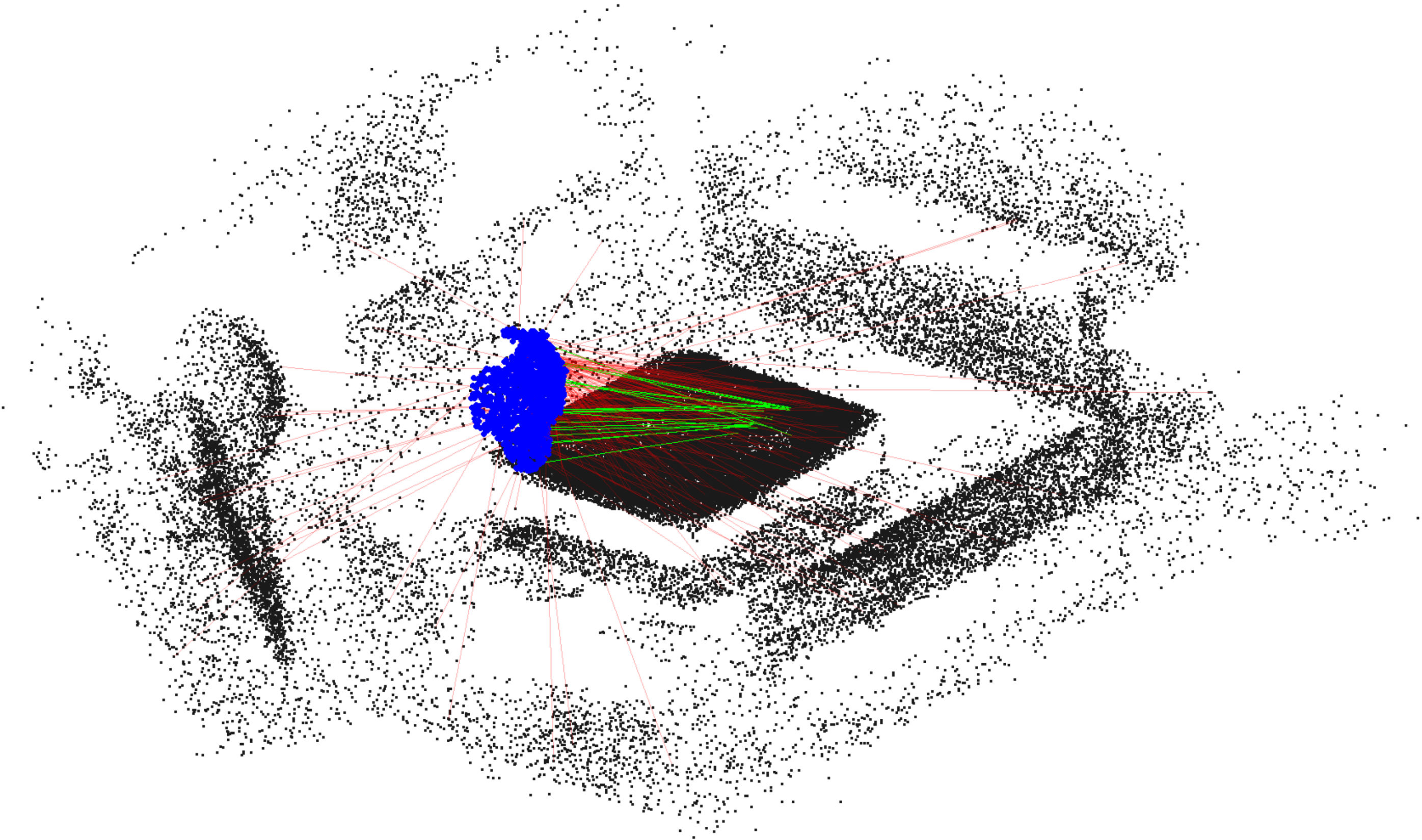}
\includegraphics[width=0.24\linewidth]{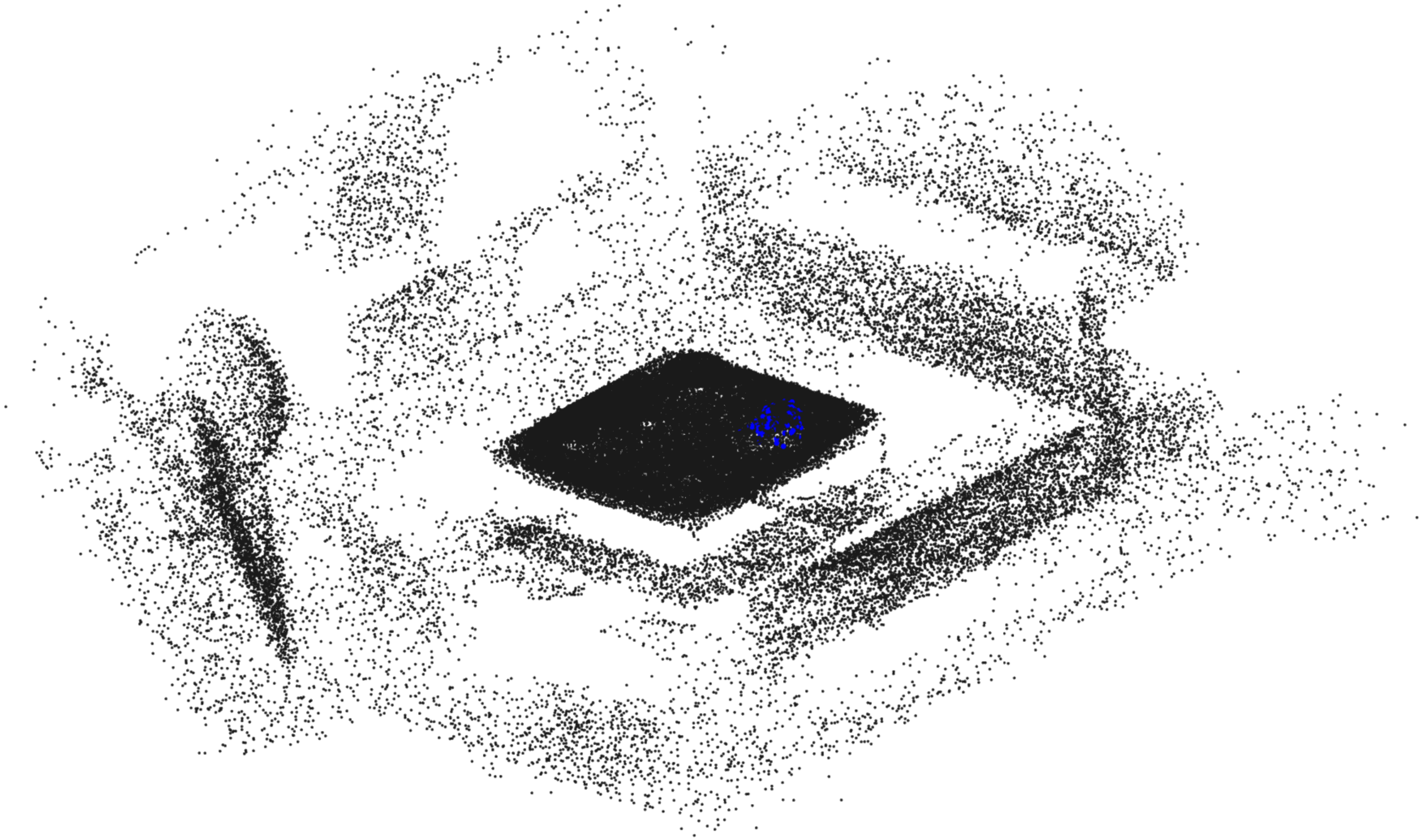}
\end{minipage}
}%

\subfigure[\textit{Scene-10}, Known Scale: ($582,\,91.58\%,\,{0.1894}^{\circ},0.0013 m$), Unknown Scale: ($521,\,91.94\%,\,2.88\times10^{-7},{0.0171}^{\circ},6.79\times10^{-4} m$)]{
\begin{minipage}[t]{1\linewidth}
\centering
\includegraphics[width=0.24\linewidth]{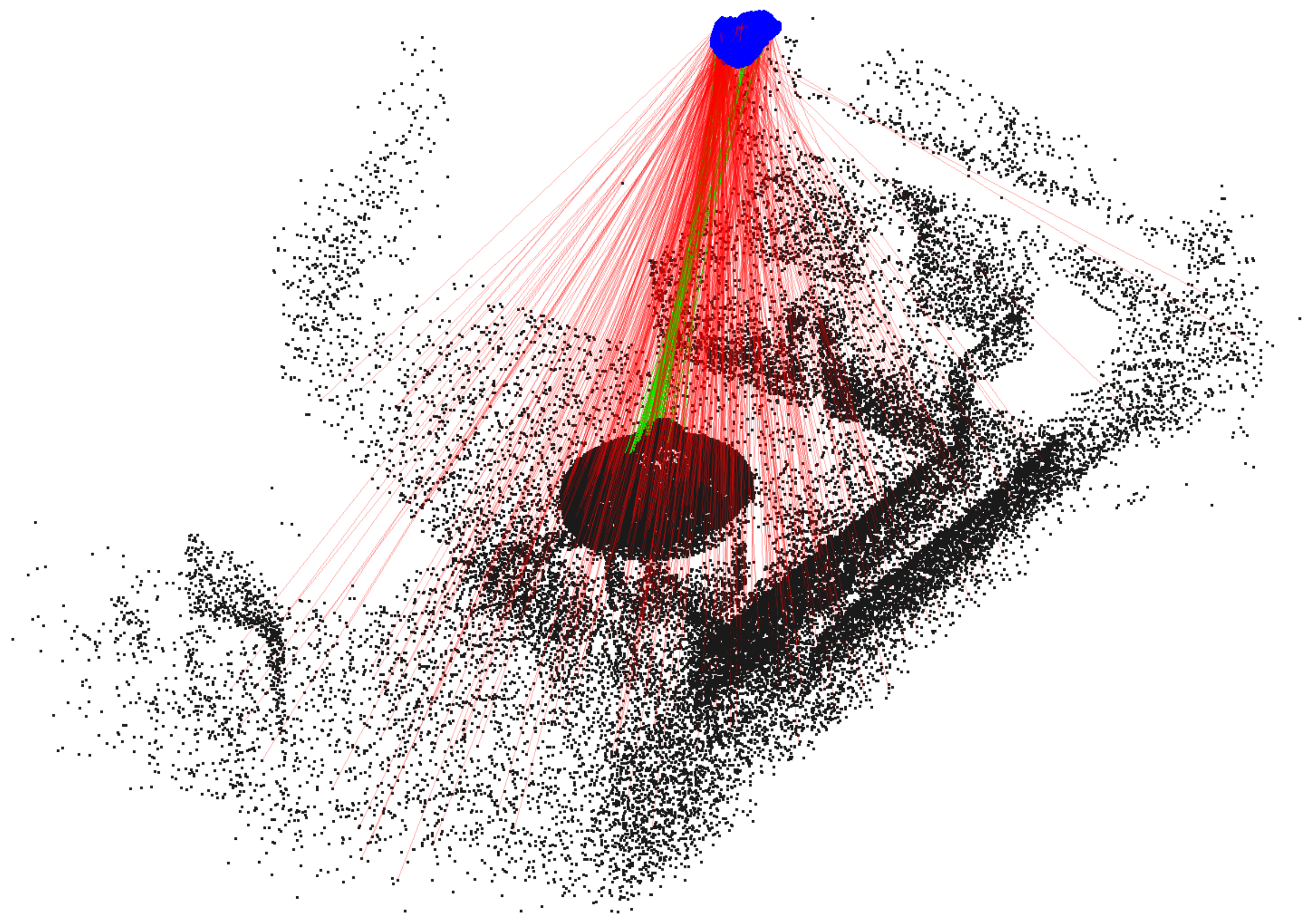}
\includegraphics[width=0.24\linewidth]{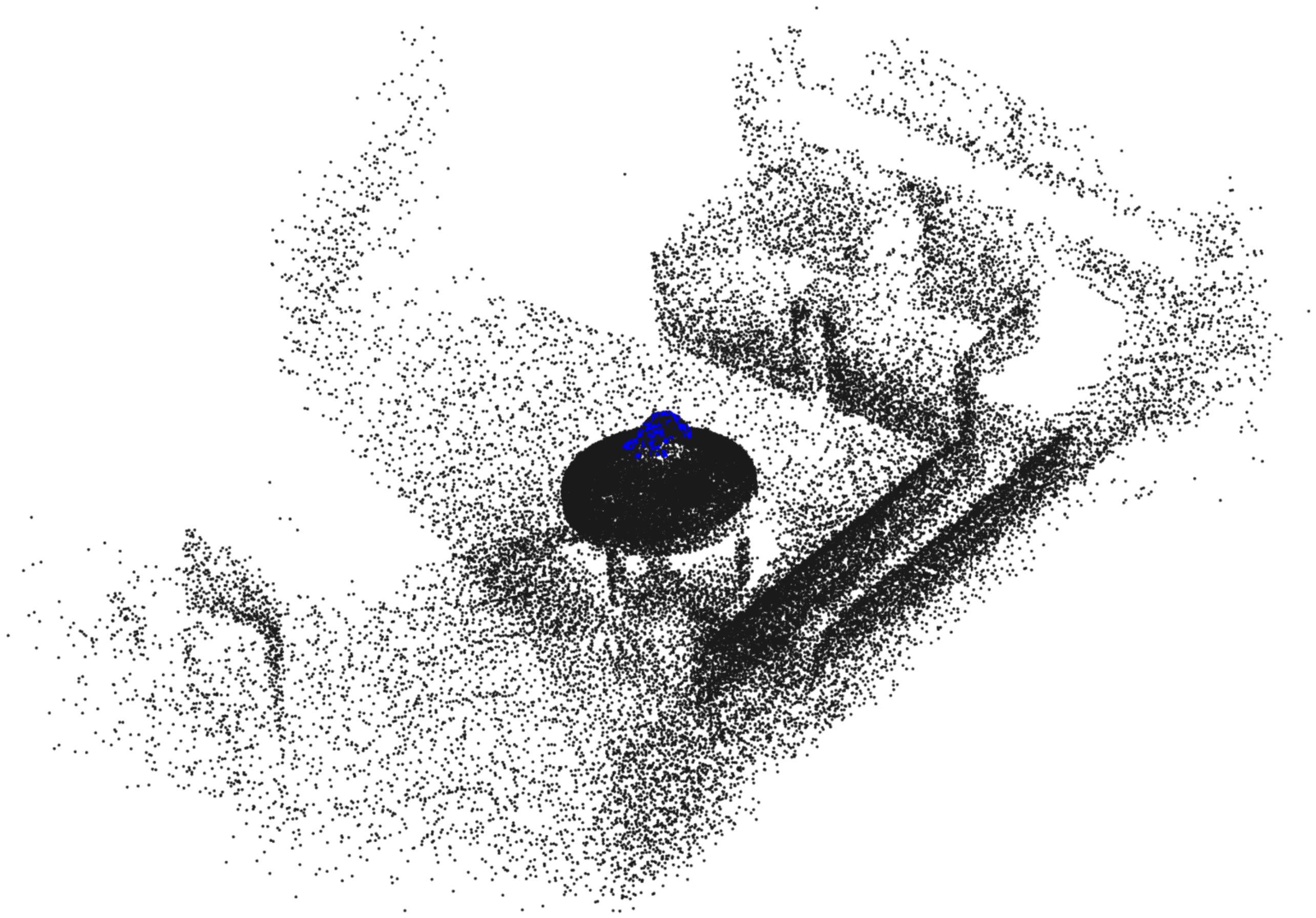}
\includegraphics[width=0.24\linewidth]{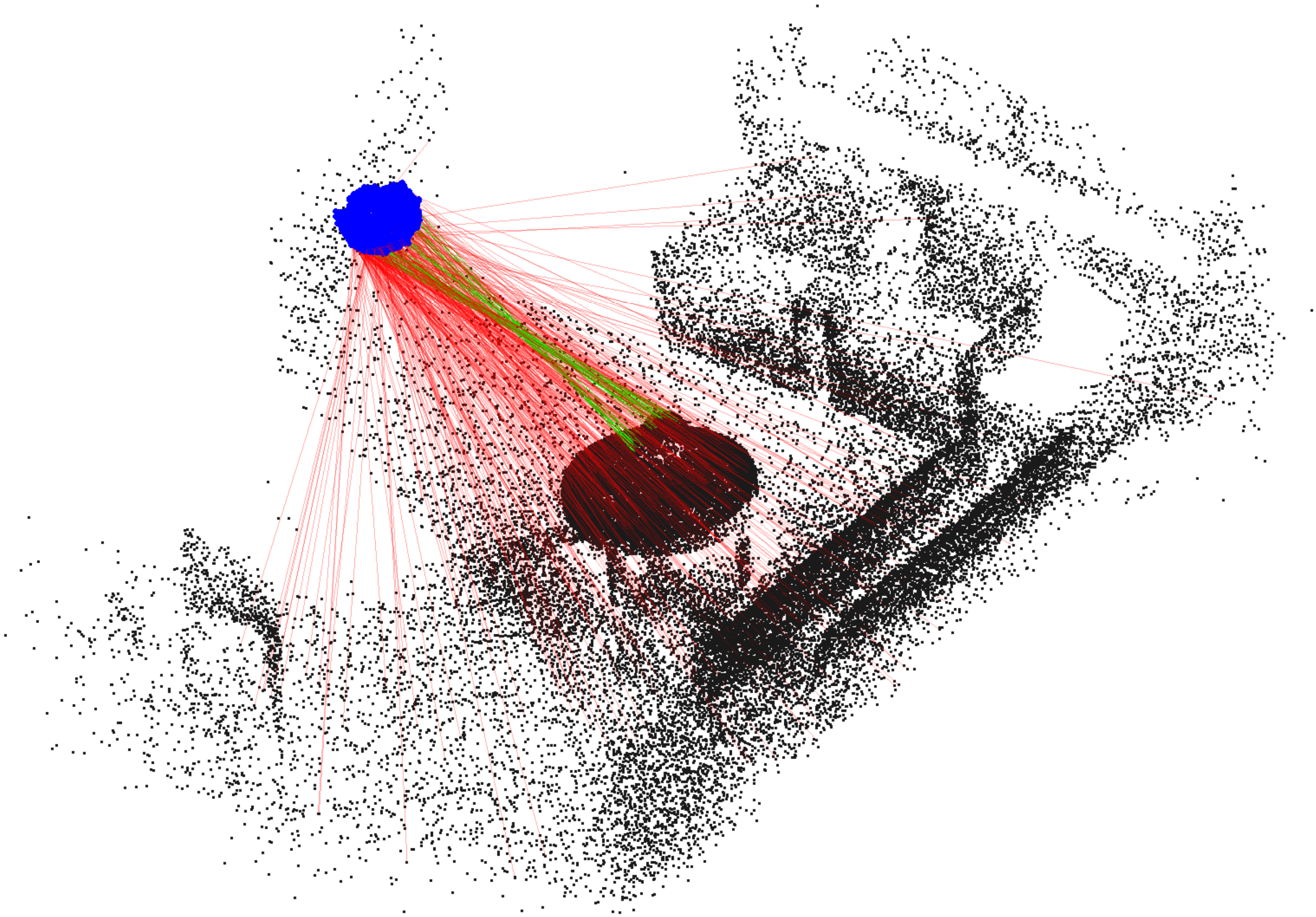}
\includegraphics[width=0.24\linewidth]{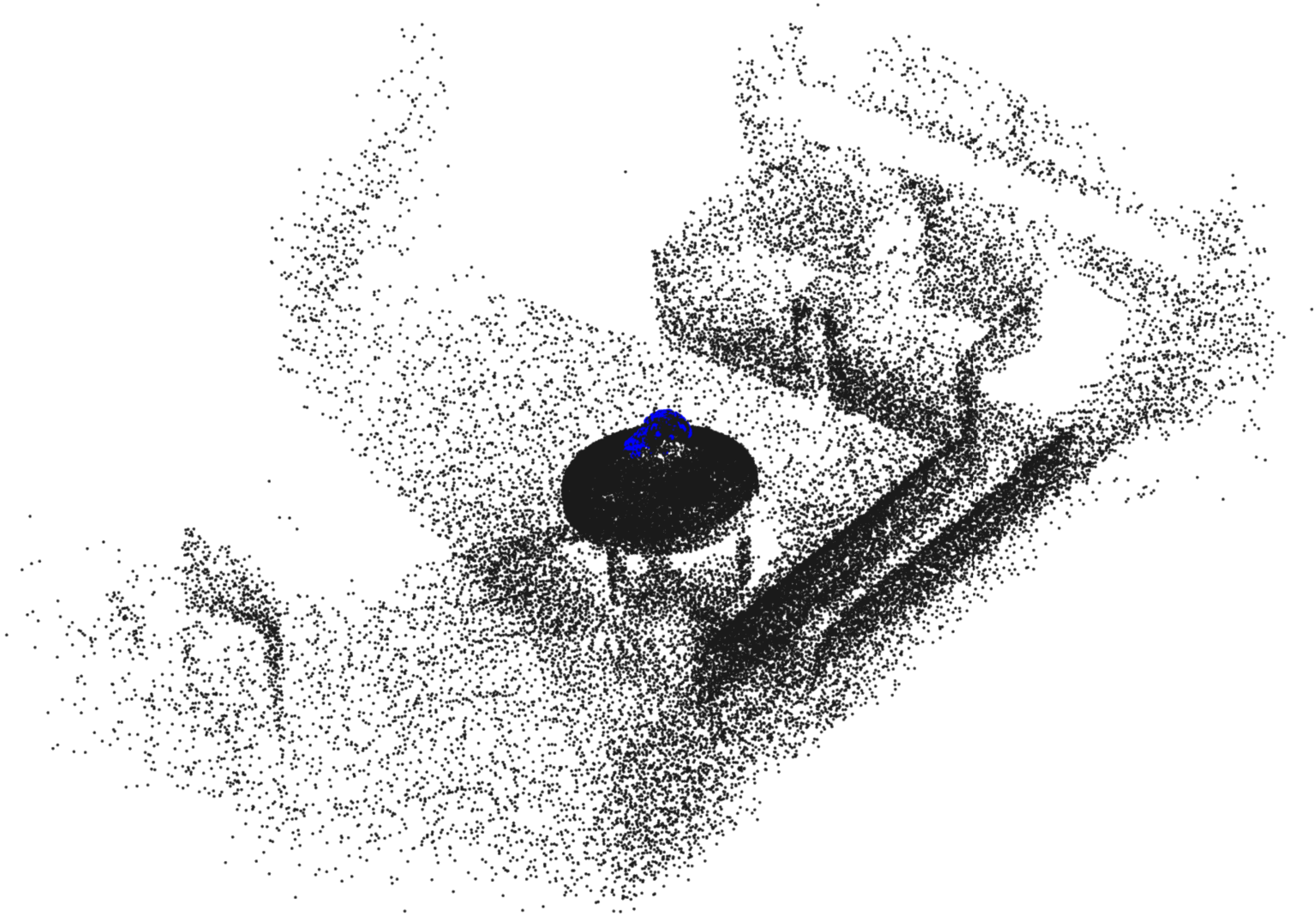}
\end{minipage}
}%

\subfigure[\textit{Scene-12}, Known Scale: ($843,\,85.85\%,\,{0.0835}^{\circ},0.0014 m$), Unknown Scale: ($721,\,90.01\%,\,3.87\times10^{-8},{0.0506}^{\circ},0.0022 m$)]{
\begin{minipage}[t]{1\linewidth}
\centering
\includegraphics[width=0.24\linewidth]{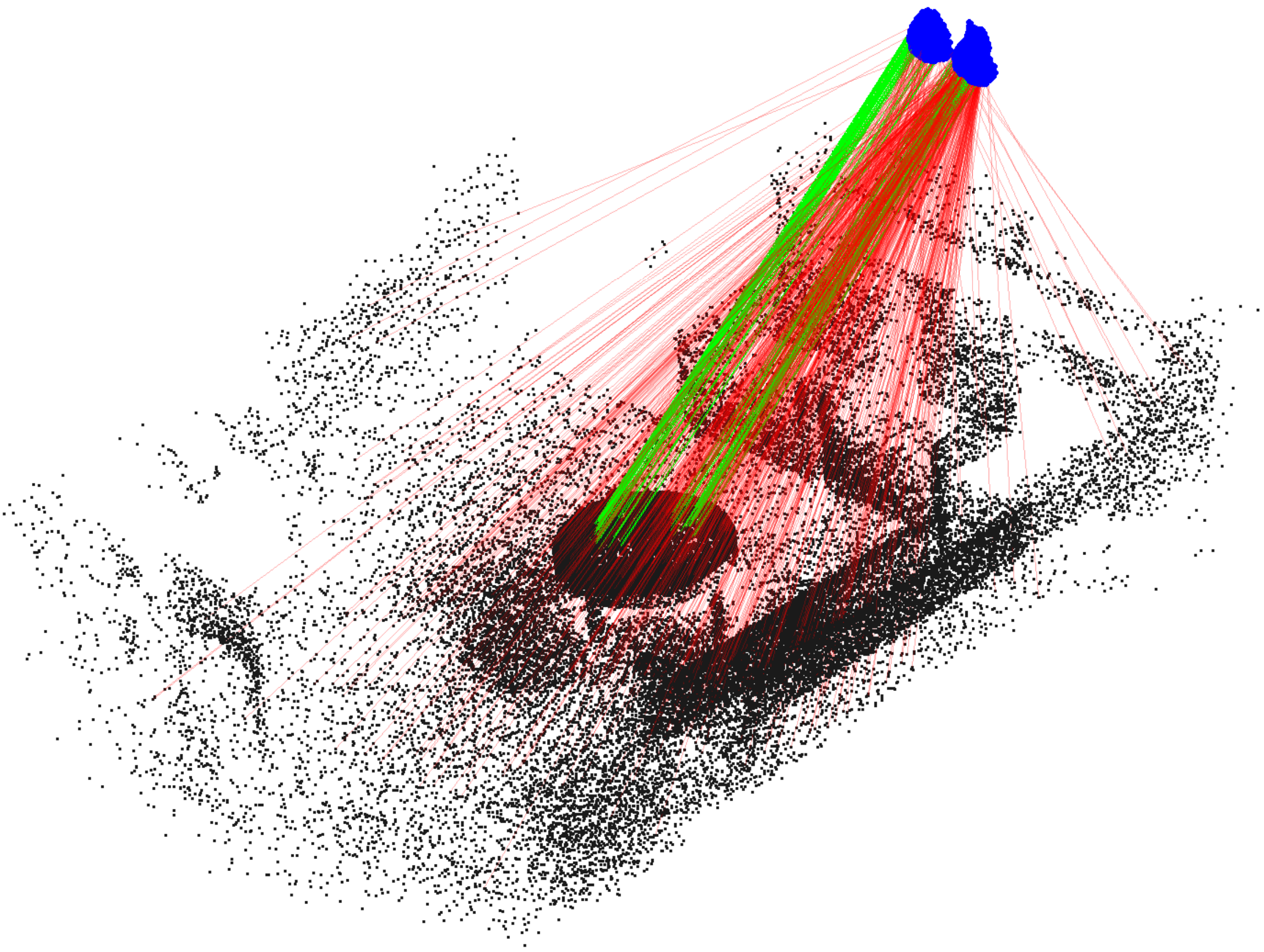}
\includegraphics[width=0.24\linewidth]{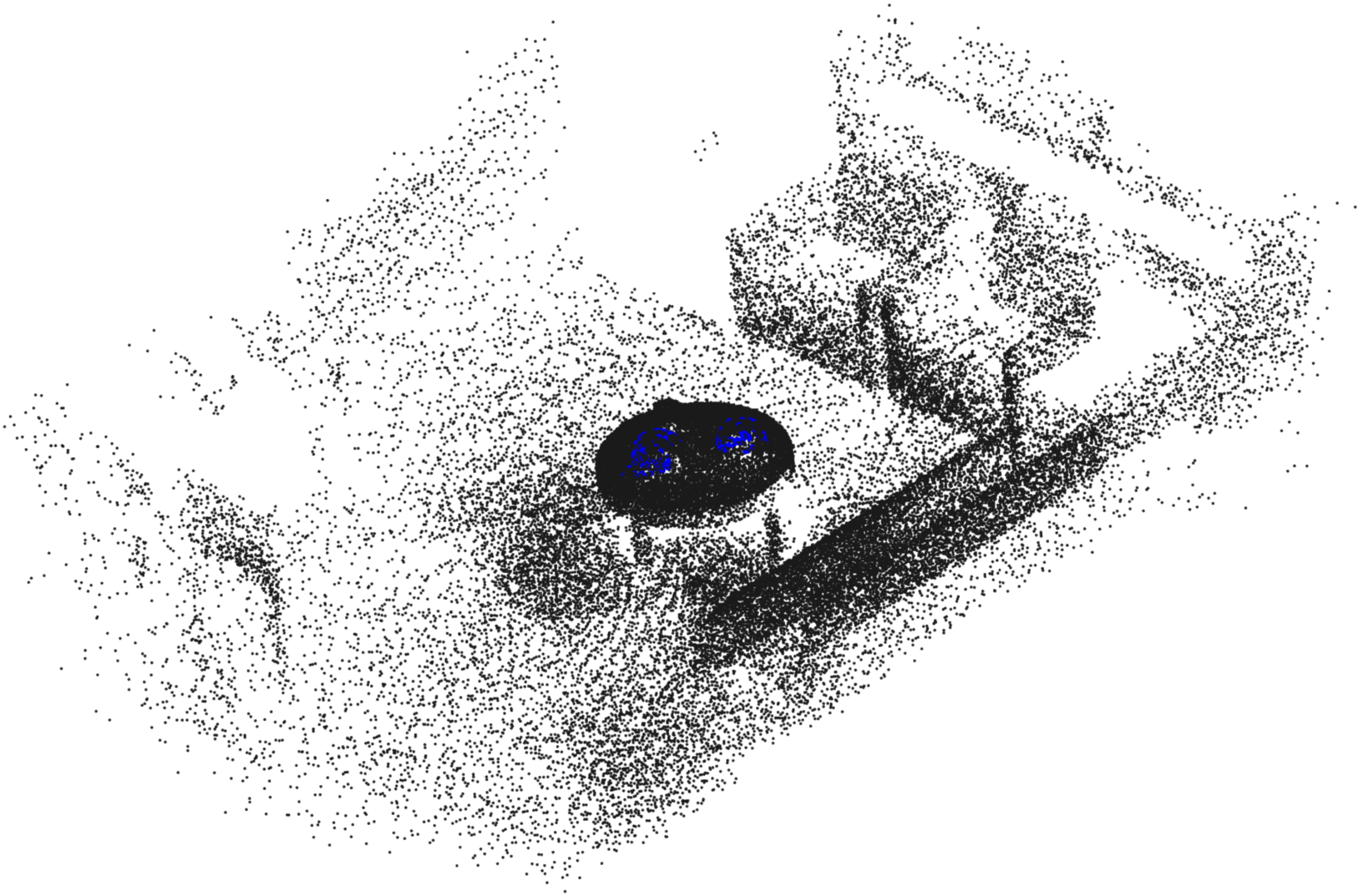}
\includegraphics[width=0.24\linewidth]{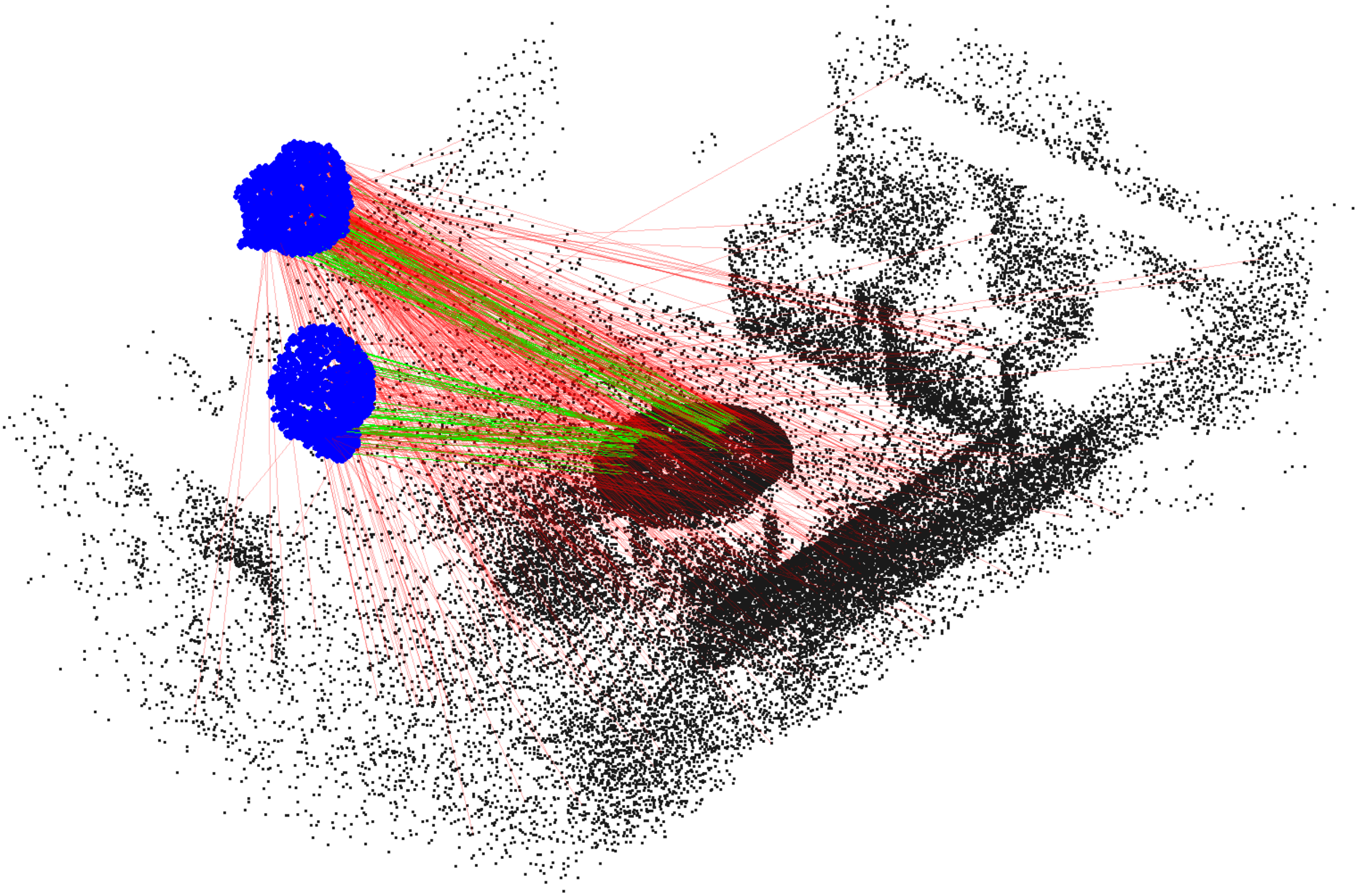}
\includegraphics[width=0.24\linewidth]{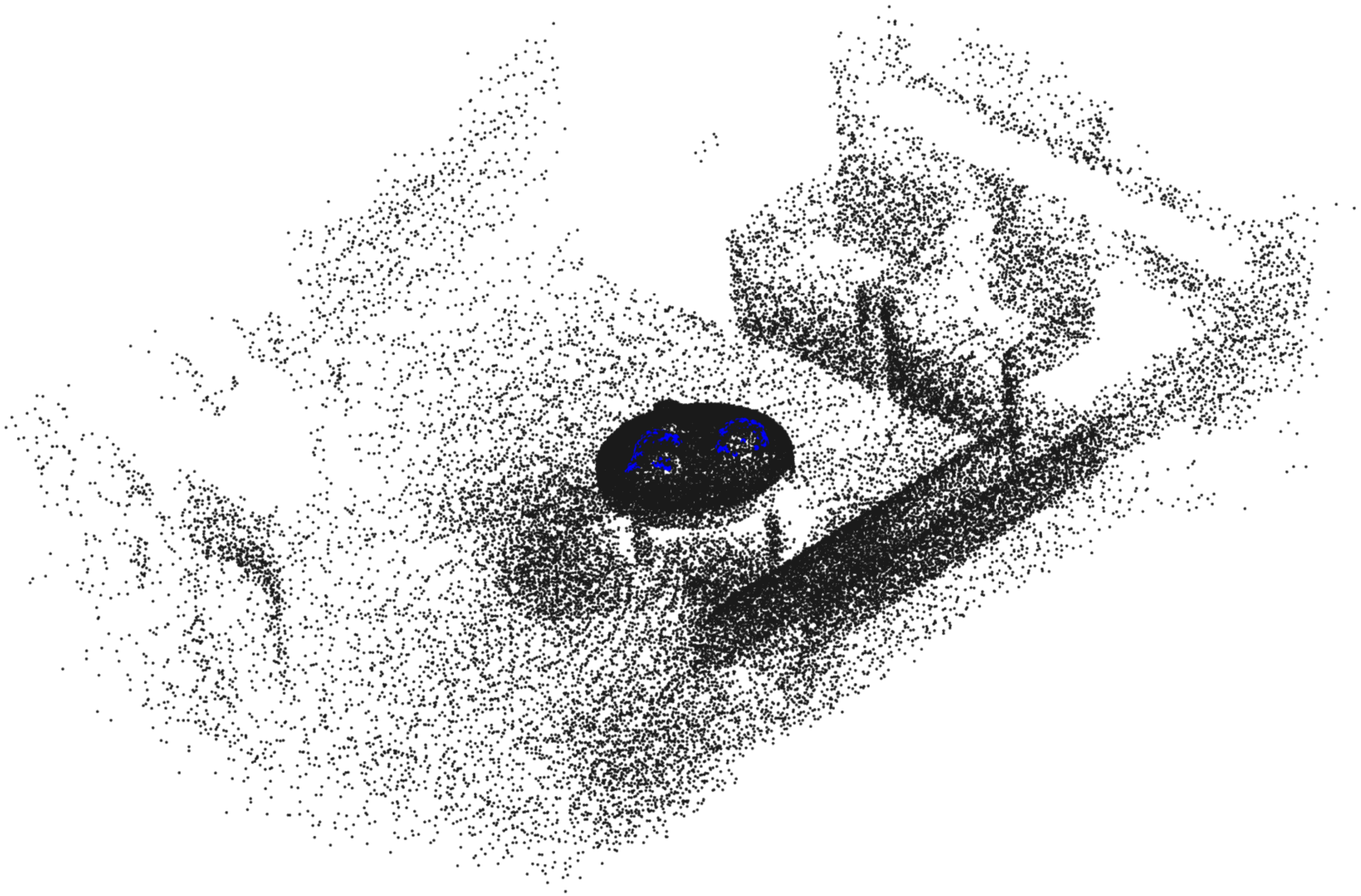}
\end{minipage}
}%

\centering
\caption{3D object localization of \textit{cap} over dataset~\cite{lai2011large}. In each scene from left to right, the images show: correspondences for known-scale registration, qualitative reprojection result of known-scale registration, correspondences for unknown-scale registration, and qualitative reprojection result of unknown-scale registration, and the data below denote: correspondence number, outlier ratio, (scale error), rotation error, and translation error.}
\label{Cap}
\vspace{-9pt}
\end{figure*}

We provide some supplementary experiments in addition to that in the main article. We benchmark the solvers included in the main article over 'Armadillo' and 'Lucy' point clouds in~\cite{curless1996volumetric}, as shown in Figure~\ref{OB2}. We also validate IRON in 3D object localization problems over more scenes from~\cite{lai2011large}, as shown in Figures~\ref{CB} and \ref{Cap}.

\end{document}